\documentclass[letterpaper]{article} 
\usepackage{aaai25}  
\usepackage{times}  
\usepackage{helvet}  
\usepackage{courier}  
\usepackage[hyphens]{url}  
\usepackage{graphicx} 
\usepackage{natbib}  
\usepackage{caption} 
\usepackage{algorithm}

\usepackage{newfloat}
\usepackage{listings}
\DeclareCaptionStyle{ruled}{labelfont=normalfont,labelsep=colon,strut=off} 
\floatstyle{ruled}
\newfloat{listing}{tb}{lst}{}
\floatname{listing}{Listing}
\usepackage{harvey-tables}
\usepackage{common}

\title{\ourtitle}
\author{
    \qzhao, \chris
    \\
}
\affiliations{
    \kastel\\
    \kit
}

\def\appendixref{1}

\usepackage{bibentry}

\begin{document}

\maketitle


\begin{abstract}
\noindent
The community has recently developed various training-time defenses to counter neural backdoors introduced through data~poisoning.
In light of the observation that a model learns poisonous samples responsible for the backdoor easier than benign samples, these approaches either use a fixed threshold of the training loss for splitting~\cite{Li2021antibackdoor, Huang2022backdoor, Chen2022DST} or iteratively learn a reference model as an oracle for identifying benign samples~\cite{Gao2023asd, Zhang2023backdoor}.
In particular, the latter has proven effective for anti-backdoor learning. 
%
Our~method, \ourmethod, \mbox{leverages} a~similar yet crucially different technique: learning an oracle for poisonous rather than benign samples.
Learning a~\emph{backdoored reference model} is significantly easier than learning a reference model on benign~data.
Consequently, we can identify poisonous samples much more accurately than related work identifies benign samples. 
This crucial difference enables near-perfect backdoor removal as we demonstrate in our evaluation.
\ourmethod substantially outperforms related approaches across attack types, datasets, and architectures, lowering the attack success rate to the very minimum at a negligible loss in natural accuracy. 
\end{abstract}



\section{Introduction}

%
Learning an expressive deep neural network (DNN) requires large amounts of training data, which is oftentimes retrieved from third-party resources in practice~\citep{Carlini2023poisoning}.
Using such an external dataset without review may introduce security threats via data poisoning~\citep{Biggio2018Wild}.
The adversary may sneak a small portion of poisonous samples into the training dataset to introduce a neural backdoor.
Such a backdoor shortcuts the prediction toward a predefined target label based on a trigger pattern~\cite{Gu2017badnet, Chen2017blend, Liu2018trojan, Nguyen2020IAB, Nguyen2021wanet, Barni2019sig} and can be established via data poisoning in two ways: First, \emph{dirty-label attacks}~\citep{Gu2017badnet, Chen2017blend, Liu2018trojan, Nguyen2020IAB, Nguyen2021wanet} construct the trigger pattern on poisonous samples and relabel them to the target.
Second, \emph{clean-label attacks}~\citep{turner2019labelconsistent, Shafahi2018poison, Zhao2023clean} strategically modify samples from the target class but do \emph{not} change their labels.

\newcommand{\method}[1]{\makebox[0pt][l]{#1}\phantom{\dst}}

\begin{table}[!t]\vspace*{-1mm}
  \caption{Comparison of training-time backdoor defenses.}
  \label{tab:funcCompare}\vspace*{-1mm}
  \centering
  \setlength{\tabcolsep}{6pt}
  \tablesize
  \begin{ablmethodstable}{\linewidth}
	\method{\abl}
	& Unlearn     
	& Fixed    
	& --
	& \arrowlower 
	\\
	\method{\cbd}
	& Suppress 
	& Adaptive 
	& --
	& \arrowlower 
	\\
	\method{\dbd}
	& Data Split  
	& Fixed    
	& --
	& \arrowlower 
	\\
	\method{\dst}
	& Data Split  
	& Fixed    
	& --
	& \arrowlower 
	\\
	\midrule
	\bfseries \ourmethod
	& Data Split 
	& Adaptive  
	& \fullcirc
	& \makecell[t]{\arrowsame} 
	\\ 
	\bottomrule
\end{ablmethodstable}\vspace*{-1mm}
\end{table}

A wide variety of strategies have been proposed to alleviate backdooring attacks.
\iftrue
\emph{Type-1:}%
~{Model-based defenses} either 
reverse-engineer the trigger pattern~\citep[][]{wang2023unicorn, Wang2022rethinking,  Wang2019neural}, 
merely detect the existence of the backdoor~\citep[][]{cai2022randomized, xu2021detecting, wang2020practical}, 
or erase the backdoor from the model~\citep[][]{ Liu2018fineprune, Zhao2020Bridging, Li2021NAD}.
\emph{Type-2:}%
~{Runtime~defenses} conduct differential testing~\citep[][]{Doan2020februus}, break the trigger functionality via data preprocessing~\citep[][]{Qiu2021DeepSweep} or filter out abnormal inputs~\citep[][]{hayase2021SPECTRE, gao2019strip}.
\emph{Type-3:}%
~{Training-time defenses}, in turn, suppress the backdoor during the training, either using prior knowledge of a clean dataset as the reference~\citep[][]{zhou2023dataelixir, Gao2023asd}, or without such~\citep[][]{Li2021antibackdoor, Chen2022DST, Huang2022backdoor, Zhang2023backdoor}.
\else
  (1) \emph{Data preprocessing}~\citep{Qiu2021DeepSweep, Doan2020februus} should break the relation of backdoor trigger to malicious functionality.
  (2) \emph{Post-training backdoor removal}~\citep{Zeng2021adversarial, Zheng2022data, Wu2021adversarial, Li2021NAD, Liu2018fineprune, Zhao2020Bridging} attempts to remove the backdoor functionality of an existing model.
  (3) \emph{Trigger reverse-engineering}~\citep{wang2023unicorn, Chai2022oneshot, Wang2022rethinking, Guan2022fewshot, Wang2019neural} to derive backdoor's trigger pattern to eventually unlearn it.
  (4) \emph{Backdoor detection}~\citep{xiang2022BP2Class, cai2022randomized, xu2021detecting, wang2020practical} to indentify to examine suspicious attack existence, 
  (5) \emph{sample filtering}~\citep{guo2023scale, du2020robust, hayase2021SPECTRE, gao2019strip} to test and filter abnormal inputs,
and (6) \emph{poison suppression}~\citep{Li2021antibackdoor,  Huang2022backdoor, Zhang2023backdoor, Gao2023asd} that aim to suppress the backdoor during the training.
\fi

All the strategies above have slightly different threat models, and only the latter tackles data poisoning at its roots.
\emph{
Mitigating a backdoor during training without having a clean reference dataset is the most practical setting but it also is exceptionally difficult.
}

In this paper, we propose \ourmethod, a novel training-time defense in this setting, preventing backdoor injection by removing poisonous samples from the training data.
In contrast to related work using unlearning~\citep[\abl;][]{Li2021antibackdoor} or backdoor suppression~\citep[\cbd;][]{Zhang2023backdoor}, dataset splitting allows to preserve natural accuracy better.
However, \ourmethod splits the dataset adaptively without a fixed splitting ratio~\citep[\dbd \& \dst;][]{Huang2022backdoor, Chen2022DST}, allowing us to preserve the natural performance much better as summarized in \cref{tab:funcCompare}.

Our method builds on two crucial observations: 
First, we find that the reverse cross-entropy (\rce) component of the symmetrical cross-entropy (\sce) loss as used by prior work~\citep{Gao2023asd, Huang2022backdoor} is mainly responsible for effectively splitting poisonous and benign samples.
%
Second, using the loss to assemble a set of benign samples \citep{Huang2022backdoor, Chen2022DST} is much more difficult than gathering poisonous samples in the same setting. 

We thus learn a \emph{strongly backdoored reference~model} to continuously isolate poisonous samples using this model's \rce loss. 
%
\ourmethod consists of the following four~stages:
\oneb~\textbf{Initialization:} We naively train a model and split the dataset half-and-half in poisoned and benign subsets using the \rce loss.
\twob~\textbf{Learning the backdoor:} We iteratively train the reference model by learning poisonous samples and unlearning benign samples using the split determined in the previous iteration. 
Over multiple rounds, the reference model becomes more and more specific to the backdoor functionality, which in turn improves \rce-loss-based splitting over time.
\threeb~\textbf{Meta-splitting:} Using the \emph{first} reference model from the previous stage, we refine the poisonous subset yield through the \emph{last} reference model.
The former still has a notion of benign functionality allowing to isolate the remaining benign samples from the poisoned subset.
\mbox{\fourb~\textbf{Final~training:}} 
Eventually, we train on the determined benign dataset to yield a perfectly clean model.

This procedure allows \ourmethod to successfully remove a large variety of backdoor attacks across different model architectures and datasets, lowering the attack success rate~(\asr) below \perc{2} \emph{in the worst case} while preserving the natural accuracy.
We hence outperform related work by a large margin. 
%
%
In summary, our contributions are three-fold:
\begin{itemize}[topsep=0.6em]
\item \textbf{Dataset splitting using \rce loss.}
We analyze the commonly used symmetric cross-entropy (\sce)~loss for dataset splitting, finding that its \rce component alone is suited much better for solving the task.
\ourmethod benefits from using \rce yielding more solid and stable splitting performance than related work.
	
\item \textbf{Paradigm shift on using reference models.}
We find that a \emph{strongly backdoored reference model} tells poisonous and benign samples more reliably apart than any benign reference model can.
This is inline with the early observation that poisonous samples are easier to learn than benign samples, but stands in contrast to how reference models are used in related work.	

\item \textbf{Decisive improvement in training-time defense.}
We evaluate against various attacks across three model architectures and three datasets.
None of the related approaches resist all attacks, while \ourmethod suppresses the backdoor consistently and maintains natural accuracy on average and in the worst-case. 
\end{itemize}



\section{Taming Backdoors at Training-Time}
\label{sec:taming}

We start by briefly describing the problem setting, before we analyze using \sce loss for dataset splitting (\cref{sec:sce-split}) and discuss the distribution of benign and poisonous samples based on the \rce loss (\cref{sec:learn-unlearn-analyse}).  

\paragraph{Problem definition}
We consider backdoor injection through data-poisoning, that is, the adversary has no access to the training process, but can manipulate the training data.
Naively training on this manipulated/poisoned training data~\poisonset learns the ``primary task'' (\ie image classification) but also introduces some additional malicious functionality as a ``secondary task'' (\ie the neural backdoor).

More formally, the adversary poisons $\numdata_p$ samples of an existing dataset~$\dataset = \left\{ \left( \inputx_i, \labely_i \right) \right\}_{i=1}^{\numdata}$ containing $\numdata$~samples $\inputx_i \in \mathbb{R}^d$ with the ground-truth label $\labely_i \in \left\{0, 1, \dots \numClass-1 \right\}$, where \numClass denotes the number of classes.
The resulting dataset $\poisonset$ comprises a poisonous subset~\poisonsubset and a clean subset~\cleanset, \ie $\poisonset = \poisonsubset \cup \cleanset$, and has the same size as the original dataset, $|\dataset| = |\poisonset|$. 
The poisoning rate in \poisonset is $\pratio = \frac{\numpoison}{\numdata}$.
%

Training-time defense, aka.~``anti-backdoor learning,'' aims to train a model on \poisonset with high natural accuracy and simultaneously counter the backdoor.
Note, that the model trainer has no prior knowledge of the attack at all.
%
Defenses using ``dataset splitting''~\citep{Huang2022backdoor, Gao2023asd} separate the dataset~\poisonset into a poisoned set~\poisonD and a benign set~\cleanD, which is then used to train the final model.
Thus, \cleanD should contain as few poisonous samples as possible, $\pratio_{bng} = \frac{\sizeof{\cleanD\setminus\cleanset}}{\sizeof{\cleanD}} \approx 0.0$, ideally $\cleanD = \cleanset$~.

\subsection{Analysis of Dataset Splitting Using \sce Loss}
\label{sec:sce-split}
\label{sec:sce}

Prior defenses~\cite{Gao2023asd, Huang2022backdoor} treat poisonous samples as ``label noise,'' and thus, use the \sce loss~\cite{Wang2019symmetric} to isolate them.
\sce loss consists of two terms, cross-entropy~(\ce) and reverse~cross-entropy~(\rce), weighted by $\alpha$ and $\beta$, respectively:
\begin{align*}
	\centering
	\losssce = \alpha \lossce + \beta \losssce
\end{align*}

In terms of the KL divergence~\citep{Kullback1951KL}, optimizing the \ce loss draws the prediction probability $p\left(k|\inputx\right)$ of an input $\inputx$ wrt. a class~$k$ near the ground-truth probability distribution $q\left(k|\inputx\right)$, minimizing $\KL\left(q||p\right)$.
\begin{align*}
	\centering
	\begin{split}
		\lossce & = - \sum_{k=0}^{\numClass-1} q\left(k|\inputx\right) \cdot \log p\left(k|\inputx\right)
		\\
	\end{split}
\end{align*}

When training on data samples with noisy labels~\cite{Wang2019symmetric}, $q\left(k|\inputx\right)$ does not represent the real ground truth, though.
Hence, \sce additionally considers $\KL\left(p||q\right)$ to push predictions on mislabeled samples toward $p\left(k|\inputx\right)$ 
 known as reverse \ce:
\begin{align*}
	\centering
	\begin{split}
		\lossrce & = - \sum_{k=0}^{\numClass-1} p\left(k|\inputx\right) \cdot \log q\left(k|\inputx\right)
		\\
	\end{split}
\end{align*}

Given an arbitrary input $\inputx$ with the fixed label $\labely$, the distribution $q\left(\labely|\inputx\right)$ equals \num{1} while for all other classes $q\left(k|\inputx\right)|_{k \neq \labely}$ is $\epsilon$-small and close to \num{0}.
We define $\RCEconst = - \log \epsilon$ as being positive and rewrite the \rce loss as:
\begin{align*}
	\begin{split}
		\lossrce 
		& =
		- p\left(\labely|\inputx\right) \cdot \log1
		+ \sum_{k \neq \labely} p\left(k|\inputx\right) \cdot \RCEconst
		\\
		& = \RCEconst \cdot (1 - p\left(\labely|\inputx\right))
		\\
	\end{split}
\end{align*}

Apparently, only the prediction of the ground-truth \labely matters in the \rce loss.
Hence, the training converges to the ground-truth distribution for easy-to-learn samples, yielding a \rce loss very close to \num{0}.
Meanwhile, samples with hard-to-learn features have higher \rce loss close to value~\RCEconst. 

\sce loss with default parameters~\cite{Wang2019symmetric} ($\RCEconst = -log 10^{-5}$, $\alpha=0.1$, $\beta=1.0$ for \cifar{10}) weighs the \rce loss significantly stronger than the \ce loss. 
Still, comparing the distributions of all three losses using the example of \dbd~\cite{Huang2022backdoor} reveals that the \ce loss makes \sce loss less effective in distinguishing benign and poisonous samples, while \rce alone allows a clear separation of poisonous samples.
Using \rce rather than \sce loss, hence, is sufficient (and even beneficial in terms of stability) for dataset splitting.
\if\appendixref1
More details are in~\cref{app:splitting-loss}.
\else
More details are in the appendix.
\fi

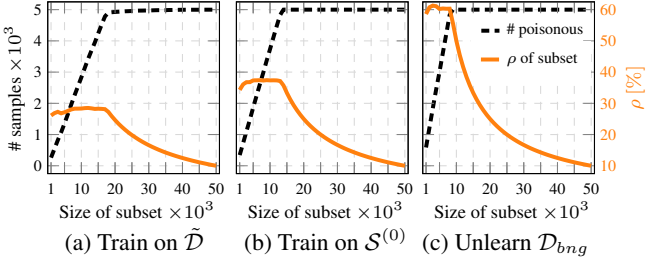
\begin{figure}[h]
	\hskip -4pt
	\begin{subfigure}{.27\linewidth}
		\centering
		\pgfplotstableread[col sep=space]{%
Num	pnum	prate
1	261	26.10
2	537	26.85
3	818	27.27
4	1072	26.80
5	1357	27.14
6	1654	27.57
7	1963	28.04
8	2263	28.29
9	2547	28.30
10	2833	28.33
11	3131	28.46
12	3423	28.52
13	3691	28.39
14	3949	28.21
15	4240	28.27
16	4516	28.22
17	4782	28.13
18	4906	27.26
19	4917	25.88
20	4927	24.64
21	4933	23.49
22	4941	22.46
23	4948	21.51
24	4951	20.63
25	4958	19.83
26	4965	19.10
27	4969	18.40
28	4970	17.75
29	4973	17.15
30	4976	16.59
31	4978	16.06
32	4982	15.57
33	4984	15.10
34	4985	14.66
35	4985	14.24
36	4989	13.86
37	4990	13.49
38	4992	13.14
39	4995	12.81
40	4996	12.49
41	4997	12.19
42	4997	11.90
43	4997	11.62
44	4997	11.36
45	4997	11.10
46	4999	10.87
47	4999	10.64
48	4999	10.41
49	4999	10.20
50	5000	10.00
}\poisondist

\begin{tikzpicture}
	
	\begin{axis}[
		height=\linewidth,
		width=\linewidth, 
		enlarge x limits=0.02,
		enlarge y limits=0.05,
		grid=major, 
		grid style={dashed,gray!30},
		xlabel= {Size of subset $\times 10^3$}, 
		ylabel= {\# samples $\times 10^3$},
		xmin=1,
		xmax=50,
		ymin=0,
		ymax=5000,
		xtick={1, 5, 10, 15, 20, 25, 30, 35, 40, 45, 50},
		xticklabels={1,~,10,~,20,~,30,~,40,~,50},
		ytick={0, 1000, ..., 5000},
		yticklabels={0, 1, 2, 3, 4, 5},
		yticklabel style = {font=\ticksize, yshift=0ex, xshift=.3ex},
		xticklabel style = {font=\ticksize},
		ylabel style = {font=\scriptsize, yshift=-5ex},
		xlabel style = {font=\scriptsize, yshift=1.8ex, xshift=0.3ex},
		legend style={at={(0.2,0.55)}, font=\legendsize, anchor=north west, legend columns=1, fill=white, draw=none, 
			nodes={scale=1.0, transform shape}, column sep=0pt},
		legend image post style={scale=0.5, mark size=2pt},
		legend cell align={left},
		scale only axis,
		every axis plot/.append style={line width=0.7pt},
		]
		
		\addplot+[mark=none, densely dashed, mark size=1pt, mark options={solid}, color=black, line width=1.5pt] table[x index = 0, y index=1] 
		{\poisondist};
		
		
	\end{axis}
	
	\begin{axis}[
		height=\linewidth,
		width=\linewidth, 
		enlarge x limits=0.02,
		enlarge y limits=0.05,
		ylabel near ticks, yticklabel pos=right,
		ymajorticks=false,
		xmin=1,
		xmax=50,
		ymin=10,
		ymax=60,
		xtick={1, 5, 10, 15, 20, 25, 30, 35, 40, 45, 50},
		xticklabels=\empty,
		ytick={10, 20, ..., 60},
		yticklabels=\empty, 
		yticklabel style = {font=\scriptsize, yshift=0.3ex, color=secondarycolor},
		xticklabel style = {font=\scriptsize},
		ylabel style = {font=\ticksize, yshift=-5ex},
		xlabel style = {font=\ticksize, yshift=1.5ex},
		legend image post style={scale=1.0},
		legend style={at={(0.1,0.25)}, font=\legendsize, anchor=north west, legend columns=1, fill=white, draw=white, 
			nodes={scale=1.0, transform shape},, column sep=3pt},
		legend cell align={left},
		scale only axis,
		every axis plot/.append style={line width=0.7pt},
		]
		
		\addplot+[mark=none, mark size=.3pt, line width=1pt, mark options={solid}, color=secondarycolor, line width=1.5pt] table[x index = 0, y index=2] 
		{\poisondist};
		
		
	\end{axis}
	
\end{tikzpicture}
		\captionsetup{margin*={0pt, -40pt}}
		\vskip -20pt
		\caption{Train on \poisonset}
		\label{fig:dist_cifar10_blend}
	\end{subfigure}
	\hskip 20pt
	\begin{subfigure}{.27\linewidth}
		\centering
		\vskip 0.7pt
		\pgfplotstableread[col sep=space]{%
Num	pnum	prate
1	343	34.30
2	721	36.05
3	1103	36.77
4	1471	36.78
5	1867	37.34
6	2242	37.37
7	2619	37.41
8	2985	37.31
9	3368	37.42
10	3735	37.35
11	4111	37.37
12	4481	37.34
13	4827	37.13
14	5000	35.71
15	5000	33.33
16	5000	31.25
17	5000	29.41
18	5000	27.78
19	5000	26.32
20	5000	25.00
21	5000	23.81
22	5000	22.73
23	5000	21.74
24	5000	20.83
25	5000	20.00
26	5000	19.23
27	5000	18.52
28	5000	17.86
29	5000	17.24
30	5000	16.67
31	5000	16.13
32	5000	15.62
33	5000	15.15
34	5000	14.71
35	5000	14.29
36	5000	13.89
37	5000	13.51
38	5000	13.16
39	5000	12.82
40	5000	12.50
41	5000	12.20
42	5000	11.90
43	5000	11.63
44	5000	11.36
45	5000	11.11
46	5000	10.87
47	5000	10.64
48	5000	10.42
49	5000	10.20
50	5000	10.00
}\poisondist

\begin{tikzpicture}
	
	\begin{axis}[
		height=\linewidth,
		width=\linewidth, 
		enlarge x limits=0.02,
		enlarge y limits=0.05,
		grid=major, 
		grid style={dashed,gray!30},
		xlabel= {Size of subset $\times 10^3$}, 
		ymajorticks=false,
		xmin=1,
		xmax=50,
		ymin=0,
		ymax=5000,
		xtick={1, 5, 10, 15, 20, 25, 30, 35, 40, 45, 50},
		xticklabels={1,~,10,~,20,~,30,~,40,~,50},
		ytick={0, 1000, ..., 5000},
		yticklabels=\empty, 
		smooth,
		tension=0.1,
		yticklabel style = {font=\ticksize, yshift=0.3ex},
		xticklabel style = {font=\ticksize},
		ylabel style = {font=\scriptsize, yshift=-4.5ex},
		xlabel style = {font=\scriptsize, yshift=1.8ex, xshift=0.3ex},
		legend style={at={(0.27,0.92)}, font=\legendsize, anchor=north west, legend columns=1, fill=white, draw=none, minimum height=0cm, nodes={scale=1.0, transform shape}, column sep=0pt},
		legend image post style={scale=0.5, mark size=2pt},
		legend cell align={left},
		scale only axis,
		every axis plot/.append style={line width=0.7pt},
		]
		
		\addplot+[mark=none, densely dashed, mark size=1pt, mark options={solid}, color=black, line 
		width=1.5pt] table[x index = 0, y index=1] 
		{\poisondist};
		
	\end{axis}
	
	\begin{axis}[
		height=\linewidth,
		width=\linewidth, 
		enlarge x limits=0.02,
		enlarge y limits=0.05,
		ylabel near ticks, yticklabel pos=right,
		xmin=1,
		xmax=50,
		ymin=10,
		ymax=60,
		xtick={1, 5, 10, 15, 20, 25, 30, 35, 40, 45, 50},
		xticklabels=\empty,
		ytick={10, 20, ..., 60},
		yticklabels=\empty, 
		smooth,
		tension=0.1,
		yticklabel style = {font=\ticksize, yshift=0.3ex, color=secondarycolor},
		xticklabel style = {font=\ticksize},
		ylabel style = {font=\scriptsize, yshift=0.5ex},
		xlabel style = {font=\scriptsize, yshift=1.5ex},
		legend image post style={scale=1.0},
		legend style={at={(0.27,0.78)}, font=\legendsize, anchor=north west, legend columns=1, fill=white, draw=white, 
			nodes={scale=1.0, transform shape},, column sep=3pt},
		legend cell align={left},
		scale only axis,
		every axis plot/.append style={line width=0.7pt},
		]
		
		\addplot+[mark=, mark size=.7pt, mark options={solid}, color=secondarycolor, line width=1.5pt] table[x index = 0, y index=2] 
		{\poisondist};
		
	\end{axis}
	
\end{tikzpicture}
		\captionsetup{margin*={0pt, -10pt}}
		\vskip -20pt
		\caption{Train on \subsetP[0]}
		\label{fig:dist_c1_cifar10_blend}
	\end{subfigure}
	\hskip 3pt
	\begin{subfigure}{.27\linewidth}
		\centering
		\pgfplotstableread[col sep=space]{%
Num	pnum	prate
1	586	58.60
2	1215	60.75
3	1839	61.30
4	2444	61.10
5	3011	60.22
6	3622	60.37
7	4220	60.29
8	4825	60.31
9	5000	55.56
10	5000	50.00
11	5000	45.45
12	5000	41.67
13	5000	38.46
14	5000	35.71
15	5000	33.33
16	5000	31.25
17	5000	29.41
18	5000	27.78
19	5000	26.32
20	5000	25.00
21	5000	23.81
22	5000	22.73
23	5000	21.74
24	5000	20.83
25	5000	20.00
26	5000	19.23
27	5000	18.52
28	5000	17.86
29	5000	17.24
30	5000	16.67
31	5000	16.13
32	5000	15.62
33	5000	15.15
34	5000	14.71
35	5000	14.29
36	5000	13.89
37	5000	13.51
38	5000	13.16
39	5000	12.82
40	5000	12.50
41	5000	12.20
42	5000	11.90
43	5000	11.63
44	5000	11.36
45	5000	11.11
46	5000	10.87
47	5000	10.64
48	5000	10.42
49	5000	10.20
50	5000	10.00
}\poisondist

\begin{tikzpicture}
	
	\begin{axis}[
		height=\linewidth,
		width=\linewidth, 
		enlarge x limits=0.02,
		enlarge y limits=0.05,
		grid=major, 
		grid style={dashed,gray!30},
		xlabel= {Size of subset $\times 10^3$}, 
		ymajorticks=false,
		xmin=1,
		xmax=50,
		ymin=0,
		ymax=5000,
		xtick={1, 5, 10, 15, 20, 25, 30, 35, 40, 45, 50},
		xticklabels={1,~,10,~,20,~,30,~,40,~,50},
		ytick={0, 1000, ..., 5000},
		yticklabels={0, 10, 20, 30, 40, 50, 60, 70, 80, 90, 100},
		smooth,
		tension=0.1,
		yticklabel style = {font=\ticksize, yshift=0.3ex},
		xticklabel style = {font=\ticksize},
		ylabel style = {font=\scriptsize, yshift=-4.5ex},
		xlabel style = {font=\scriptsize, yshift=1.8ex, xshift=0.3ex},
		legend style={at={(0.28,0.94)}, font=\legendsize, anchor=north west, legend columns=1, fill=white, draw=none, 
		nodes={scale=1.0, transform shape}, column sep=0pt},
		legend image post style={scale=0.4, mark size=2pt},
		legend cell align={left},
		scale only axis,
		every axis plot/.append style={line width=0.7pt},
		]
		
		\addplot+[mark=none, densely dashed, mark size=1pt, mark options={solid}, color=black, line 
		width=1.5pt] table[x index = 0, y index=1] 
		{\poisondist};
		
		\addlegendentry{\# poisonous};
		
	\end{axis}
	
	\begin{axis}[
		height=\linewidth,
		width=\linewidth, 
		enlarge x limits=0.02,
		enlarge y limits=0.05,
		ylabel= \textcolor{secondarycolor}{\pratio $\left[\%\right]$},
		ylabel near ticks, yticklabel pos=right,
		xmin=1,
		xmax=50,
		ymin=10,
		ymax=60,
		xtick={1, 5, 10, 15, 20, 25, 30, 35, 40, 45, 50},
		xticklabels=\empty,
		ytick={10, 20, ..., 60},
		yticklabels={10, 20, 30, 40, 50, 60},
		yticklabel style = {font=\ticksize, yshift=0ex, xshift=-0.3ex, color=secondarycolor},
		xticklabel style = {font=\ticksize},
		ylabel style = {font=\scriptsize, yshift=.7ex},
		xlabel style = {font=\scriptsize, yshift=1.5ex},
		legend image post style={scale=1.0},
		legend style={at={(0.28,0.77)}, font=\legendsize, anchor=north west, legend columns=1, fill=white, draw=none, 
		nodes={scale=1.0, transform shape}, line width=.1pt, column sep=0pt},
		legend image post style={scale=0.4, mark size=2pt},
		legend cell align={left},
		scale only axis,
		every axis plot/.append style={line width=0.7pt},
		]
		
		\addplot+[mark=none, mark size=.7pt, mark options={solid}, color=secondarycolor, line width=1.5pt] table[x index = 0, y index=2] 
		{\poisondist};
		\addlegendentry{\pratio of subset};
		
	\end{axis}
	
\end{tikzpicture}
		\captionsetup{margin*={0pt, -5pt}}
		\vskip -20pt
		\caption{Unlearn \cleanD}
		\label{fig:dist_c1_unlearn_cifar10_blend}
	\end{subfigure}
	\caption{
		\revision{%
		Subsets of incremental size chosen in ascending order of \rce loss based on a \resnet{18} trained on a \cifar{10} under \blend attack~\citep{Chen2017blend} at $\pratio=\perc{10}$.
	}
	}
	\vskip -\baselineskip
	\label{fig:dist_cifar10}
\end{figure}

\begin{figure*}[!t]
	\vspace*{-2mm}%
	\includegraphics[width=\linewidth, page=1]{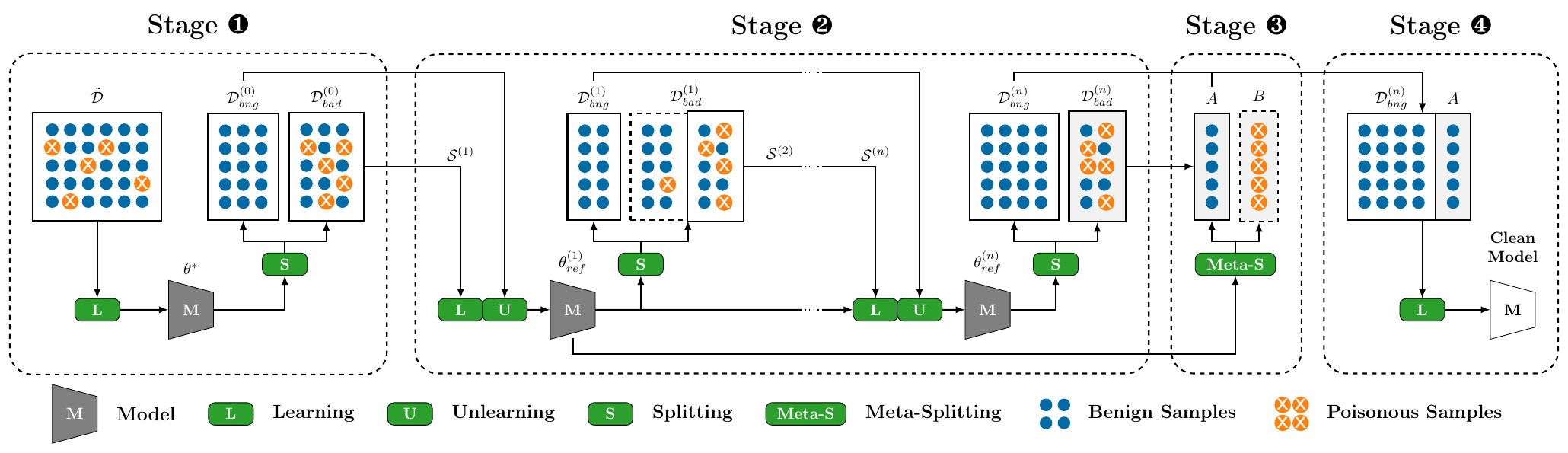}
	\vspace*{-2mm}
	\caption{Overview depiction of our method \ourmethod. The individual stages are described in \cref{sec:stage-1,sec:stage-2,sec:stage-3,sec:stage-4}.}
	\label{fig:overview}
\end{figure*}

\subsection{Distribution of Benign and Poisonous Samples}
\label{sec:learn-unlearn-analyse}

Although the model learns poisonous samples (and thus the backdoor) easier than benign samples in the general case~\cite{Li2021antibackdoor}, after training, some benign samples still have a similar loss as poisonous samples.
Hence, a fixed threshold on the loss \emph{cannot} ensure splitting the dataset precisely.
However, the majority of samples with low loss are indeed poisonous allowing to sample a subset with a high poisoning rate.
Increasing the poisoning rate can intensifies the backdoor in a model~\cite{Li2024Theoretical, Wu2022backdoorbench}.

Training on the entire dataset~\poisonset yields low loss values for most poisonous samples already as shown in~\cref{fig:dist_cifar10_blend}.
If we consider the \perc{50} of the samples with the lowest \rce loss to form a subset \subsetP[1], this subset will naturally have a higher poisoning rate.
Retraining on \subsetP[1] yields the distribution shown in \cref{fig:dist_c1_cifar10_blend} with even more poisonous samples having low loss.
This gives rise to adaptive dataset splitting similar as \citet{Zhang2023backdoor} proposed for \cbd \emph{but focusing on poisonous samples instead of benign samples}.

If we additionally unlearn the \perc{50} not selected for \subsetP[1] (samples with high \rce loss and thus mostly benign) by maximizing the training loss, we yield even more poisonous samples with low loss in the resulting model as shown in \cref{fig:dist_c1_unlearn_cifar10_blend}.
The subset formed by the \SI{10}{k}~samples with lowest loss has an exceptionally high poisoning rate, forming the ideal basis for the next round of splitting.


Learning a \emph{strongly backdoored model} can serve as an oracle for poisonous samples.
In the next section, we introduce our method that uses such an oracle as ``reference model'' to identify and filter out poisonous samples, before learning a backdoor-free model on the remaining clean data samples.

\section{\ourmethod: Learning a Backdoor Oracle}
\label{sec:method}

We propose a novel training-time defense against neural backdoors based on dataset splitting.
In contrast to related work, we tackle the problem from a different perspective: We focus on poisonous samples rather than benign samples and \emph{iteratively learn a strongly backdoored model as an oracle for poisonous samples.}
This way our method does not require a clean reference dataset, but bootstraps iterative splitting with the most obvious poisonous samples and improves continuously.
\cref{fig:overview} depicts the multi-stage, working principle of our method, \ourmethod. Additionally, we provide an algorithmic description in the appendix at~\cref{alg:method}.
In the following subsections, we elaborate on each stage individually.

\subsection{Initialization (\stage{1})} 
\label{sec:stage-1}

We begin by training \EpochsInit epochs on \poisonset to obtain a naively trained model \paramsnaive.
Based on the observations made in \cref{sec:taming}, we retrieve the \perc{50} of the samples with the lowest \rce loss as the initial poisonous subset \poisonD[0], which we use for the further processing in the next stage entirely, $\subsetP[1] = \poisonD[0]$.
The remaining \perc{50} form the initial benign set \cleanD[0].
Note that this yields a reasonable split because poisonous samples are learned easier than benign samples in general~\citep{Li2021antibackdoor}.
The following stages refine this split over multiple rounds to constantly improve separation up until we are ready to learn the final backdoor-free model.

\subsection{Learning the Backdoor (\stage{2})}
\label{sec:stage-2}

We discard the initial naively trained model~\paramsnaive and train a new reference model~\paramsref from scratch that we iteratively fine-tune over \Niters~rounds.
Each round fulfills two tasks:
First, we enhance the reference model's notion of the backdoor by fine-tuning\footnote{\paramsref[0] is randomly initialized at the beginning of \stage{2}.} \paramsref[i-1] yielding~\paramsref[i].
Second, we use the latter to split \poisonset into a poisoned set \poisonD[i] and a benign set~\cleanD[i].
The poisoned set serves as basis for sub-sampling data into \subsetP[i+1] used for the next iteration.

%
\paragraph{Enhancing the backdoored reference model}
We use a learning-unlearning strategy to strongly internalize the backdoor. 
First, we train on \subsetP[i] for \EpochsPoi epochs and employ the local gradient ascent (\lga)~loss~\citep{Li2021antibackdoor} with $\gamma$ set to \num{0.01} to avoid overfitting to the few benign samples contained in the subset \subsetP[i]:
\begin{equation*}
	\label{eq:LGA}
	\centering
	\lgaloss
	=
	\hskip -11pt
	\sum_{\left( \inputx, \labely \right) \in \subsetP[i]}
	\hskip -10pt
	sign\left( \lossce \left( \inputx, \labely, \paramsref[i] \right) - \gamma \right) 
	\cdot
	\lossce \left( \inputx, \labely, \paramsref[i] \right)
\end{equation*}

Next, we unlearn the benign samples in the benign subset \cleanD[i-1] by \emph{maximizing} the \ce loss.
We use the following (negative) loss function, with $\lambda=0.01$ to stabilize the training.
The indicator function $\mathds{1}\left(\cdot\right)$ ensures to only consider samples for unlearning that were predicted correctly by the reference model: 
\begin{equation*}
	\label{eq:UL}
	\centering
	\ulloss
	= 
	\sum_{\left( \inputx, \labely \right) \in \cleanD[i-1]}
	-
	\lambda 
	\cdot 
	\mathds{1} \left( f_{\paramsref[i]} \left(\inputx\right) = \labely \right)
	\cdot
	\lossce \left( \inputx, \labely, \paramsref[i] \right)
\end{equation*}
We perform unlearning for only one epoch per round.
This way, we avoid a learning divergence and help the model to stably unlearn the benign samples.

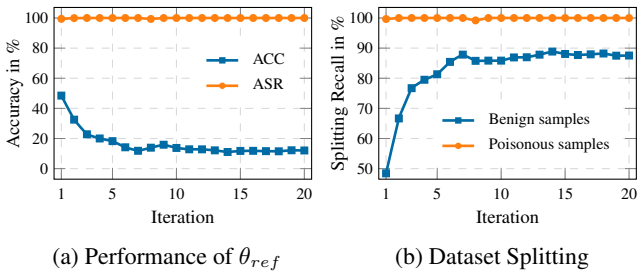
\begin{figure}[!b]
	\hskip -2pt
	\begin{subfigure}{.45\linewidth}
		\pgfplotstableread[col sep=space]{%
	cycle	prec	crec	acc	asr
	0	96.80	55.20	82.38	99.01
}\pretrain

\pgfplotstableread[col sep=space]{%
	cycle	prec	crec	acc	asr
	1	99.66	48.44	48.48	99.42
	2	99.98	66.66	32.52	99.92
	3	100.00	76.71	22.80	100.00
	4	100.00	79.50	20.02	100.00
	5	100.00	81.32	18.25	100.00
	6	100.00	85.43	14.17	100.00
	7	100.00	87.84	11.85	100.00
	8	99.20	85.78	13.91	99.37
	9	100.00	85.85	15.87	99.99
	10	100.00	85.85	13.73	99.99
	11	100.00	86.91	12.83	99.99
	12	100.00	86.97	12.81	100.00
	13	100.00	87.83	12.14	100.00
	14	100.00	88.86	11.06	100.00
	15	100.00	88.06	11.75	100.00
	16	100.00	87.72	11.82	100.00
	17	100.00	87.94	11.61	100.00
	18	100.00	88.21	11.51	100.00
	19	100.00	87.49	12.19	100.00
	20	100.00	87.54	12.05	100.00
}\poisontrain

\begin{tikzpicture}	
	\begin{axis}[
		height=.6\linewidth,
		width=.9\linewidth, 
		enlarge x limits=0.03,
		enlarge y limits=0.07,
		grid=major, 
		grid style={dashed,gray!30},
		xlabel= Iteration, 
		ylabel= Accuracy in \%,
		xmin=1,
		xmax=20,
		ymin=0,
		ymax=100,
		xtick={1, 5, 10, 15, 20},
		xticklabels={1, 5, 10, 15, 20},
		ytick={0, 20, ..., 100},
		yticklabels={0, 20, 40, 60, 80, 100},
		smooth,
		tension=0.1,
		yticklabel style = {font=\ticksize, xshift=0.3ex},
		xticklabel style = {font=\ticksize},
		ylabel style = {font=\scriptsize, yshift=-4.5ex},
		xlabel style = {font=\scriptsize, yshift=1.5ex},
		legend style={at={(0.95,0.80)}, font=\legendsize, anchor=north east, 
			legend columns=1, fill=white, draw=white, 
			nodes={scale=1.0, transform shape}, column sep=3pt},
		legend cell align={left},
		legend image post style={scale=0.7, mark size=3pt},
		scale only axis,
		every axis plot/.append style={line width=0.7pt},
		]
		
		\addplot+[mark=square*, mark options={solid, mark size=1pt}, 
		color=primarycolor, line width=1pt] table[x index = 0, y index=3] 
		{\poisontrain};
		
		\addplot+[mark=*, mark options={solid, mark size=1pt}, 
		color=secondarycolor, line width=1pt] table[x index = 0, y index=4]
		{\poisontrain};
		
		\addlegendentry{\acc};
		\addlegendentry{\asr};
		
		
		
	\end{axis}
	
\end{tikzpicture}
		\vskip -12pt
		\captionsetup{margin*={20pt, 0pt}}
		\caption{Performance of \paramsref}
		\label{fig:split-acc}
	\end{subfigure}
	\hskip 12pt
	\begin{subfigure}{.45\linewidth}
		\pgfplotstableread[col sep=space]{%
	cycle	prec	crec	acc	asr
	0	96.80	55.20	82.38	99.01
}\pretrain

\pgfplotstableread[col sep=space]{%
cycle	prec	crec	acc	asr
1	99.66	48.44	48.48	99.42
2	99.98	66.66	32.52	99.92
3	100.00	76.71	22.80	100.00
4	100.00	79.50	20.02	100.00
5	100.00	81.32	18.25	100.00
6	100.00	85.43	14.17	100.00
7	100.00	87.84	11.85	100.00
8	99.20	85.78	13.91	99.37
9	100.00	85.85	15.87	99.99
10	100.00	85.85	13.73	99.99
11	100.00	86.91	12.83	99.99
12	100.00	86.97	12.81	100.00
13	100.00	87.83	12.14	100.00
14	100.00	88.86	11.06	100.00
15	100.00	88.06	11.75	100.00
16	100.00	87.72	11.82	100.00
17	100.00	87.94	11.61	100.00
18	100.00	88.21	11.51	100.00
19	100.00	87.49	12.19	100.00
20	100.00	87.54	12.05	100.00
}\poisontrain

\begin{tikzpicture}
	
	\begin{axis}[
		height=.6\linewidth,
		width=.9\linewidth, 
		enlarge x limits=0.03,
		enlarge y limits=0.07,
		grid=major, 
		grid style={dashed,gray!30},
		xlabel= Iteration, 
		ylabel= Splitting Recall in \%,
		xmin=1,
		xmax=20,
		ymin=50,
		ymax=100,
		xtick={1, 5, 10, 15, 20},
		xticklabels={1, 5, 10, 15, 20},
		ytick={50, 60, ..., 100},
		yticklabels={50, 60, 70, 80, 90, 100},
		smooth,
		tension=0.1,
		yticklabel style = {font=\ticksize, xshift=0.3ex},
		xticklabel style = {font=\ticksize},
		ylabel style = {font=\scriptsize, yshift=-4.5ex},
		xlabel style = {font=\scriptsize, yshift=1.5ex},
		legend style={at={(0.95,0.45)}, font=\legendsize, anchor=north east, 
			legend columns=1, fill=white, draw=white, 
			nodes={scale=1.0, transform shape}, column sep=3pt},
		legend cell align={left},
		legend image post style={scale=0.7, mark size=3pt},
		scale only axis,
		every axis plot/.append style={line width=0.7pt},
		]
		
		\addplot+[mark=square*, mark options={solid, mark size=1pt}, color=primarycolor, line width=1pt] table[x index = 0, y index=2]
		{\poisontrain};
		
		\addplot+[mark=*, mark options={solid, mark size=1pt}, color=secondarycolor, line width=1pt] table[x index = 0, y index=1] 
		{\poisontrain};
		
		\addlegendentry{Benign samples};
		\addlegendentry{Poisonous samples};
		
		
		
	\end{axis}
	
\end{tikzpicture}
		\vskip -12pt
		\captionsetup{margin*={20pt, 0pt}}
		\caption{Dataset Splitting}
		\label{fig:split-prec}
	\end{subfigure}
	\caption{\ourmethod's splitting performance using \resnet{18} on \cifar{10} poisoned by \blend attack~\citep{Chen2017blend}.
	}
	\label{fig:learn_cifar10_badnets}
\end{figure}

\paragraph{Splitting dataset \poisonset and composing \subsetP[i+1]}
The reference model \paramsref[i] encodes the backdoor much better than benign data.
There hence exists a huge gap in the \rce loss (\cf~\cref{fig:tsne-cifar10-badnets-c1}) that separates poisonous and most benign samples.
The \rce loss has a fixed output range of $\left[0, \RCEconst\right]$, so that a threshold of $\sfrac{\RCEconst}{2}$ splits the dataset well in a poisonous part \poisonD[i] (low \rce loss) and clean part \cleanD[i] (high \rce loss).
We then retrieve \perc{50} of the samples with the lowest \rce loss from \poisonD[i] to be used as \subsetP[i+1] for the next iteration.

Note that it may happen that some classes are not represented in \subsetP[i+1]. 
To counteract any bias, we additionally append samples with low \rce loss from $\poisonset\setminus\subsetP[i+1]$ so that all classes have at least \perc{1} of the original amount of samples. 
Doing so will not weaken the backdoor in the reference model but preserves the learning behavior across all classes.

\cref{fig:learn_cifar10_badnets} shows the attack success rate (\asr) and natural accuracy (\acc) of the reference models~\paramsref[i] on the left and the splitting performance measured as recall on the right.
The process stabilizes starting at round~\num{10}, yielding a decent split in poisonous and benign samples.
However, we cannot ensure perfect splitting this way as some benign samples' loss overlaps with that of poisonous samples (\cf~\cref{fig:tsne-cifar10-badnets-end}). 
We thus employ ``meta splitting'' as described next.

\begin{figure}[!b]
\begin{minipage}{.421\linewidth}
	\input{figures/cifar10-resnet18-blend-wo-metasplit-main.tex}
	\captionsetup{margin*={0pt, 0pt}}
	\captionof{figure}{Train on \cleanD[n] of \cifar{10} poisoned by \blend w/o meta-splitting.}
	\label{fig:no-meta-split}
\end{minipage}
~~
\begin{minipage}{.555\linewidth}
	\input{figures/cifar10-resnet18-blend-RCE-split-meta.tex}
	\captionof{figure}{
		\revision{
		Meta-splitting with using \paramsref[1] on \poisonD[n] of \cifar{10} against \blend~\citep{Chen2017blend}.}
	}
	\label{fig:resplit}
\end{minipage}
\end{figure}

\begin{figure*}[!t]
	\iftrue
	\hskip 21.8pt
	\begin{subfigure}{.217\textwidth}
		\includegraphics[width=\textwidth, trim=89.8 78 70 85, clip]{
			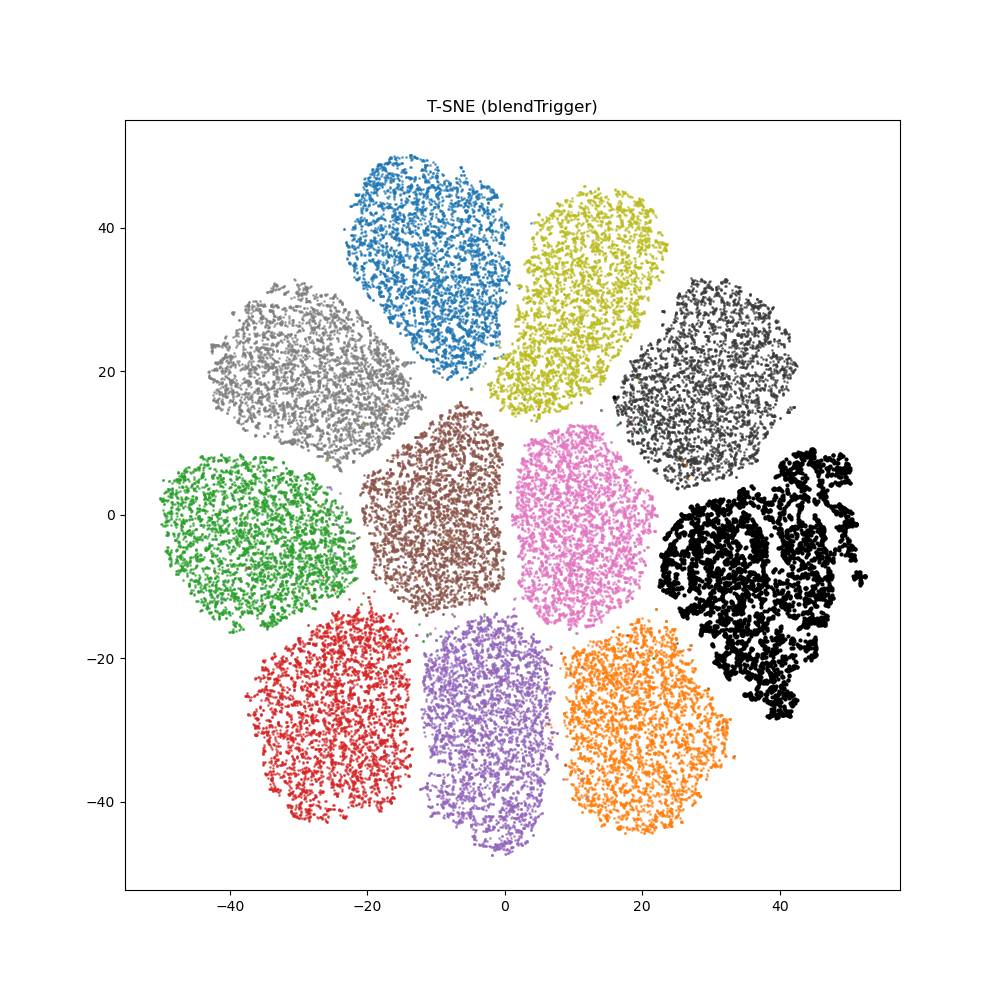
		}
	\end{subfigure}
	\hskip 11.3pt 
	\begin{subfigure}{.217\textwidth}
		\includegraphics[width=\textwidth, trim=89.8 78 70 85, clip]{
			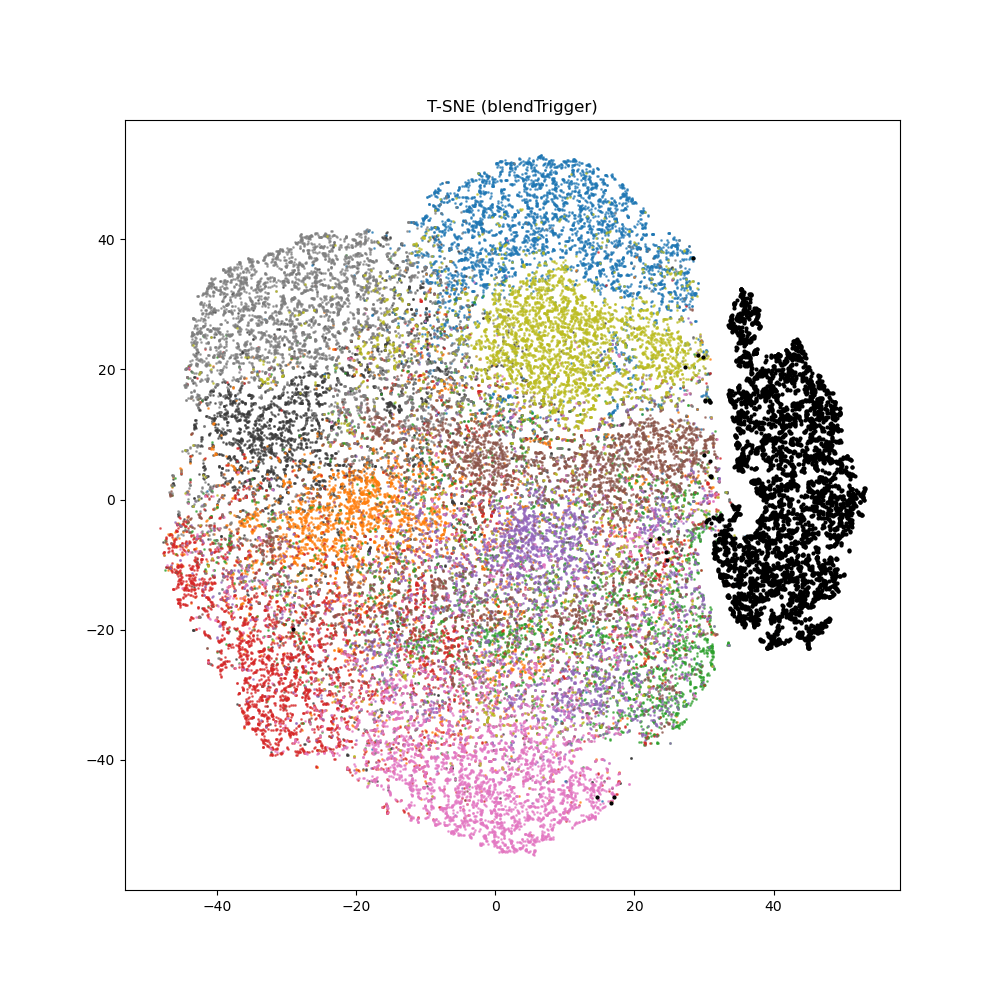
		}
	\end{subfigure}
	\hskip 11.5pt
	\begin{subfigure}{.217\textwidth}
		\includegraphics[width=\textwidth, trim=89.8 78 70 85, clip]{
			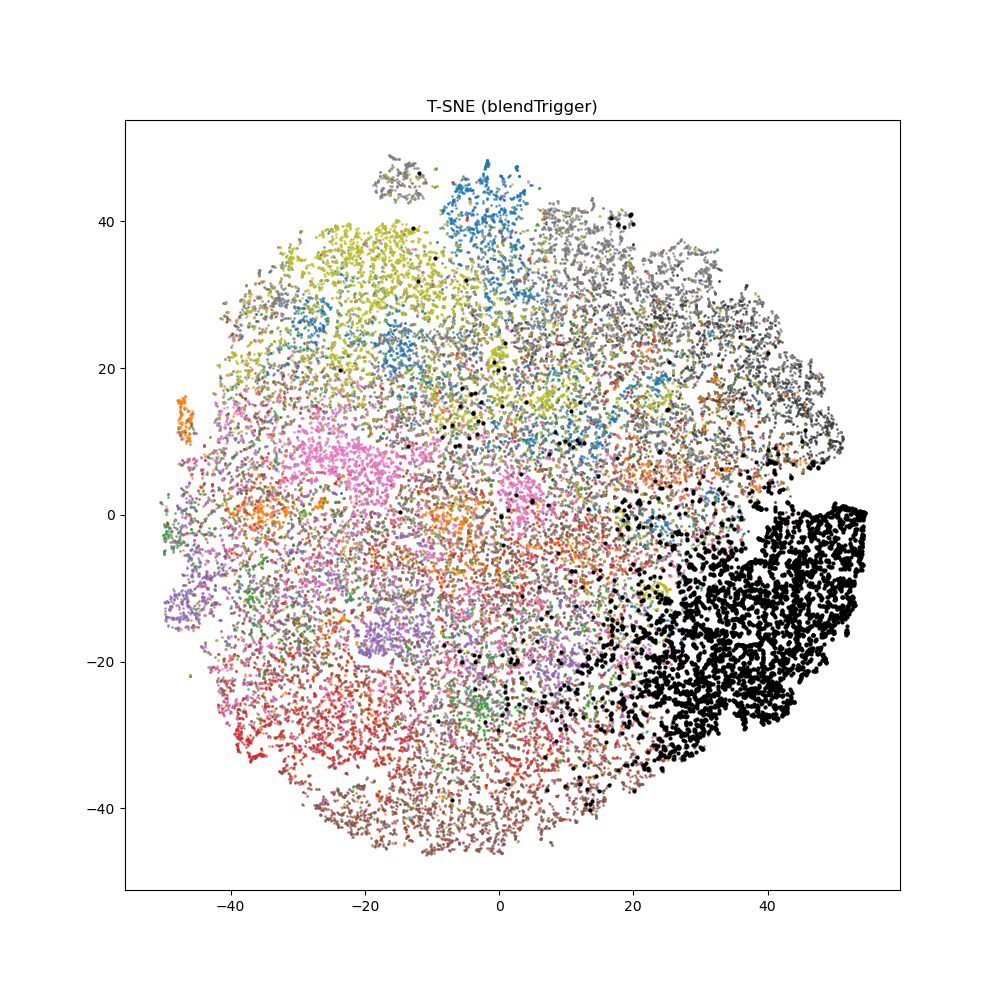
		}
	\end{subfigure}
	\hskip 11.7pt
	\begin{subfigure}{.217\textwidth}
		\includegraphics[width=\textwidth, trim=89.8 78 70 85, clip]{
			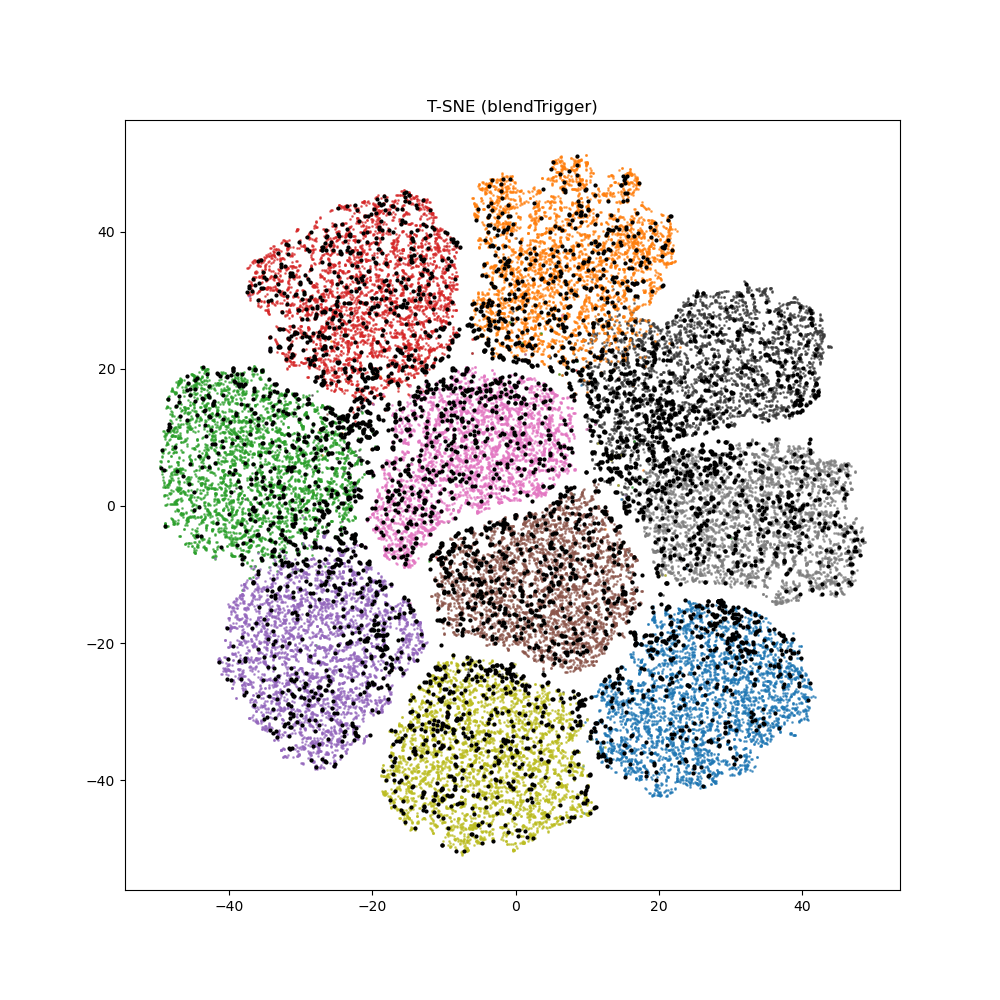
		}
	\end{subfigure}
	
	\smallskip
	
	\begin{subfigure}{.24\textwidth}
		\vskip -0.7pt
		\pgfplotstableread[col sep=space]{%
	bins	clean
	8.289306606457103e-06	38.567
	0.046059250831604004	1.311
	0.09211020916700363	0.692
	0.13816116750240326	0.455
	0.18421213328838348	0.333
	0.23026308417320251	0.253
	0.27631404995918274	0.225
	0.32236501574516296	0.168
	0.3684159815311432	0.151
	0.414466917514801	0.128
	0.46051788330078125	0.111
	0.5065688490867615	0.093
	0.5526198148727417	0.087
	0.5986707806587219	0.097
	0.6447217464447021	0.069
	0.6907727122306824	0.069
	0.7368236780166626	0.064
	0.782874584197998	0.064
	0.8289255499839783	0.052
	0.8749765157699585	0.048
	0.9210274815559387	0.054
	0.967078447341919	0.031
	1.0131293535232544	0.041
	1.0591803789138794	0.035
	1.1052312850952148	0.033
	1.1512823104858398	0.041
	1.1973332166671753	0.028
	1.2433842420578003	0.035
	1.2894351482391357	0.023
	1.3354861736297607	0.027
	1.3815370798110962	0.02
	1.4275881052017212	0.025
	1.4736390113830566	0.026
	1.519689917564392	0.031
	1.565740942955017	0.014
	1.6117918491363525	0.021
	1.6578428745269775	0.03
	1.703893780708313	0.025
	1.749944806098938	0.013
	1.7959957122802734	0.023
	1.8420467376708984	0.019
	1.8880976438522339	0.019
	1.9341486692428589	0.019
	1.9801995754241943	0.02
	2.0262506008148193	0.012
	2.0723013877868652	0.014
	2.1183524131774902	0.012
	2.1644034385681152	0.028
	2.2104544639587402	0.016
	2.256505250930786	0.019
	2.302556276321411	0.014
	2.348607301712036	0.016
	2.394658327102661	0.008
	2.440709114074707	0.017
	2.486760139465332	0.019
	2.532811164855957	0.013
	2.578861951828003	0.016
	2.624912977218628	0.012
	2.670964002609253	0.016
	2.717015027999878	0.014
	2.763065814971924	0.014
	2.809116840362549	0.016
	2.855167865753174	0.012
	2.901218891143799	0.009
	2.9472696781158447	0.014
	2.9933207035064697	0.002
	3.0393717288970947	0.014
	3.0854225158691406	0.004
	3.1314735412597656	0.006
	3.1775245666503906	0.018
	3.2235755920410156	0.008
	3.2696263790130615	0.016
	3.3156774044036865	0.013
	3.3617284297943115	0.011
	3.4077794551849365	0.012
	3.4538302421569824	0.008
	3.4998812675476074	0.006
	3.5459322929382324	0.014
	3.5919830799102783	0.013
	3.6380341053009033	0.017
	3.6840851306915283	0.004
	3.7301361560821533	0.013
	3.776186943054199	0.005
	3.822237968444824	0.012
	3.868288993835449	0.009
	3.914339780807495	0.004
	3.96039080619812	0.003
	4.006441593170166	0.004
	4.052492618560791	0.004
	4.098543643951416	0.014
	4.144594669342041	0.005
	4.190645694732666	0.012
	4.236696720123291	0.005
	4.282747745513916	0.008
	4.328798770904541	0.007
	4.374849319458008	0.007
	4.420900344848633	0.002
	4.466951370239258	0.013
	4.513002395629883	0.012
	4.559053421020508	0.008
	4.605104446411133	0.003
	4.651155471801758	0.007
	4.697206020355225	0.01
	4.74325704574585	0.006
	4.789308071136475	0.018
	4.8353590965271	0.008
	4.881410121917725	0.009
	4.92746114730835	0.003
	4.973512172698975	0.014
	5.019562721252441	0.005
	5.065613746643066	0.004
	5.111664772033691	0.014
	5.157715797424316	0.006
	5.203766822814941	0.006
	5.249817848205566	0.006
	5.295868873596191	0.005
	5.341919898986816	0.004
	5.387970447540283	0.006
	5.434021472930908	0.006
	5.480072498321533	0.006
	5.526123523712158	0.004
	5.572174549102783	0.004
	5.618225574493408	0.005
	5.664276599884033	0.001
	5.7103271484375	0.007
	5.756378173828125	0.008
	5.80242919921875	0.003
	5.848480224609375	0.004
	5.89453125	0.006
	5.940582275390625	0.002
	5.98663330078125	0.009
	6.032683849334717	0.001
	6.078734874725342	0.007
	6.124785900115967	0.002
	6.170836925506592	0.006
	6.216887950897217	0.006
	6.262938976287842	0.007
	6.308990001678467	0.003
	6.355040550231934	0.003
	6.401091575622559	0.004
	6.447142601013184	0.009
	6.493193626403809	0.01
	6.539244651794434	0.007
	6.585295677185059	0.007
	6.631346702575684	0.006
	6.677397727966309	0.003
	6.723448276519775	0.008
	6.7694993019104	0.004
	6.815550327301025	0.006
	6.86160135269165	0.006
	6.907652378082275	0.007
	6.9537034034729	0.005
	6.999754428863525	0.003
	7.045804977416992	0.008
	7.091856002807617	0.003
	7.137907028198242	0.003
	7.183958053588867	0.008
	7.230009078979492	0.005
	7.276060104370117	0.006
	7.322111129760742	0.007
	7.368161678314209	0.005
	7.414212703704834	0.006
	7.460263729095459	0.007
	7.506314754486084	0.009
	7.552365779876709	0.005
	7.598416805267334	0.003
	7.644467830657959	0.003
	7.690518856048584	0.007
	7.736569404602051	0.002
	7.782620429992676	0.003
	7.828671455383301	0.006
	7.874722480773926	0.004
	7.920773506164551	0.006
	7.966824531555176	0.006
	8.0128755569458	0.005
	8.058926582336426	0.007
	8.10497760772705	0.006
	8.151028633117676	0.006
	8.1970796585083	0.008
	8.24312973022461	0.007
	8.289180755615234	0.003
	8.33523178100586	0.002
	8.381282806396484	0.01
	8.42733383178711	0.012
	8.473384857177734	0.009
	8.51943588256836	0.004
	8.565486907958984	0.012
	8.61153793334961	0.008
	8.657588958740234	0.008
	8.70363998413086	0.007
	8.749691009521484	0.011
	8.79574203491211	0.01
	8.841793060302734	0.013
	8.887843132019043	0.009
	8.933894157409668	0.01
	8.979945182800293	0.018
	9.025996208190918	0.013
	9.072047233581543	0.012
	9.118098258972168	0.014
	9.164149284362793	0.031
}\cleantable

\pgfplotstableread[col sep=space]{%
	bins	poison
	8.289306606457103e-06	4.983
	0.046059250831604004	0.006
	0.09211020916700363	0.004
	0.13816116750240326	0.002
}\poisontable

\begin{tikzpicture}
	\begin{axis}[
		ybar,
		height=0.5\linewidth,
		width=0.9\linewidth,
		legend pos=outer north east,
		enlarge x limits=0.02,
		grid=major, 
		grid style={dashed,gray!30},
		xlabel={\rce Loss},
		ylabel={\# Samples $\times 10^3$},
		xmin=0,
		xmax=10,
		ymin=0,
		ymax=14.5,
		xtick={0,1,...,10},
		xtick align=inside,
		ytick={0, 2, 4, ..., 14},
		xticklabels={0,1,2,3,4,5,6,7,8,9,10},
		yticklabels={0, 2, 4, 6, 8, 10, 12, 14},
		yticklabel style = {font=\ticksize, yshift=0.0ex},
		xticklabel style = {font=\ticksize},
		ylabel style = {font=\scriptsize, yshift=-4.5ex},
		xlabel style = {font=\scriptsize, yshift=1ex},
		scale only axis,
		legend image post style={scale=0.8},
		legend style={at={(0.3,0.95)}, font=\legendsize, anchor=north 
			west, legend columns=1, fill=white, draw=white, nodes={scale=1.0, 
				transform shape}, column sep=1pt},
		legend cell align={left}
		]
		
		\addplot [ybar, bar width=1.5, draw=primarycolor, draw opacity=0, line width=0, fill=primarycolor, fill opacity=0.9, mark options={scale=0.5}] table[x index = 0, y index=1] {\cleantable};
		
	\end{axis}
	
	\begin{axis}[
		ybar,
		height=0.5\linewidth,
		width=0.9\linewidth,
		legend pos=outer north east,
		enlarge x limits=0.02,
		xlabel=\empty,
		ylabel=\empty,
		xmin=0,
		xmax=10,
		ymin=0,
		ymax=12.5,
		xtick=\empty,
		xtick align=inside,
		ytick=\empty,
		xticklabels=\empty,
		yticklabels=\empty,
		yticklabel style = {font=\ticksize, yshift=0.0ex},
		xticklabel style = {font=\ticksize},
		ylabel style = {font=\scriptsize, yshift=-4.5ex},
		xlabel style = {font=\scriptsize, yshift=1ex},
		scale only axis,
		legend image post style={scale=0.8},
		legend style={at={(0.3,0.82)}, font=\legendsize, anchor=north west, legend columns=1, fill=white, draw=white, nodes={scale=1.0, 		transform shape}, column sep=1pt}, 
		legend cell align={left}
		]
		
		\addplot[ybar, bar width=1.5, draw=secondarycolor, draw opacity=0, line width=0, fill=secondarycolor, fill opacity=0.8, mark options={scale=0.5}] table[x index = 0, y index=1] {\poisontable};
		
	\end{axis}
\end{tikzpicture}
		\captionsetup{margin*={20pt, 0pt}}
		\vskip -7.9pt
		\caption{Initial model \paramsnaive}
		\label{fig:tsne-cifar10-badnets-c0}
	\end{subfigure}
	\hskip 3.5pt
	\begin{subfigure}{.24\textwidth}
		\pgfplotstableread[col sep=space]{%
	bins	clean
	8.289306606457103e-06	14.385
	0.0460599884390831	1.107
	0.09211168438196182	0.677
	0.13816338777542114	0.484
	0.18421508371829987	0.378
	0.2302667796611786	0.303
	0.2763184905052185	0.248
	0.32237017154693604	0.243
	0.36842188239097595	0.191
	0.41447359323501587	0.184
	0.4605252742767334	0.173
	0.5065769553184509	0.166
	0.5526286959648132	0.137
	0.5986803770065308	0.131
	0.6447320580482483	0.114
	0.6907837986946106	0.121
	0.7368354797363281	0.088
	0.7828871607780457	0.117
	0.828938901424408	0.089
	0.8749905824661255	0.1
	0.921042263507843	0.083
	0.9670939445495605	0.066
	1.0131456851959229	0.095
	1.0591974258422852	0.058
	1.105249047279358	0.073
	1.1513007879257202	0.099
	1.197352409362793	0.079
	1.2434041500091553	0.063
	1.2894558906555176	0.075
	1.3355075120925903	0.069
	1.3815592527389526	0.06
	1.427610993385315	0.061
	1.4736626148223877	0.062
	1.51971435546875	0.066
	1.5657660961151123	0.048
	1.611817717552185	0.061
	1.6578694581985474	0.061
	1.7039211988449097	0.046
	1.7499728202819824	0.06
	1.7960245609283447	0.061
	1.842076301574707	0.054
	1.8881279230117798	0.046
	1.934179663658142	0.05
	1.9802314043045044	0.051
	2.026283025741577	0.055
	2.0723347663879395	0.043
	2.1183865070343018	0.05
	2.164438247680664	0.045
	2.2104897499084473	0.042
	2.2565414905548096	0.044
	2.302593231201172	0.055
	2.348644971847534	0.058
	2.3946967124938965	0.046
	2.440748453140259	0.044
	2.486799955368042	0.048
	2.5328516960144043	0.046
	2.5789034366607666	0.04
	2.624955177307129	0.031
	2.671006917953491	0.035
	2.7170586585998535	0.033
	2.7631101608276367	0.046
	2.809161901473999	0.035
	2.8552136421203613	0.031
	2.9012653827667236	0.039
	2.947317123413086	0.036
	2.993368625640869	0.044
	3.0394203662872314	0.037
	3.0854721069335938	0.052
	3.131523847579956	0.038
	3.1775755882263184	0.045
	3.2236273288726807	0.039
	3.269678831100464	0.033
	3.315730571746826	0.038
	3.3617823123931885	0.035
	3.407834053039551	0.037
	3.453885793685913	0.04
	3.4999375343322754	0.035
	3.5459890365600586	0.042
	3.592040777206421	0.029
	3.638092517852783	0.034
	3.6841442584991455	0.037
	3.730195999145508	0.031
	3.77624773979187	0.027
	3.8222992420196533	0.037
	3.8683509826660156	0.036
	3.914402723312378	0.03
	3.9604544639587402	0.035
	4.006505966186523	0.042
	4.052557945251465	0.04
	4.098609447479248	0.038
	4.1446614265441895	0.038
	4.190712928771973	0.035
	4.236764430999756	0.047
	4.282816410064697	0.039
	4.3288679122924805	0.037
	4.374919891357422	0.034
	4.420971393585205	0.037
	4.467022895812988	0.035
	4.51307487487793	0.049
	4.559126377105713	0.033
	4.605178356170654	0.032
	4.6512298583984375	0.044
	4.697281837463379	0.034
	4.743333339691162	0.042
	4.789384841918945	0.037
	4.835436820983887	0.038
	4.88148832321167	0.025
	4.927540302276611	0.042
	4.9735918045043945	0.034
	5.019643306732178	0.032
	5.065695285797119	0.041
	5.111746788024902	0.045
	5.157798767089844	0.04
	5.203850269317627	0.041
	5.24990177154541	0.036
	5.295953750610352	0.044
	5.342005252838135	0.042
	5.388057231903076	0.029
	5.434108734130859	0.036
	5.480160713195801	0.036
	5.526212215423584	0.036
	5.572263717651367	0.03
	5.618315696716309	0.038
	5.664367198944092	0.045
	5.710419178009033	0.029
	5.756470680236816	0.032
	5.8025221824646	0.046
	5.848574161529541	0.041
	5.894625663757324	0.039
	5.940677642822266	0.036
	5.986729145050049	0.034
	6.032780647277832	0.047
	6.078832626342773	0.035
	6.124884128570557	0.043
	6.170936107635498	0.042
	6.216987609863281	0.038
	6.263039588928223	0.035
	6.309091091156006	0.037
	6.355142593383789	0.051
	6.4011945724487305	0.052
	6.447246074676514	0.041
	6.493298053741455	0.04
	6.539349555969238	0.038
	6.5854010581970215	0.05
	6.631453037261963	0.036
	6.677504539489746	0.043
	6.7235565185546875	0.05
	6.769608020782471	0.04
	6.815659999847412	0.052
	6.861711502075195	0.053
	6.9077630043029785	0.05
	6.95381498336792	0.054
	6.999866485595703	0.043
	7.0459184646606445	0.057
	7.091969966888428	0.05
	7.138021469116211	0.05
	7.184073448181152	0.052
	7.2301249504089355	0.06
	7.276176929473877	0.059
	7.32222843170166	0.06
	7.368279933929443	0.051
	7.414331912994385	0.062
	7.460383415222168	0.058
	7.506435394287109	0.053
	7.552486896514893	0.051
	7.598538875579834	0.048
	7.644590377807617	0.065
	7.6906418800354	0.067
	7.736693859100342	0.068
	7.782745361328125	0.073
	7.828797340393066	0.076
	7.87484884262085	0.074
	7.920900344848633	0.07
	7.966952323913574	0.077
	8.013004302978516	0.093
	8.05905532836914	0.075
	8.105107307434082	0.079
	8.151159286499023	0.075
	8.197210311889648	0.116
	8.24326229095459	0.097
	8.289314270019531	0.09
	8.335366249084473	0.111
	8.381417274475098	0.128
	8.427469253540039	0.118
	8.47352123260498	0.12
	8.519572257995605	0.14
	8.565624237060547	0.146
	8.611676216125488	0.124
	8.65772819519043	0.168
	8.703779220581055	0.171
	8.749831199645996	0.194
	8.795883178710938	0.188
	8.841934204101562	0.246
	8.887986183166504	0.25
	8.934038162231445	0.307
	8.98008918762207	0.38
	9.026141166687012	0.511
	9.072193145751953	0.672
	9.118245124816895	1.173
	9.16429615020752	12.621
}\cleantable

\pgfplotstableread[col sep=space]{%
	bins	poison
	8.289306606457103e-06	4.956
	0.13816338777542114	0.005
	0.2302667796611786	0.002
	0.2763184905052185	0.002
	0.5986803770065308	0.002
	8.47352123260498	0.002
	8.887986183166504	0.002
	9.16429615020752	0.007
}\poisontable

\begin{tikzpicture}
	\begin{axis}[
		ybar,
		height=0.5\linewidth,
		width=0.9\linewidth,
		legend pos=outer north east,
		enlarge x limits=0.02,
		grid=major, 
		grid style={dashed,gray!30},
		xlabel={\rce Loss},
		ylabel=\empty,
		xmin=0,
		xmax=10,
		ymin=0,
		ymax=14.5,
		xtick={0,1,...,10},
		xtick align=inside,
		ytick={0, 2, 4, ..., 14},
		xticklabels={0,1,2,3,4,5,6,7,8,9,10},
		yticklabels={0, 2, 4, 6, 8, 10, 12, 14},
		yticklabel style = {font=\ticksize, yshift=0.0ex},
		xticklabel style = {font=\ticksize},
		ylabel style = {font=\scriptsize, yshift=-4.5ex},
		xlabel style = {font=\scriptsize, yshift=1ex},
		scale only axis,
		legend image post style={scale=0.8},
		legend style={at={(0.3,0.95)}, font=\legendsize, anchor=north 
			west, legend columns=1, fill=white, draw=white, nodes={scale=1.0, 
				transform shape}, column sep=1pt},
		legend cell align={left}
		]
		
		\addplot [ybar, bar width=1.5, draw=primarycolor, draw opacity=0, line width=0, fill=primarycolor, fill opacity=0.9, mark options={scale=0.5}] table[x index = 0, y index=1] {\cleantable};
		
	\end{axis}
	
	\begin{axis}[
		ybar,
		height=0.5\linewidth,
		width=0.9\linewidth,
		legend pos=outer north east,
		enlarge x limits=0.02,
		xlabel=\empty,
		ylabel=\empty,
		xmin=0,
		xmax=10,
		ymin=0,
		ymax=14.5,
		xtick=\empty,
		xtick align=inside,
		ytick=\empty,
		xticklabels=\empty,
		yticklabels=\empty,
		yticklabel style = {font=\ticksize, yshift=0.0ex},
		xticklabel style = {font=\ticksize},
		ylabel style = {font=\scriptsize, yshift=-4.5ex},
		xlabel style = {font=\scriptsize, yshift=1ex},
		scale only axis,
		legend image post style={scale=0.8},
		legend style={at={(0.3,0.82)}, font=\legendsize, anchor=north west, legend columns=1, fill=white, draw=white, nodes={scale=1.0, 		transform shape}, column sep=1pt}, 
		legend cell align={left}
		]
		
		\addplot[ybar, bar width=1.5, draw=secondarycolor, draw opacity=0, line width=0, fill=secondarycolor, fill opacity=0.8, mark options={scale=0.5}] table[x index = 0, y index=1] {\poisontable};
		
	\end{axis}
	
	\begin{axis}[
		height=0.5\linewidth,
		width=0.9\linewidth,
		legend pos=outer north east,
		enlarge x limits=0.02,
		xlabel=\empty,
		ylabel=\empty,
		xmin=0,
		xmax=10,
		ymin=0,
		ymax=10,
		xtick=\empty,
		xtick align=inside,
		ytick=\empty,
		xticklabels=\empty,
		yticklabels=\empty,
		yticklabel style = {font=\ticksize, yshift=0.0ex},
		xticklabel style = {font=\ticksize},
		ylabel style = {font=\scriptsize, yshift=-4.5ex},
		xlabel style = {font=\scriptsize, yshift=1ex},
		scale only axis,
		legend image post style={scale=0.37},
		legend style={at={(0.3,0.69)}, font=\legendsize, anchor=north west, legend columns=1, fill=white, draw=white, nodes={scale=1.0, 		transform shape}, column sep=1pt},
		legend cell align={left}
		]
		
		\addplot[draw=black, densely dashed, line width=1pt] %
		coordinates {
			(4.6052, 0.0)
			(4.6052, 14.5)
		};
		
		\node[
		at={(25, 35)},
		anchor=south,
		] {\scriptsize\poisonD[1]}
		;
		
		\draw [
		-stealth,
		black,
		very thick,
		] (4, 28) to [out=70, in=-170] (15, 45);
		
		\node[
		at={(67, 34)},
		anchor=south,
		] {\scriptsize\cleanD[1]}
		;
		
		\draw [
		-stealth,
		black,
		very thick,
		] (88, 28) to [out=110, in=-20] (77, 45);
		
	\end{axis}
\end{tikzpicture}
		\captionsetup{margin*={10pt, 0pt}}
		\vskip -10pt
		\caption{\nth{1} Reference model $\paramsref[1]$}
		\label{fig:tsne-cifar10-badnets-c1}
	\end{subfigure}
	\begin{subfigure}{.24\textwidth}
		\pgfplotstableread[col sep=space]{%
	bins	clean
	8.289306606457103e-06	4.549
	0.0460599884390831	0.022
	0.09211168438196182	0.006
	0.13816338777542114	0.004
	0.18421508371829987	0.008
	0.2302667796611786	0.009
	0.2763184905052185	0.004
	0.32237017154693604	0.002
	0.36842188239097595	0.004
	0.41447359323501587	0.003
	0.4605252742767334	0.004
	0.5065769553184509	0.002
	0.5526286959648132	0.002
	0.5986803770065308	0.004
	0.6447320580482483	0.0
	0.6907837986946106	0.001
	0.7368354797363281	0.003
	0.7828871607780457	0.002
	0.828938901424408	0.002
	0.8749905824661255	0.001
	0.921042263507843	0.001
	0.9670939445495605	0.001
	1.0131456851959229	0.002
	1.0591974258422852	0.002
	1.105249047279358	0.004
	1.1513007879257202	0.002
	1.197352409362793	0.003
	1.2434041500091553	0.0
	1.2894558906555176	0.005
	1.3355075120925903	0.001
	1.3815592527389526	0.0
	1.427610993385315	0.003
	1.4736626148223877	0.002
	1.51971435546875	0.003
	1.5657660961151123	0.0
	1.611817717552185	0.002
	1.6578694581985474	0.0
	1.7039211988449097	0.0
	1.7499728202819824	0.001
	1.7960245609283447	0.001
	1.842076301574707	0.003
	1.8881279230117798	0.0
	1.934179663658142	0.001
	1.9802314043045044	0.001
	2.026283025741577	0.0
	2.0723347663879395	0.001
	2.1183865070343018	0.002
	2.164438247680664	0.0
	2.2104897499084473	0.0
	2.2565414905548096	0.001
	2.302593231201172	0.002
	2.348644971847534	0.0
	2.3946967124938965	0.001
	2.440748453140259	0.002
	2.486799955368042	0.002
	2.5328516960144043	0.001
	2.5789034366607666	0.0
	2.624955177307129	0.002
	2.671006917953491	0.0
	2.7170586585998535	0.0
	2.7631101608276367	0.003
	2.809161901473999	0.002
	2.8552136421203613	0.001
	2.9012653827667236	0.0
	2.947317123413086	0.001
	2.993368625640869	0.001
	3.0394203662872314	0.001
	3.0854721069335938	0.0
	3.131523847579956	0.0
	3.1775755882263184	0.0
	3.2236273288726807	0.001
	3.269678831100464	0.001
	3.315730571746826	0.0
	3.3617823123931885	0.0
	3.407834053039551	0.001
	3.453885793685913	0.0
	3.4999375343322754	0.001
	3.5459890365600586	0.001
	3.592040777206421	0.001
	3.638092517852783	0.0
	3.6841442584991455	0.0
	3.730195999145508	0.001
	3.77624773979187	0.002
	3.8222992420196533	0.0
	3.8683509826660156	0.002
	3.914402723312378	0.001
	3.9604544639587402	0.004
	4.006505966186523	0.001
	4.052557945251465	0.0
	4.098609447479248	0.001
	4.1446614265441895	0.001
	4.190712928771973	0.0
	4.236764430999756	0.0
	4.282816410064697	0.001
	4.3288679122924805	0.002
	4.374919891357422	0.0
	4.420971393585205	0.001
	4.467022895812988	0.0
	4.51307487487793	0.0
	4.559126377105713	0.002
	4.605178356170654	0.001
	4.6512298583984375	0.0
	4.697281837463379	0.001
	4.743333339691162	0.0
	4.789384841918945	0.0
	4.835436820983887	0.0
	4.88148832321167	0.001
	4.927540302276611	0.002
	4.9735918045043945	0.001
	5.019643306732178	0.001
	5.065695285797119	0.0
	5.111746788024902	0.0
	5.157798767089844	0.002
	5.203850269317627	0.002
	5.24990177154541	0.002
	5.295953750610352	0.002
	5.342005252838135	0.0
	5.388057231903076	0.001
	5.434108734130859	0.0
	5.480160713195801	0.001
	5.526212215423584	0.0
	5.572263717651367	0.002
	5.618315696716309	0.001
	5.664367198944092	0.003
	5.710419178009033	0.0
	5.756470680236816	0.0
	5.8025221824646	0.0
	5.848574161529541	0.001
	5.894625663757324	0.001
	5.940677642822266	0.0
	5.986729145050049	0.001
	6.032780647277832	0.002
	6.078832626342773	0.002
	6.124884128570557	0.001
	6.170936107635498	0.0
	6.216987609863281	0.0
	6.263039588928223	0.0
	6.309091091156006	0.001
	6.355142593383789	0.0
	6.4011945724487305	0.001
	6.447246074676514	0.002
	6.493298053741455	0.001
	6.539349555969238	0.0
	6.5854010581970215	0.001
	6.631453037261963	0.0
	6.677504539489746	0.0
	6.7235565185546875	0.0
	6.769608020782471	0.001
	6.815659999847412	0.0
	6.861711502075195	0.001
	6.9077630043029785	0.002
	6.95381498336792	0.001
	6.999866485595703	0.001
	7.0459184646606445	0.0
	7.091969966888428	0.001
	7.138021469116211	0.0
	7.184073448181152	0.002
	7.2301249504089355	0.0
	7.276176929473877	0.001
	7.32222843170166	0.002
	7.368279933929443	0.001
	7.414331912994385	0.001
	7.460383415222168	0.001
	7.506435394287109	0.001
	7.552486896514893	0.001
	7.598538875579834	0.0
	7.644590377807617	0.001
	7.6906418800354	0.001
	7.736693859100342	0.0
	7.782745361328125	0.0
	7.828797340393066	0.004
	7.87484884262085	0.004
	7.920900344848633	0.0
	7.966952323913574	0.002
	8.013004302978516	0.001
	8.05905532836914	0.001
	8.105107307434082	0.001
	8.151159286499023	0.001
	8.197210311889648	0.002
	8.24326229095459	0.004
	8.289314270019531	0.002
	8.335366249084473	0.004
	8.381417274475098	0.002
	8.427469253540039	0.004
	8.47352123260498	0.004
	8.519572257995605	0.003
	8.565624237060547	0.002
	8.611676216125488	0.002
	8.65772819519043	0.002
	8.703779220581055	0.003
	8.749831199645996	0.005
	8.795883178710938	0.011
	8.841934204101562	0.009
	8.887986183166504	0.011
	8.934038162231445	0.01
	8.98008918762207	0.017
	9.026141166687012	0.009
	9.072193145751953	0.02
	9.118245124816895	0.042
	9.16429615020752	40.044
}\cleantable

\pgfplotstableread[col sep=space]{%
	bins	poison
	8.289306606457103e-06	5.0
}\poisontable

\begin{tikzpicture}
	
	\begin{axis}[
		ybar,
		height=0.5\linewidth,
		width=0.9\linewidth,
		legend pos=outer north east,
		enlarge x limits=0.02,
		grid=major, 
		grid style={dashed,gray!30},
		xlabel={\rce Loss},
		ylabel=\empty, 
		xmin=0,
		xmax=10,
		ymin=0,
		ymax=14.5,
		xtick={0,1,...,10},
		xtick align=inside,
		ytick={0, 2, 4, ..., 14},
		xticklabels={0,1,2,3,4,5,6,7,8,9,10},
		yticklabels={0, 2, 4, 6, 8, 10, 12, 14},
		yticklabel style = {font=\ticksize, yshift=0.0ex},
		xticklabel style = {font=\ticksize},
		ylabel style = {font=\scriptsize, yshift=-4.5ex},
		xlabel style = {font=\scriptsize, yshift=1ex},
		scale only axis,
		legend image post style={scale=0.8},
		legend style={at={(0.25,0.96)}, font=\legendsize, anchor=north 
			west, legend columns=1, fill=white, draw=white, nodes={scale=1.0, 
				transform shape}, column sep=1pt},
		legend cell align={left}
		]
		
		\addplot [ybar, bar width=1.5, draw=primarycolor, draw opacity=0, line width=0, fill=primarycolor, fill opacity=0.9, mark options={scale=0.5}] table[x index = 0, y index=1] {\cleantable};
		
	\end{axis}
	
	\begin{axis}[
		ybar,
		height=0.5\linewidth,
		width=0.9\linewidth,
		legend pos=outer north east,
		enlarge x limits=0.02,
		xlabel=\empty,
		ylabel=\empty,
		xmin=0,
		xmax=10,
		ymin=0,
		ymax=14.5,
		xtick=\empty,
		xtick align=inside,
		ytick=\empty,
		xticklabels=\empty,
		yticklabels=\empty,
		yticklabel style = {font=\ticksize, yshift=0.0ex},
		xticklabel style = {font=\ticksize},
		ylabel style = {font=\scriptsize, yshift=-4.5ex},
		xlabel style = {font=\scriptsize, yshift=1ex},
		scale only axis,
		legend image post style={scale=0.8},
		legend style={at={(0.25,0.8)}, font=\legendsize, anchor=north west, legend columns=1, fill=white, draw=white, nodes={scale=1.0, 		transform shape}, column sep=1pt}, 
		legend cell align={left}
		]
		
		\addplot[ybar, bar width=1.5, draw=secondarycolor, draw opacity=0, line width=0, fill=secondarycolor, fill opacity=0.8, mark options={scale=0.5}] table[x index = 0, y index=1] {\poisontable};
		
	\end{axis}
	
	\begin{axis}[
		height=0.5\linewidth,
		width=0.9\linewidth,
		legend pos=outer north east,
		enlarge x limits=0.02,
		xlabel=\empty,
		ylabel=\empty,
		xmin=0,
		xmax=10,
		ymin=0,
		ymax=10.5,
		xtick=\empty,
		xtick align=inside,
		ytick=\empty,
		xticklabels=\empty,
		yticklabels=\empty,
		yticklabel style = {font=\ticksize, yshift=0.0ex},
		xticklabel style = {font=\ticksize},
		ylabel style = {font=\scriptsize, yshift=-4.5ex},
		xlabel style = {font=\scriptsize, yshift=1ex},
		scale only axis,
		legend image post style={scale=0.37},
		legend style={at={(0.25,0.65)}, font=\legendsize, anchor=north west, legend columns=1, fill=white, draw=white, nodes={scale=1.0, 		transform shape}, column sep=1pt},
		legend cell align={left}
		]
		
		\addplot[draw=black, densely dashed, line width=1pt] %
		coordinates {
			(4.6052, 0.0)
			(4.6052, 14.5)
		};
		
		\node[
		at={(25, 35)},
		anchor=south,
		] {\scriptsize\poisonD[n]}
		;
		
		\draw [
		-stealth,
		black,
		very thick,
		] (4, 28) to [out=70, in=-170] (15, 45);
		
		\node[
		at={(67, 34)},
		anchor=south,
		] {\scriptsize\cleanD[n]}
		;
		
		\draw [
		-stealth,
		black,
		very thick,
		] (88, 28) to [out=110, in=-20] (77, 45);
		
	\end{axis}
	
%
%
\end{tikzpicture}
		\captionsetup{margin*={10pt, 0pt}}
		\vskip -10pt
		\caption{$n^{\textnormal{th}}$ Reference model $\paramsref[n]$}
		\label{fig:tsne-cifar10-badnets-end}
	\end{subfigure}
	\begin{subfigure}{.24\textwidth}
		\vskip -1.1pt
		\input{
			figures/cifar10-resnet18-blend-ref-c20-main.tex
		}
		\captionsetup{margin*={15pt, 0pt}}
		\vskip -8pt
		\caption{Final clean model \params}
		\label{fig:tsne-cifar10-badnets-main}
	\end{subfigure}
	\else
	\label{fig:tsne-cifar10-badnets-c0}
	\label{fig:tsne-cifar10-badnets-c1}
	\label{fig:tsne-cifar10-badnets-end}
	\label{fig:tsne-cifar10-badnets-main}
	\vspace*{87mm}
	\fi
	\caption{
		Dataset distribution of the latent space by \tsne (top row) and \rce loss (bottom row) for \resnet{18} on \cifar{10} poisoned by \blend attack. Poisonous samples are marked black, and other colors represent benign samples of different classes. 
	}
	\label{fig:tSNE}
\end{figure*}

\subsection{Meta-Splitting (\stage{3})} 
\label{sec:stage-3}
As the last reference model \paramsref[\Niters] is strongly biased toward the backdoor, most of benign samples of the backdoor's target class remain in \poisonD[\Niters]. Hence, directly training on \cleanD[\Niters] right after \stage{2} results in low natural performance (\cf~\cref{fig:no-meta-split}). 
%
%
The first reference model \paramsref[1], in turn, yields two clusters in latent space (\cf~\cref{fig:tsne-cifar10-badnets-c1}), which can separate poisonous and benign samples, especially from the target class. 
Consequently, we use \paramsref[1] to split \poisonD[\Niters] (\cf~\cref{fig:tsne-cifar10-badnets-end}) in two parts, \MSclean and \MSpoison, according to high and low \rce loss for benign and poisonous samples, respectively (\cf~\cref{fig:resplit}).
%
 


\subsection{Final Training (\stage{4})} 
\label{sec:stage-4}

For the final training, we combine the isolated benign subsets, \cleanD[n] (\stage{2}) and \MSclean (\stage{3}), yielding the final clean dataset~\cleanD, and reject the rest.
We then directly apply supervised learning using \ce~loss on this data to obtain the final clean model without the backdoor:
\begin{equation*}
	\argmin_\params
	~
	\sum_{\left( \inputx, \labely \right) \in \cleanD} 
	\lossce \left( \inputx, \labely, \params \right)
\end{equation*}

Analyzing the model in latent space (\cf~\cref{fig:tsne-cifar10-badnets-main}) shows that it forms clearly separable groups for classes and classifies poisonous samples in the original class when being applied to the original, poisoned dataset~\poisonset.

\section{Evaluation}
\label{sec:eval}
\label{sec:exp-setup}

We conduct extensive experiments across datasets and model architectures to evaluate our method in comparison to other training-time defenses.
Specifically, we evaluate on the small-scale dataset \cifar{10}~\citep{CIFAR} with  \wrn~\citep{Zagoruyko2016Wide} and \resnet{18}~\citep{He2016Resnet}, and the large-scale dataset \tinyimagenet~\citep{Le2015TinyImageNet} with \resnet{34}~\citep{He2016Resnet}.
\if\appendixref1
In~\cref{app:expm-gtsrb},
\else
In the appendix,
\fi
we also provide results for a second small-scale dataset, \gtsrb~\citep{Stall2022GTSRB}.

Below, we elaborate on the experimental setup wrt. the considered attacks, the related work to compare to and the evaluation metrics used, before we present an overview of \ourmethod's performance in \cref{sec:main-compare}.
In \cref{sec:split-efficient,sec:ablation-study}, we then analyze the splitting performance of our method and perform ablation experiments, respectively.
\revision{
\if\appendixref1
Moreover, we provide the detailed experimental setup in~\cref{app:expm-details}, additional ablation studies in~\cref{app:ablation-study-settings}, and the evaluation with an adaptive attack in~\cref{app:adaptive-attack}.
\else
Moreover, we provide the detailed experimental setup, additional ablation studies and the evaluation with an adaptive attack in the appendix.
\fi
}

\paragraph{Considered attacks}
We evaluate with six poisoning attacks, including
\badnets~\citep{Gu2017badnet}, 
\trojan~\citep{Liu2018trojan}, 
\blend attack~\citep{Chen2017blend}, 
\clbs~\citep[\clb;][]{turner2019labelconsistent},
\iab~\citep{Nguyen2020IAB}
and \wanet~\citep{Nguyen2021wanet}. 
We choose class~\num{0} as target, $\labelyt=0$, and use a poisoning rate of \perc{10}, $\rho=0.1$, except for \clb, where we poison \perc{50} and \perc{100} of the samples of the target class for the small-scale and large-scale dataset, respectively. 

\paragraph{Considered defenses}
We compare \ourmethod to four other defenses that also do \emph{not} require a clean reference dataset: 
\abl~\citep{Li2021antibackdoor}, 
\dbd~\citep{Huang2022backdoor}, 
\dst~\citep{Chen2022DST}, and 
\cbd~\citep{Zhang2023backdoor} with the parameters proposed in the corresponding publications. 
%
%
%
For \ourmethod, we set $\epsilon=10^{-5}$, so that the splitting threshold is $\sfrac{\RCEconst}{2} \approx 4.6052$. 
We use $\EpochsInit = 20$ epochs for the pretraining and $\EpochsPoi=10$ epochs for each round/iteration in \stage{2}.
Finally, we train the clean model for $\EpochsCln = 100$ epochs. 

\paragraph{Evaluation Metrics}
We present the defensive performance with two metrics: \accs (\acc) and \asrs (\asr). 
An optimal backdooring attack has a high \acc and an $\text{\asr}~\text{of}~\perc{100}$. In contrast, an effective backdoor defense should achieve high \acc while the attack is ineffective, $\text{\asr} \approx \perc{0}$.


\begin{table*}[!t]\vspace*{-1mm}
	\caption{Comparison of \ourmethod with prior defenses.
		All results are shown in \perc{}. The best results across all defenses are highlighted in \textbf{bold} font. Settings where the backdoor injection succeeds (\asr > \perc{90}) are marked in {\HLFail orange bold} font.\\[-5mm]}
	\label{tab:compare-sota}
	\tablesize
	\iftrue
		\hspace*{-11.5pt}
		\begin{tikzpicture}
		\newcommand{\boxy}{+1.74}
		\draw [fill=tabbgcolor,draw=tabbgcolor] (-6.117,\boxy) rectangle ($(8.969,\boxy)-(0,0.785)$);
		\renewcommand{\boxy}{-1.13}
		\draw [fill=tabbgcolor,draw=tabbgcolor] (-6.117,\boxy) rectangle ($(8.969,\boxy)-(0,0.785)$);
		\renewcommand{\boxy}{-4.00}
		\draw [fill=tabbgcolor,draw=tabbgcolor] (-6.117,\boxy) rectangle ($(8.969,\boxy)-(0,0.785)$);
		\node (table) {\newcommand{\mycsvreader}[3]{%
	\csvreader[
	head to column names,
	filter = \equal{\dataset}{#1} \and \equal{\arch}{#2} \and 
	\equal{\attack}{#3},
	late after line=\\,
	]%
	{res/main_results.csv}%
	{}
	{
		& \noacctxt
		& \noasrtxt
		& \ablacctxt
		& \ablasrtxt
		& \dbdacctxt
		& \dbdasrtxt
		& \dstacctxt
		& \dstasrtxt
		& \cbdacctxt
		& \cbdasrtxt
		& \oursacctxt
		& \oursasrtxt
	}
}


\begin{overalltable}{\textwidth}
	\multirow{16}{*}{\rotatebox{90}{\cifar{10}}}
	& \multirow{8}{*}{\wrn}
	& \badnets	\mycsvreader{cifar10}{wrn}{badnets}
	&
	& \trojan	\mycsvreader{cifar10}{wrn}{trojan}
	&
	& \blend	\mycsvreader{cifar10}{wrn}{blend}
	&
	& \clb		\mycsvreader{cifar10}{wrn}{clb}
	&
	& \iab		\mycsvreader{cifar10}{wrn}{iab}
	&
	& \wanet	\mycsvreader{cifar10}{wrn}{wanet}
	\cmidrule{3-15}
	& 
	& \avg	\mycsvreader{cifar10}{wrn}{avg}
	& 
	& \worst	\mycsvreader{cifar10}{wrn}{worstcase}
	\cmidrule{2-15}
	& \multirow{8}{*}{\resnet{18}}
	& \badnets	\mycsvreader{cifar10}{resnet18}{badnets}
	&
	& \trojan	\mycsvreader{cifar10}{resnet18}{trojan}
	&
	& \blend	\mycsvreader{cifar10}{resnet18}{blend}
	&
	& \clb		\mycsvreader{cifar10}{resnet18}{clb}
	&
	& \iab		\mycsvreader{cifar10}{resnet18}{iab}
	&
	& \wanet	\mycsvreader{cifar10}{resnet18}{wanet}
	\cmidrule{3-15}
	&
	& \avg		\mycsvreader{cifar10}{resnet18}{avg}
	&
	& \worst	\mycsvreader{cifar10}{resnet18}{worstcase}
	\midrule
	\multirow{8}{*}{\rotatebox{90}{\tinyimagenet}}
	& \multirow{8}{*}{\resnet{34}}
	& \badnets	\mycsvreader{tinyimagenet}{resnet34}{badnets}
	&
	& \trojan	\mycsvreader{tinyimagenet}{resnet34}{trojan}
	&
	& \blend	\mycsvreader{tinyimagenet}{resnet34}{blend}
	&
	& \clb		\mycsvreader{tinyimagenet}{resnet34}{clb}
	&
	& \iab		\mycsvreader{tinyimagenet}{resnet34}{iab}
	&
	& \wanet	\mycsvreader{tinyimagenet}{resnet34}{wanet}
	\cmidrule{3-15}
	&
	& \avg		\mycsvreader{tinyimagenet}{resnet34}{avg}
	&
	& \worst	\mycsvreader{tinyimagenet}{resnet34}{worstcase}
	\bottomrule
\end{overalltable}
};
		\end{tikzpicture}
	\else
		
	\fi
\end{table*}

\subsection{Defensive Capability of \ourmethod}
\label{sec:main-compare}

We provide the overall comparison of our method with other training-time defenses in \cref{tab:compare-sota}, where we highlight the highest accuracy (\acc) and the lowest attack success rate (\asr).
Additionally, we mark all settings where the backdoor cannot be mitigated (\asr > \perc{90}) in orange~color.

\paragraph{\cifar{10}} 
All defenses are robust against patch-based attacks (\eg \badnets) with \resnet{18}, while their performance decreases for stealthier attacks, such as \blend, \clb and \wanet.
In particular, \dbd and \dst are affected and also with \wrn results are worse due to its low model capacity.
\abl and \cbd, in turn, prevent various backdoors successfully but at the cost of natural accuracy.
In contrast, our method is effective in both model architectures, also maintaining the natural accuracy.
\ourmethod reduces the attack success rate below \perc{2} \emph{in the worst case} and preserves the highest natural performance of all defenses.

\paragraph{\tinyimagenet}
Also on \tinyimagenet, our method excels with on par natural accuracy and a reduction of the attack success rate to merely \perc{0.48} in the worst case.
\dst performs well in preserving the natural accuracy but mis-splits the dataset specifically for \blend attacks.
\dbd, in turn fails completely due to the difficulty of learning benign features as the loss distribution of benign and poisonous samples is more interleaved.
\abl and \cbd ensure a better defense but sacrifice natural accuracy decisively.

\paragraph{Summary}
Related approaches cannot defend against backdoors reliably, whereas our method can.
The reason is that we learn an easier task: learning the backdoor rather than the benign, natural task.
\ourmethod yields a natural accuracy on par with the ``\nodefense'' setting and suppresses each backdoor to the very minimum. 


\subsection{Performance in Dataset Splitting}
\label{sec:split-efficient}

Next, we analyze the performance of the methods based on dataset splitting (\dbd, \dst, and \ourmethod) by measuring the \Fone score and the rate of remaining poisonous samples. 
\cref{tab:comp_split} summarizes the splitting results.

For \dst, we merge ``poisoned'' and ''suspicious'' subsets for the evaluation as \dst only learns on it ``clean'' subset \citep{Chen2022DST}.
The method falls behind for the smaller \wrn model in particular as it cannot capture the benign task completely during splitting.
For larger models (\resnet{18} for \cifar{10} and \resnet{34} for \tinyimagenet), \dst leaves over less poisonous samples, except for \wanet and \blend. 
Moreover, \dbd's shortcomings on \tinyimagenet are also visible in the splitting performance, where more than \perc{5} of the final dataset are still poisonous (in light of an overall poisoning rate of \perc{10}).
Also for \resnet{18} on \cifar{10}, the recall is significantly lower than \perc{90} for \trojan, \blend, and \clb attacks. 
%
%

\ourmethod is largely independent of the model's capacity as we learn the backdoor rather than the benign task for the reference model.
For \cifar{10} with \wrn, we yield \Fone scores over \perc{66}, which is \num{49} percentage points higher than the best related work. 
For \tinyimagenet, our method even reaches \Fone $> \perc{95}$ on average and a $\pratio_{bng}<\perc{0.1}$. 

\begin{table*}[!t]\vspace*{-1mm}
	\caption{Comparing \ourmethod's dataset splitting with \dbd and \dst. Precision (\precision) measures the ratio of poisonous samples in \poisonD. \recall shows the isolation percentage of poisonous samples. $\pratio_{bng}$ is the poisoning ratio in \cleanD. All results are shown in \perc{}. We highlight the best \Fone score and $\pratio_{bng}$ of all defenses in \textbf{bold} font. $\pratio_{bng}$ above \perc{5} is marked in {\HLFail orange bold} font.\\[-5mm]
	}
	\label{tab:comp_split}
	\centering
	\iftrue
	\hspace*{-11.5pt}
	\begin{tikzpicture}
		\newcommand{\boxy}{+1.74}
		\draw [fill=tabbgcolor,draw=tabbgcolor] (-6.155,\boxy) rectangle ($(8.955,\boxy)-(0,0.785)$);
		\renewcommand{\boxy}{-1.13}
		\draw [fill=tabbgcolor,draw=tabbgcolor] (-6.155,\boxy) rectangle ($(8.955,\boxy)-(0,0.785)$);
		\renewcommand{\boxy}{-4.00}
		\draw [fill=tabbgcolor,draw=tabbgcolor] (-6.155,\boxy) rectangle ($(8.955,\boxy)-(0,0.785)$);
		\node (table) {\newcommand{\splitreader}[4][white]{%
	\csvreader[
	head to column names,
	filter = \equal{\dataset}{#2} \and \equal{\arch}{#3} \and 
	\equal{\attack}{#4},
	late after line=\\,
	]%
	{res/compare_split.csv}%
	{}
	{
		& \dbdPrec
		& \dbdRecall
		& \dbdFone
		& \dbdPrate
		& \dstPrec
		& \dstRecall
		& \dstFone
		& \dstPrate
		& \oursPrec
		& \oursRecall
		& \oursFone
		& \oursPrate
	}
}

\begin{splittable}{\textwidth}
	\multirow{16}{*}{\rotatebox{90}{\cifar{10}}}
	& \multirow{8}{*}{\wrn}
	& \badnets	\splitreader{cifar10}{wrn}{badnets}
	&
	& \trojan	\splitreader{cifar10}{wrn}{trojan}
	&
	& \blend	\splitreader{cifar10}{wrn}{blend}
	&
	& \clb		\splitreader{cifar10}{wrn}{clb}
	&
	& \iab		\splitreader{cifar10}{wrn}{iab}
	&
	& \wanet	\splitreader{cifar10}{wrn}{wanet}
	\cmidrule{3-15}
	&
	& \avg
	\splitreader[tabbgcolor]{cifar10}{wrn}{avg}
	&
	& \worst
	\splitreader[tabbgcolor]{cifar10}{wrn}{worstcase}
	\cmidrule{2-15}
	& \multirow{8}{*}{\resnet{18}}
	& \badnets	\splitreader{cifar10}{resnet18}{badnets}
	&
	& \trojan	\splitreader{cifar10}{resnet18}{trojan}
	&
	& \blend	\splitreader{cifar10}{resnet18}{blend}
	&
	& \clb		\splitreader{cifar10}{resnet18}{clb}
	&
	& \iab		\splitreader{cifar10}{resnet18}{iab}
	&
	& \wanet	\splitreader{cifar10}{resnet18}{wanet}
	\cmidrule{3-15}
	&
	& \avg
	\splitreader[tabbgcolor]{cifar10}{resnet18}{avg}
	&
	& \worst
	\splitreader[tabbgcolor]{cifar10}{resnet18}{worstcase}
	\midrule
	\multirow{8}{*}{\rotatebox{90}{\tinyimagenet}}
	& \multirow{8}{*}{\resnet{34}}
	& \badnets	\splitreader{tinyimagenet}{resnet34}{badnets}
	&
	& \trojan	\splitreader{tinyimagenet}{resnet34}{trojan}
	&
	& \blend	\splitreader{tinyimagenet}{resnet34}{blend}
	&
	& \clb		\splitreader{tinyimagenet}{resnet34}{clb}
	&
	& \iab		\splitreader{tinyimagenet}{resnet34}{iab}
	&
	& \wanet	\splitreader{tinyimagenet}{resnet34}{wanet}
	\cmidrule{3-15}
	&
	& \avg
	\splitreader[tabbgcolor]{tinyimagenet}{resnet34}{avg}
	&
	& \worst
	\splitreader[tabbgcolor]{tinyimagenet}{resnet34}{worstcase}
	\bottomrule
\end{splittable}};
	\end{tikzpicture}
	\else
	
	\fi
\end{table*}

\begin{table}[!b]
	\caption{Ablation study on \ourmethod. We highlight the best in \textbf{boldface} and mark the second-best in {\secondbest{gray bold}} font.}
	\label{tab:ablation-study}
	\begin{ablationtable}{\linewidth}{\bf \makecell[l]{Ablation}}
{\ourmethod}
& \bf 93.34			& 0.56
& \secondbest{93.37}& \secondbest{0.58}	
& \bf 93.11			& 0.77	
& \bf 93.56			& 1.22
\\
\midrule
{\rce~\textrightarrow~\ce}
& 89.85				& 0.23
& 87.82				& 3.01	
& 88.44				& \secondbest{0.58}	
& 84.07				& \bf 0.08
\\
{\rce~\textrightarrow~\sce}
& 90.90				& \secondbest{0.16}
& 93.01				& 0.61
& 92.76				& 0.68
& 89.82				& \secondbest{0.26} 
\\
{Unlearn \cleanD}
& 86.69				& 1.58	
& 88.12				& 2.53	
& 80.92				& 7.16
& 80.89				& 8.14
\\
{SL~\textrightarrow~\ssl}
& \secondbest{92.36}& \bf 0.07	
& \bf 94.12			& \bf 0.22	
& \secondbest{92.87}& \bf 0.16	
& \secondbest{92.44}& 0.48
\\
\bottomrule
\end{ablationtable}
\end{table}

\subsection{Ablation Study}
\label{sec:ablation-study}


We study the impact of different components of our method using a \resnet{18} model on poisoned \cifar{10} with the poisoning rate \perc{10}.
\cref{tab:ablation-study} summarizes the results.

\begin{enumerate}[label=\bf (\alph*), wide=0em] 
\item\textbf{Replacing \rce loss for dataset splitting.}
We use \ce loss and \sce loss with a threshold of \num{5.0} to show the impact of using \rce loss. 
While the backdoor is decently removed with \ce loss, the natural accuracy degrades significantly as the final \cleanD lacks benign samples.
The \rce term in \sce loss makes it remove poisonous samples better (and thus reduce the \asr) and also improves the natural accuracy.
We can further improve the result by tuning the threshold, which, however, requires ground-truth knowledge of the poisoned samples.
The fixed output range of \rce loss as used for \ourmethod, in turn, allows to trivially split the data at $\nicefrac{\RCEconst}{2}$, being much more practical.

\item\textbf{Impact of unlearning.} 
Next, we evaluate the influence of our unlearning step using \cleanD[i-1].
As can be seen in \cref{tab:ablation-study} unlearning is crucial for maintaining natural accuracy.
Learning a reference model on \subsetP[i] alone that still contains many benign samples lowers splitting effectiveness. 

\item\textbf{Final training using semi-supervised learning (\ssl).}
\ssl has been proven effective for training-time defenses before~\citep{Huang2022backdoor, Chen2022DST}.
Using \mixmatch~\citep{Berthelot2019mixmatch} for the final training on \cleanD and \poisonD achieves a comparable natural accuracy to standard supervised learning~(SL).
Despite a slight improvement in defense, \ssl's increased time consumption ($6\times$ compared to SL) outweighs \mbox{the benefit}.
\end{enumerate}


\begin{table}[!b]
	\caption{The time consumption (in hours) of all defenses across different datasets. Naive denotes the naive training.}
	\label{tab:comp_time}
	\tablesize
	\begin{timeconsumetable}{\columnwidth}
	\cifar{10}
	& 0.52	& 0.56	& 19.25	& 1.76	& 0.60	& 0.95
	\\
	\gtsrb
	& 0.42	& 0.55	& 16.21	& 1.43	& 0.47	& 0.74
	\\
	\tinyimagenet
	& 3.19	& 3.48	& 114.13& 10.59	& 3.60	& 5.68
	\\
	\bottomrule
\end{timeconsumetable}
\end{table}

\cref{tab:comp_time} summarizes the \textbf{time consumption} of all defenses. \ourmethod complete the defense slightly faster than \abl and \cbd.
In contrast, other defenses are slower by another magnitude. In particular, \dbd takes nearly one day on \cifar{10}.
While \ourmethod is not the fastest, it is in a comparable range as the naive training while achieving a significant performance increase as shown in \cref{tab:compare-sota} and \cref{tab:compare-sota-gtsrb}.




\section{Conclusion}
\label{sec:conclusion}
\iftrue
Neural backdoors are difficult to remove once established in a machine-learning model.
\ourmethod counters such attacks early by splitting poisonous samples off the training dataset to prevent learning the backdoors in the first place. 
High recall guarantees complete backdoor removal, while high precision ensures that we learn a performant (clean) model on the remaining benign samples. 
We find that these challenges are best met by learning a model that overfits the backdoor and use it as an oracle for poisonous samples in iterative dataset splitting.
This strategy represents a paradigm shift in splitting-based defenses, allowing for near-perfect removal of poisonous samples without harming the natural performance.
Most importantly, \ourmethod does not require any clean reference data.
Moreover, it works across model architectures and datasets both on average and in the worst-case, setting a new standard in training-time defenses.
\revision{
	The implementation of \ourmethod is publicly available at \smaller[0]{\projecturl}.
}

\else
Backdoor attacks establish a solid connection between the trigger pattern and the target label. Given a poisoned dataset, anti-backdoor learning must cope with the backdoor elimination and reproduce the natural performance in the model training. Thus, it is crucial to precisely isolate poisonous samples and preserve as many benign samples as possible. Our method, \ourmethod, takes advantage of the easy learning on the backdoor to capture poisonous samples and simultaneously bootstraps the forgetting of the natural performance. 
The resulting gap between the backdoor attack and natural performance, inversely, enables the splitting of benign samples in a poisoned dataset. 
Across different model architectures and datasets, learning the backdoor is always easy in our method, which underlines the vital meaning of learning the backdoor for backdoor removal in anti-backdoor learning.

\fi

\section*{Acknowledgment}

\revision{
The authors gratefully acknowledge funding by the Helmholtz Association (HGF) within topic “46.23 Engineering Secure Systems”, by SAP S.E. under project DE-2020-021, and by the German Federal Ministry of Education and Research (BMBF) under the project DataChainSec (FKZ 16KIS1700).
}


\bibliography{bib/references.bib}


\newpage

\appendix

%


In the appendix, we first discuss the data distribution using different loss functions in \cref{app:splitting-loss}, and provide \ourmethod's algorithmic description in \cref{alg:method}. In \cref{app:expm-details}, we summarize implementation details of all considered backdoor attacks and defenses, including \ourmethod. In \cref{app:expm-gtsrb}, we provide additional experiments on the dataset \gtsrb. In \cref{app:ablation-study-settings}, we conduct additional ablation study on \ourmethod. Finally, we investigate \ourmethod's robustness against the adaptive attack in \cref{app:adaptive-attack}.


\captionsetup[algorithm]{labelsep=colon}	
\begin{algorithm}[!t]
	\caption{Anti-Backdoor Learning of \ourmethod}
	\label{alg:method}
	\newcommand{\mystage}[2][]{\hspace*{\fill}\\\stage{#1}: #2}
\newcommand{\myinput}[1]{\textbf{Input:} #1}
\newcommand{\myoutput}[1]{\textbf{Output:} #1}

\newcommand{\mycomment}[1]{}

\myinput{
	the model with parameters \params,
	the poisoned dataset \poisonset, 
	the number of iterations \Niters in \stage{2}, and training epochs in \stage{1} - \threeb: \EpochsInit, \EpochsPoi, \EpochsCln.
}

\myoutput{clean model $\params$ and the benign set \cleanD.}

\mystage[1]{
	Initialization of the poisoned subset \subsetP[0]
}

\begin{algorithmic}[1]
	\State Clone model \params as the initial model \paramsnaive
	\State Naively train the model \paramsnaive for \EpochsInit epochs on $\poisonset$ by:
	
	\begin{equation*}
		\label{eq:poison_train}
		\centering
		\underset{\paramsnaive}{\min} ~\trainloss 
		= 
		\underset{\left( \inputx, \labely \right) \in \poisonset}{\sum} ~ 
		\sloss \left( \inputx, \labely, \paramsref \right)
	\end{equation*}
	
	\State \poisonD[0] $\leftarrow$ \perc{50} samples of \poisonset with lowest \rce loss 
	
	\State \cleanD[0] $\leftarrow$ $\poisonset\setminus\poisonD[0]$	
	\State Use \poisonD[0] as the subset \subsetP[1]
\end{algorithmic}

\mystage[2]{
	Learning the backdoor
}

\begin{algorithmic}[1]
	\State Initialize the model \params as the reference model \paramsref
	\For {i in $\left\{1, ..., \Niters \right\}$}
	
	\State Train $\paramsref^{(i)}$ for \EpochsPoi epochs on \subsetP[i] with loss:
	
	\begin{equation*}
		\hskip -8pt
		\sum_{\left( \inputx, \labely \right) \in \subsetP[i]} 
		sign\left( \lossce \left( \inputx, \labely, \paramsref[i] \right) - \gamma \right) 
		\cdot
		\lossce \left( \inputx, \labely, \paramsref[i] \right)
	\end{equation*}

	\State Unlearn \cleanD[i-1] for one epoch with the loss:
	
	\begin{equation*}
		\hskip -20pt
		\sum_{\left( \inputx, \labely \right) \in \cleanD[i-1]}
		-\lambda 
		\cdot
		\mathds{1} \left( f_{\paramsref[i]} \left(\inputx\right) = \labely \right)
		\cdot
		\lossce \left( \inputx, \labely, \paramsref[i] \right)
	\end{equation*}
	
	\State Calculate \rce loss of the entire \poisonset with $\paramsref[i]$
	
	\State Split \poisonset to \poisonD[i] and \cleanD[i] by the threshold of $\sfrac{\RCEconst}{2}$%
	
	\State \subsetP[i+1] $\gets$ \perc{50} samples of \poisonD[i] with lowest \rce 
	
	\EndFor
\end{algorithmic}

\mystage[3]{Meta-splitting on \poisonD[\Niters] with $\paramsref[1]$}
\begin{algorithmic}[1]
	\item $\{\MSclean, \MSpoison \} \leftarrow$ \text{Meta-split}  \poisonD[\Niters] by the model \paramsref[1]
	\item $\cleanD \leftarrow \cleanD[\Niters] + \MSclean$
\end{algorithmic}

\mystage[4]{Train the final model \params on \cleanD with the loss:}
\begin{equation*}
	\label{eq:ssl}
	\small
	\underset{\left( \inputx, \labely \right) \in \cleanD}{\sum} 
	\trainloss_{\ce} \left( \inputx, \labely, \params \right)
\end{equation*}

%
\end{algorithm}


\section{Extended Analysis of Loss-based Splitting}
\label{app:splitting-loss}

\begin{figure*}[!t]
	\hskip -4pt
	\begin{subfigure}{0.29\textwidth}
		\pgfplotstableread[col sep=space]{%
	bins	clean
	0.0	21.915
	0.10796628147363663	6.355
	0.21593256294727325	3.189
	0.3238988518714905	2.101
	0.4318651258945465	1.506
	0.5398313999176025	1.232
	0.647797703742981	1.029
	0.7557639479637146	0.857
	0.863730251789093	0.765
	0.9716964960098267	0.716
	1.079662799835205	0.61
	1.1876291036605835	0.525
	1.295595407485962	0.489
	1.4035615921020508	0.422
	1.5115278959274292	0.364
	1.6194941997528076	0.347
	1.727460503578186	0.302
	1.8354268074035645	0.252
	1.9433929920196533	0.248
	2.0513594150543213	0.203
	2.15932559967041	0.188
	2.267291784286499	0.19
	2.375258207321167	0.141
	2.483224391937256	0.133
	2.591190814971924	0.109
	2.6991569995880127	0.084
	2.8071231842041016	0.084
	2.9150896072387695	0.077
	3.0230557918548584	0.07
	3.1310222148895264	0.053
	3.2389883995056152	0.055
	3.346954584121704	0.047
	3.454921007156372	0.039
	3.562887191772461	0.031
	3.670853614807129	0.043
	3.7788197994232178	0.024
	3.8867859840393066	0.029
	3.9947524070739746	0.019
	4.102718830108643	0.011
	4.210684776306152	0.013
	4.31865119934082	0.011
	4.426617622375488	0.016
	4.534583568572998	0.011
	4.642549991607666	0.01
	4.750516414642334	0.008
	4.858482360839844	0.01
	4.966448783874512	0.004
	5.07441520690918	0.009
	5.182381629943848	0.009
	5.290347576141357	0.003
	5.398313999176025	0.007
	5.506280422210693	0.005
	5.614246368408203	0.003
	5.722212791442871	0.0
	5.830179214477539	0.001
	5.938145637512207	0.0
	6.046111583709717	0.001
	6.154078006744385	0.0
	6.262044429779053	0.0
	6.3700103759765625	0.004
	6.4779767990112305	0.001
	6.585943222045898	0.0
	6.693909168243408	0.0
	6.801875591278076	0.0
	6.909842014312744	0.001
	7.017807960510254	0.002
	7.125774383544922	0.0
	7.23374080657959	0.001
	7.341707229614258	0.002
	7.449673175811768	0.001
	7.5576395988464355	0.0
	7.6656060218811035	0.001
	7.773571968078613	0.0
	7.881538391113281	0.0
	7.989504814147949	0.0
	8.097471237182617	0.0
	8.205437660217285	0.0
	8.313403129577637	0.0
	8.421369552612305	0.0
	8.529335975646973	0.002
}\cleantable

\pgfplotstableread[col sep=space]{%
	bins	poison
	0.0	0.507
	0.10796628147363663	0.118
	0.21593256294727325	0.116
	0.3238988518714905	0.112
	0.4318651258945465	0.12
	0.5398313999176025	0.125
	0.647797703742981	0.122
	0.7557639479637146	0.116
	0.863730251789093	0.138
	0.9716964960098267	0.152
	1.079662799835205	0.164
	1.1876291036605835	0.158
	1.295595407485962	0.174
	1.4035615921020508	0.166
	1.5115278959274292	0.175
	1.6194941997528076	0.206
	1.727460503578186	0.18
	1.8354268074035645	0.169
	1.9433929920196533	0.16
	2.0513594150543213	0.151
	2.15932559967041	0.15
	2.267291784286499	0.142
	2.375258207321167	0.122
	2.483224391937256	0.123
	2.591190814971924	0.122
	2.6991569995880127	0.101
	2.8071231842041016	0.098
	2.9150896072387695	0.109
	3.0230557918548584	0.089
	3.1310222148895264	0.088
	3.2389883995056152	0.075
	3.346954584121704	0.054
	3.454921007156372	0.066
	3.562887191772461	0.052
	3.670853614807129	0.045
	3.7788197994232178	0.04
	3.8867859840393066	0.031
	3.9947524070739746	0.026
	4.102718830108643	0.028
	4.210684776306152	0.018
	4.31865119934082	0.019
	4.426617622375488	0.009
	4.534583568572998	0.011
	4.642549991607666	0.009
	4.750516414642334	0.01
	4.858482360839844	0.01
	4.966448783874512	0.006
	5.07441520690918	0.005
	5.182381629943848	0.005
	5.290347576141357	0.003
	5.398313999176025	0.004
	5.614246368408203	0.002
	5.722212791442871	0.003
}\poisontable

\begin{tikzpicture}
	\begin{axis}[
		ybar,
		height=0.4\linewidth,
		width=0.9\linewidth,
		legend pos=outer north east,
		enlarge x limits=0.02,
		grid=major, 
		grid style={dashed,gray!30},
		xlabel={\ce Loss},
		ylabel={\# Samples $\times 10^3$},
		xmin=0,
		xmax=11,
		ymin=0,
		ymax=8,
		xtick={0,1,...,16},
		xtick align=inside,
		ytick={0, 2, 4, 6, 8, 10},
		xticklabels={0,1,2,3,4,5,6,7,8,9,10,11,12,13,14,15,16},
		yticklabels={0, 2, 4, 6, 8, 10},
		yticklabel style = {font=\scriptsize, yshift=0.0ex},
		xticklabel style = {font=\scriptsize},
		ylabel style = {font=\footnotesize, yshift=-4ex},
		xlabel style = {font=\footnotesize, yshift=1ex},
		scale only axis,
		legend image post style={scale=0.75},
		legend style={at={(0.6,0.95)}, font=\legendsize, anchor=north west, legend columns=1, fill=white, draw=white, nodes={scale=1.0, 		transform shape}, column sep=2pt},
		legend cell align={left}
		]
		
		\addplot [ybar, bar width=1.8, draw=primarycolor, draw opacity=0, line width=0, fill=primarycolor, fill opacity=0.9, mark options={scale=0.5}] table[x index = 0, y index=1] {\cleantable};
		
	\end{axis}
	
	\begin{axis}[
		ybar,
		height=0.4\linewidth,
		width=0.9\linewidth,
		legend pos=outer north east,
		enlarge x limits=0.02,
		xlabel=\empty,
		ylabel=\empty,
		xmin=0,
		xmax=11,
		ymin=0,
		ymax=8,
		xtick=\empty,
		xtick align=inside,
		ytick=\empty,
		xticklabels=\empty,
		yticklabels=\empty,
		yticklabel style = {font=\scriptsize, yshift=0.0ex},
		xticklabel style = {font=\scriptsize},
		ylabel style = {font=\footnotesize, yshift=-4.5ex},
		xlabel style = {font=\footnotesize, yshift=1ex},
		scale only axis,
		legend image post style={scale=0.75},
		legend style={at={(0.6,0.75)}, font=\legendsize, anchor=north west, legend columns=1, fill=white, draw=white, nodes={scale=1.0, 		transform shape}, column sep=2pt},
		legend cell align={left}
		]
		
		\addplot[ybar, bar width=1.8, draw=secondarycolor, draw opacity=0, line width=0, fill=secondarycolor, fill opacity=0.8, mark options={scale=0.5}] table[x index = 0, y index=1] {\poisontable};
		
	\end{axis}
	
	%
	%
\end{tikzpicture}
		\pgfplotstableread[col sep=space]{%
	bins	clean
	0.0	13.691
	0.1619233936071396	1.789
	0.3238467872142792	1.245
	0.48577016592025757	1.008
	0.6476935744285583	0.841
	0.8096169233322144	0.757
	0.9715403318405151	0.691
	1.133463740348816	0.654
	1.2953871488571167	0.603
	1.4573105573654175	0.608
	1.6192338466644287	0.576
	1.7811572551727295	0.552
	1.9430806636810303	0.58
	2.105004072189331	0.537
	2.266927480697632	0.537
	2.4288508892059326	0.514
	2.5907742977142334	0.555
	2.752697706222534	0.477
	2.914621114730835	0.52
	3.0765442848205566	0.475
	3.2384676933288574	0.537
	3.400391101837158	0.481
	3.562314510345459	0.489
	3.7242379188537598	0.479
	3.8861613273620605	0.492
	4.048084735870361	0.504
	4.210008144378662	0.465
	4.371931552886963	0.428
	4.533854961395264	0.423
	4.6957783699035645	0.455
	4.857701778411865	0.431
	5.019625186920166	0.408
	5.181548595428467	0.431
	5.343472003936768	0.413
	5.505395412445068	0.387
	5.667318820953369	0.379
	5.82924222946167	0.412
	5.991165637969971	0.344
	6.153088569641113	0.335
	6.315011978149414	0.329
	6.476935386657715	0.316
	6.638858795166016	0.303
	6.800782203674316	0.335
	6.962705612182617	0.326
	7.124629020690918	0.322
	7.286552429199219	0.329
	7.4484758377075195	0.245
	7.61039924621582	0.26
	7.772322654724121	0.272
	7.934246063232422	0.275
	8.096169471740723	0.231
	8.258092880249023	0.216
	8.420016288757324	0.211
	8.581939697265625	0.252
	8.743863105773926	0.205
	8.905786514282227	0.193
	9.067709922790527	0.216
	9.229633331298828	0.182
	9.391556739807129	0.188
	9.55348014831543	0.192
	9.71540355682373	0.161
	9.877326965332031	0.167
	10.039250373840332	0.14
	10.201173782348633	0.146
	10.363097190856934	0.161
	10.525020599365234	0.134
	10.686944007873535	0.145
	10.848867416381836	0.124
	11.010790824890137	0.118
	11.172714233398438	0.116
	11.334637641906738	0.116
	11.496561050415039	0.113
	11.65848445892334	0.106
	11.82040786743164	0.101
	11.982331275939941	0.099
	12.144254684448242	0.085
	12.306177139282227	0.099
	12.468100547790527	0.106
	12.630023956298828	0.084
	12.791947364807129	0.086
	12.95387077331543	0.082
	13.11579418182373	0.066
	13.277717590332031	0.083
	13.439640998840332	0.058
	13.601564407348633	0.059
	13.763487815856934	0.065
	13.925411224365234	0.048
	14.087334632873535	0.051
	14.249258041381836	0.055
	14.411181449890137	0.046
	14.573104858398438	0.053
	14.735028266906738	0.063
	14.896951675415039	0.053
	15.05887508392334	0.053
	15.22079849243164	0.045
	15.382721900939941	0.041
	15.544645309448242	0.043
	15.706568717956543	0.036
	15.868492126464844	0.031
	16.03041648864746	0.034
	16.192338943481445	0.03
	16.35426139831543	0.032
	16.516185760498047	0.03
	16.67810821533203	0.023
	16.84003257751465	0.029
	17.001955032348633	0.02
	17.16387939453125	0.015
	17.325801849365234	0.018
	17.48772621154785	0.027
	17.649648666381836	0.025
	17.811573028564453	0.015
	17.973495483398438	0.014
	18.135419845581055	0.017
	18.29734230041504	0.019
	18.459266662597656	0.017
	18.62118911743164	0.026
	18.783113479614258	0.011
	18.945035934448242	0.015
	19.10696029663086	0.014
	19.268882751464844	0.016
	19.43080711364746	0.013
	19.592729568481445	0.011
	19.754653930664062	0.01
	19.916576385498047	0.011
	20.078500747680664	0.008
	20.24042320251465	0.007
	20.402347564697266	0.01
	20.56427001953125	0.007
	20.726194381713867	0.01
	20.88811683654785	0.009
	21.05004119873047	0.007
	21.211963653564453	0.007
	21.37388801574707	0.008
	21.535810470581055	0.011
	21.697734832763672	0.009
	21.859657287597656	0.002
	22.021581649780273	0.004
	22.183504104614258	0.001
	22.345428466796875	0.007
	22.50735092163086	0.003
	22.669275283813477	0.002
	22.83119773864746	0.0
	22.993122100830078	0.004
	23.155044555664062	0.003
	23.31696891784668	0.004
	23.478891372680664	0.004
	23.64081573486328	0.001
	23.802738189697266	0.002
	23.964662551879883	0.002
	24.126585006713867	0.002
	24.288509368896484	0.002
	24.45043182373047	0.0
	24.612354278564453	0.003
	24.77427864074707	0.0
	24.936201095581055	0.001
	25.098125457763672	0.003
	25.260047912597656	0.003
	25.421972274780273	0.003
	25.583894729614258	0.0
	25.745819091796875	0.004
	25.90774154663086	0.0
	26.069665908813477	0.0
	26.23158836364746	0.001
	26.393512725830078	0.001
	26.555435180664062	0.001
	26.71735954284668	0.003
	26.879281997680664	0.003
	27.04120635986328	0.0
	27.203128814697266	0.0
	27.365053176879883	0.0
	27.526975631713867	0.0
	27.688899993896484	0.001
	27.85082244873047	0.0
	28.012746810913086	0.0
	28.17466926574707	0.001
	28.336593627929688	0.0
	28.498516082763672	0.0
	28.66044044494629	0.001
	28.822362899780273	0.001
	28.98428726196289	0.0
	29.146209716796875	0.0
	29.308134078979492	0.0
	29.470056533813477	0.0
	29.631980895996094	0.0
	29.793903350830078	0.0
	29.955827713012695	0.0
	30.11775016784668	0.0
	30.279674530029297	0.0
	30.44159698486328	0.001
	30.6035213470459	0.0
	30.765443801879883	0.0
	30.9273681640625	0.0
	31.089290618896484	0.0
	31.2512149810791	0.0
	31.413137435913086	0.0
	31.575061798095703	0.0
	31.736984252929688	0.0
	31.898908615112305	0.0
	32.06083297729492	0.0
	32.222755432128906	0.002
}\cleantable

\pgfplotstableread[col sep=space]{%
	bins	poison
	0.0	4.999
}\poisontable

\begin{tikzpicture}
	\begin{axis}[
		ybar,
		height=0.4\linewidth,
		width=0.9\linewidth,
		legend pos=outer north east,
		enlarge x limits=0.02,
		grid=major, 
		grid style={dashed,gray!30},
		xlabel={\ce Loss},
		ylabel={\# Samples $\times 10^3$},
		xmin=0,
		xmax=11,
		ymin=0,
		ymax=8,
		xtick={0,1,...,16},
		xtick align=inside,
		ytick={0, 2, 4, ..., 10},
		xticklabels={0,1,2,3,4,5,6,7,8,9,10,11,12,13,14,15,16},
		yticklabels={0, 2, 4, 6, 8, 10},
		yticklabel style = {font=\scriptsize, yshift=0.0ex},
		xticklabel style = {font=\scriptsize},
		ylabel style = {font=\footnotesize, yshift=-4ex},
		xlabel style = {font=\footnotesize, yshift=1ex},
		scale only axis,
		legend image post style={scale=0.8},
		legend style={at={(0.6,0.95)}, font=\legendsize, anchor=north 
			west, legend columns=1, fill=white, draw=white, nodes={scale=1.0, 
				transform shape}, column sep=2pt},
		legend cell align={left}
		]
		
		\addplot [ybar, bar width=1.9, draw=primarycolor, draw opacity=0, line width=0, fill=primarycolor, fill opacity=0.9, mark options={scale=0.5}] table[x index = 0, y index=1] {\cleantable};
		
	\end{axis}
	
	\begin{axis}[
		ybar,
		height=0.4\linewidth,
		width=0.9\linewidth,
		legend pos=outer north east,
		enlarge x limits=0.02,
		xlabel=\empty,
		ylabel=\empty,
		xmin=0,
		xmax=11,
		ymin=0,
		ymax=8,
		xtick=\empty,
		xtick align=inside,
		ytick=\empty,
		xticklabels=\empty,
		yticklabels=\empty,
		yticklabel style = {font=\scriptsize, yshift=0.0ex},
		xticklabel style = {font=\scriptsize},
		ylabel style = {font=\footnotesize, yshift=-4.5ex},
		xlabel style = {font=\footnotesize, yshift=1ex},
		scale only axis,
		legend image post style={scale=0.8},
		legend style={at={(0.6,0.75)}, font=\legendsize, anchor=north west, legend columns=1, fill=white, draw=white, nodes={scale=1.0, 		transform shape}, column sep=2pt},
		legend cell align={left}
		]
		
		\addplot[ybar, bar width=1.9, draw=secondarycolor, draw opacity=0, line width=0, fill=secondarycolor, fill opacity=0.8, mark options={scale=0.5}] table[x index = 0, y index=1] {\poisontable};
		
	\end{axis}
	
	%
	%
\end{tikzpicture}
		\captionsetup{margin*={25pt, 0pt}}
		\label{fig:badnets-wrn-cifar10-ce}
	\end{subfigure}
	~~\qquad
	\begin{subfigure}{0.29\textwidth}
		\pgfplotstableread[col sep=space]{%
	bins	clean
	8.289306606457103e-06	28.833
	0.14751435816287994	2.07
	0.2950204312801361	1.173
	0.44252651929855347	0.819
	0.5900325775146484	0.654
	0.7375386357307434	0.515
	0.8850447535514832	0.483
	1.0325508117675781	0.41
	1.1800569295883179	0.351
	1.327562928199768	0.322
	1.4750690460205078	0.285
	1.622575044631958	0.225
	1.7700811624526978	0.243
	1.9175872802734375	0.229
	2.0650932788848877	0.218
	2.212599277496338	0.174
	2.360105514526367	0.185
	2.5076115131378174	0.191
	2.6551175117492676	0.173
	2.802623748779297	0.185
	2.950129747390747	0.148
	3.0976357460021973	0.129
	3.2451419830322266	0.14
	3.3926479816436768	0.145
	3.540153980255127	0.138
	3.6876602172851562	0.142
	3.8351662158966064	0.117
	3.9826722145080566	0.126
	4.130178451538086	0.142
	4.277684211730957	0.128
	4.425190448760986	0.115
	4.572696685791016	0.115
	4.720202445983887	0.129
	4.867708683013916	0.121
	5.015214920043945	0.107
	5.162720680236816	0.1
	5.310226917266846	0.127
	5.457733154296875	0.129
	5.605238914489746	0.12
	5.752745151519775	0.123
	5.900251388549805	0.114
	6.047757148742676	0.119
	6.195263385772705	0.122
	6.342769622802734	0.117
	6.4902753829956055	0.134
	6.637781620025635	0.124
	6.785287857055664	0.134
	6.932793617248535	0.108
	7.0802998542785645	0.127
	7.227806091308594	0.125
	7.375311851501465	0.137
	7.522818088531494	0.152
	7.670324325561523	0.155
	7.8178300857543945	0.159
	7.965336322784424	0.166
	8.112842559814453	0.169
	8.260348320007324	0.181
	8.407854080200195	0.18
	8.555360794067383	0.177
	8.702866554260254	0.194
	8.850372314453125	0.198
	8.997879028320312	0.238
	9.145384788513184	0.254
	9.292890548706055	0.293
	9.440397262573242	0.268
	9.587903022766113	0.241
	9.735408782958984	0.183
	9.882915496826172	0.072
	10.030421257019043	0.042
	10.177927017211914	0.014
	10.325433731079102	0.006
	10.472939491271973	0.008
	10.620445251464844	0.0
	10.767951965332031	0.0
	10.915457725524902	0.0
	11.062963485717773	0.0
	11.210470199584961	0.0
	11.357975959777832	0.0
	11.505481719970703	0.0
	11.65298843383789	0.0
}\cleantable

\pgfplotstableread[col sep=space]{%
	bins	poison
	8.289306606457103e-06	0.375
	0.14751435816287994	0.025
	0.2950204312801361	0.012
	0.44252651929855347	0.022
	0.5900325775146484	0.012
	0.7375386357307434	0.005
	0.8850447535514832	0.011
	1.0325508117675781	0.006
	1.1800569295883179	0.007
	1.327562928199768	0.007
	1.4750690460205078	0.003
	1.622575044631958	0.008
	1.7700811624526978	0.003
	1.9175872802734375	0.004
	2.0650932788848877	0.006
	2.360105514526367	0.003
	2.5076115131378174	0.008
	2.6551175117492676	0.004
	2.802623748779297	0.002
	2.950129747390747	0.004
	3.0976357460021973	0.007
	3.2451419830322266	0.004
	3.3926479816436768	0.004
	3.540153980255127	0.002
	3.6876602172851562	0.004
	3.8351662158966064	0.005
	3.9826722145080566	0.004
	4.130178451538086	0.002
	4.277684211730957	0.002
	4.425190448760986	0.004
	4.572696685791016	0.002
	4.867708683013916	0.005
	5.015214920043945	0.006
	5.162720680236816	0.003
	5.310226917266846	0.007
	5.457733154296875	0.007
	5.605238914489746	0.007
	5.752745151519775	0.01
	5.900251388549805	0.007
	6.047757148742676	0.007
	6.195263385772705	0.008
	6.342769622802734	0.006
	6.4902753829956055	0.008
	6.637781620025635	0.007
	6.785287857055664	0.006
	6.932793617248535	0.014
	7.0802998542785645	0.017
	7.227806091308594	0.018
	7.375311851501465	0.015
	7.522818088531494	0.016
	7.670324325561523	0.018
	7.8178300857543945	0.013
	7.965336322784424	0.017
	8.112842559814453	0.032
	8.260348320007324	0.028
	8.407854080200195	0.03
	8.555360794067383	0.038
	8.702866554260254	0.046
	8.850372314453125	0.087
	8.997879028320312	0.084
	9.145384788513184	0.122
	9.292890548706055	0.162
	9.440397262573242	0.293
	9.587903022766113	0.441
	9.735408782958984	0.547
	9.882915496826172	0.568
	10.030421257019043	0.472
	10.177927017211914	0.441
	10.325433731079102	0.341
	10.472939491271973	0.235
	10.620445251464844	0.117
	10.767951965332031	0.056
	10.915457725524902	0.042
	11.062963485717773	0.022
	11.210470199584961	0.01
	11.357975959777832	0.002
	11.65298843383789	0.002
}\poisontable

\begin{tikzpicture}
	\begin{axis}[
		ybar,
		height=0.4\linewidth,
		width=0.9\linewidth,
		legend pos=outer north east,
		enlarge x limits=0.02,
		grid=major, 
		grid style={dashed,gray!30},
		xlabel={\sce Loss},
		ylabel={\# Samples $\times 10^3$},
		xmin=0,
		xmax=11,
		ymin=0,
		ymax=8,
		xtick={0,1,...,16},
		xtick align=inside,
		ytick={0, 2, 4, 6, 8, 10},
		xticklabels={0,1,2,3,4,5,6,7,8,9,10,11,12,13,14,15,16},
		yticklabels={0, 2, 4, 6, 8, 10},
		yticklabel style = {font=\scriptsize, yshift=0.0ex},
		xticklabel style = {font=\scriptsize},
		ylabel style = {font=\footnotesize, yshift=-4ex},
		xlabel style = {font=\footnotesize, yshift=1ex},
		scale only axis,
		legend image post style={scale=0.75},
		legend style={at={(0.4,0.90)}, font=\legendsize, anchor=north west, legend columns=1, fill=white, draw=white, nodes={scale=1.0, 		transform shape}, column sep=2pt},
		legend cell align={left}
		]
		
		\addplot [ybar, bar width=1.8, draw=primarycolor, draw opacity=0, line width=0, fill=primarycolor, fill opacity=0.9, mark options={scale=0.5}] table[x index = 0, y index=1] {\cleantable};
		\addlegendentry{Benign samples};
		
	\end{axis}
	
	\begin{axis}[
		ybar,
		height=0.4\linewidth,
		width=0.9\linewidth,
		legend pos=outer north east,
		enlarge x limits=0.02,
		xlabel=\empty,
		ylabel=\empty,
		xmin=0,
		xmax=11,
		ymin=0,
		ymax=8,
		xtick=\empty,
		xtick align=inside,
		ytick=\empty,
		xticklabels=\empty,
		yticklabels=\empty,
		yticklabel style = {font=\scriptsize, yshift=0.0ex},
		xticklabel style = {font=\scriptsize},
		ylabel style = {font=\footnotesize, yshift=-4.5ex},
		xlabel style = {font=\footnotesize, yshift=1ex},
		scale only axis,
		legend image post style={scale=0.75},
		legend style={at={(0.4,0.70)}, font=\legendsize, anchor=north west, legend columns=1, fill=white, draw=white, nodes={scale=1.0, 		transform shape}, column sep=2pt},
		legend cell align={left}
		]
		
		\addplot[ybar, bar width=1.8, draw=secondarycolor, draw opacity=0, line width=0, fill=secondarycolor, fill opacity=0.8, mark options={scale=0.5}] table[x index = 0, y index=1] {\poisontable};
		\addlegendentry{Poisonous samples};
		
	\end{axis}
	
	%
	%
\end{tikzpicture}
		\pgfplotstableread[col sep=space]{%
	bins	clean
	8.289306606457103e-06	12.834
	0.06480494141578674	1.113
	0.1296015977859497	0.701
	0.19439825415611267	0.513
	0.25919491052627563	0.394
	0.3239915668964386	0.344
	0.38878822326660156	0.284
	0.4535848796367645	0.269
	0.5183815360069275	0.243
	0.5831781625747681	0.201
	0.6479748487472534	0.217
	0.712771475315094	0.161
	0.7775681614875793	0.187
	0.8423647880554199	0.163
	0.9071614742279053	0.156
	0.9719581007957458	0.134
	1.0367547273635864	0.123
	1.1015514135360718	0.112
	1.1663480997085571	0.124
	1.231144666671753	0.102
	1.2959413528442383	0.112
	1.3607380390167236	0.09
	1.425534725189209	0.105
	1.4903312921524048	0.09
	1.5551279783248901	0.092
	1.6199246644973755	0.085
	1.6847213506698608	0.09
	1.7495179176330566	0.08
	1.814314603805542	0.087
	1.8791112899780273	0.097
	1.9439079761505127	0.073
	2.008704662322998	0.078
	2.0735011100769043	0.086
	2.1382977962493896	0.073
	2.203094482421875	0.066
	2.2678911685943604	0.07
	2.3326878547668457	0.087
	2.397484540939331	0.076
	2.4622812271118164	0.067
	2.5270779132843018	0.085
	2.591874361038208	0.076
	2.6566710472106934	0.075
	2.7214677333831787	0.074
	2.786264419555664	0.068
	2.8510611057281494	0.055
	2.9158577919006348	0.076
	2.98065447807312	0.066
	3.0454509258270264	0.066
	3.1102476119995117	0.068
	3.175044298171997	0.071
	3.2398409843444824	0.069
	3.3046376705169678	0.064
	3.369434356689453	0.053
	3.4342310428619385	0.066
	3.499027729034424	0.058
	3.56382417678833	0.066
	3.6286208629608154	0.052
	3.693417549133301	0.063
	3.758214235305786	0.06
	3.8230109214782715	0.058
	3.887807607650757	0.061
	3.952604293823242	0.051
	4.017400741577148	0.064
	4.082197666168213	0.048
	4.146994113922119	0.067
	4.211791038513184	0.067
	4.27658748626709	0.062
	4.341383934020996	0.064
	4.4061808586120605	0.05
	4.470977306365967	0.049
	4.535774230957031	0.059
	4.6005706787109375	0.06
	4.665367603302002	0.053
	4.730164051055908	0.058
	4.7949604988098145	0.062
	4.859757423400879	0.052
	4.924553871154785	0.063
	4.98935079574585	0.059
	5.054147243499756	0.068
	5.11894416809082	0.065
	5.183740615844727	0.062
	5.248537063598633	0.067
	5.313333988189697	0.054
	5.3781304359436035	0.051
	5.442927360534668	0.061
	5.507723808288574	0.051
	5.572520732879639	0.062
	5.637317180633545	0.058
	5.702113628387451	0.056
	5.766910552978516	0.062
	5.831707000732422	0.065
	5.896503925323486	0.066
	5.961300373077393	0.065
	6.026097297668457	0.075
	6.090893745422363	0.065
	6.155690670013428	0.083
	6.220487117767334	0.067
	6.28528356552124	0.07
	6.350080490112305	0.069
	6.414876937866211	0.083
	6.479673862457275	0.081
	6.544470310211182	0.066
	6.609267234802246	0.071
	6.674063682556152	0.067
	6.738860130310059	0.086
	6.803657054901123	0.072
	6.868453502655029	0.071
	6.933250427246094	0.073
	6.998046875	0.08
	7.0628437995910645	0.083
	7.127640247344971	0.081
	7.192436695098877	0.064
	7.257233619689941	0.101
	7.322030067443848	0.084
	7.386826992034912	0.091
	7.451623439788818	0.096
	7.516420364379883	0.085
	7.581216812133789	0.121
	7.646013259887695	0.103
	7.71081018447876	0.12
	7.775606632232666	0.099
	7.8404035568237305	0.101
	7.905200004577637	0.103
	7.969996929168701	0.114
	8.034793853759766	0.13
	8.099590301513672	0.117
	8.164386749267578	0.117
	8.229183197021484	0.13
	8.29397964477539	0.104
	8.358777046203613	0.141
	8.42357349395752	0.158
	8.488369941711426	0.146
	8.553166389465332	0.157
	8.617962837219238	0.18
	8.682760238647461	0.147
	8.747556686401367	0.215
	8.812353134155273	0.182
	8.87714958190918	0.233
	8.941946983337402	0.236
	9.006743431091309	0.254
	9.071539878845215	0.267
	9.136336326599121	0.288
	9.201132774353027	0.331
	9.26593017578125	0.424
	9.330726623535156	0.384
	9.395523071289062	0.496
	9.460319519042969	0.565
	9.525116920471191	0.609
	9.589913368225098	0.736
	9.654709815979004	0.794
	9.71950626373291	0.872
	9.784302711486816	0.917
	9.849100112915039	0.909
	9.913896560668945	0.956
	9.978693008422852	0.867
	10.043489456176758	0.843
	10.108285903930664	0.726
	10.173083305358887	0.688
	10.237879753112793	0.609
	10.3026762008667	0.54
	10.367472648620605	0.456
	10.432270050048828	0.391
	10.497066497802734	0.385
	10.56186294555664	0.324
	10.626659393310547	0.282
	10.691455841064453	0.232
	10.756253242492676	0.221
	10.821049690246582	0.193
	10.885846138000488	0.163
	10.950642585754395	0.144
	11.015439987182617	0.112
	11.080236434936523	0.104
	11.14503288269043	0.084
	11.209829330444336	0.089
	11.274625778198242	0.06
	11.339423179626465	0.058
	11.404219627380371	0.042
	11.469016075134277	0.024
	11.533812522888184	0.039
	11.59860897064209	0.02
	11.663406372070312	0.022
	11.728202819824219	0.024
	11.792999267578125	0.02
	11.857795715332031	0.018
	11.922593116760254	0.012
	11.98738956451416	0.006
	12.052186012268066	0.013
	12.116982460021973	0.007
	12.181778907775879	0.008
	12.246576309204102	0.004
	12.311372756958008	0.005
	12.376169204711914	0.003
	12.44096565246582	0.002
	12.505762100219727	0.004
	12.57055950164795	0.0
	12.635355949401855	0.004
	12.700152397155762	0.001
	12.764948844909668	0.0
	12.82974624633789	0.0
	12.894542694091797	0.001
}\cleantable

\pgfplotstableread[col sep=space]{%
	bins	poison
	8.289306606457103e-06	4.995
}\poisontable

\begin{tikzpicture}
	\begin{axis}[
		ybar,
		height=0.4\linewidth,
		width=0.9\linewidth,
		legend pos=outer north east,
		enlarge x limits=0.02,
		grid=major, 
		grid style={dashed,gray!30},
		xlabel={\sce Loss},
		ylabel={\# Samples $\times 10^3$},
		xmin=0,
		xmax=11,
		ymin=0,
		ymax=8,
		xtick={0,1,...,16},
		xtick align=inside,
		ytick={0, 2, 4, ..., 10},
		xticklabels={0,1,2,3,4,5,6,7,8,9,10,11,12,13,14,15,16},
		yticklabels={0, 2, 4, 6, 8, 10},
		yticklabel style = {font=\scriptsize, yshift=0.0ex},
		xticklabel style = {font=\scriptsize},
		ylabel style = {font=\footnotesize, yshift=-4ex},
		xlabel style = {font=\footnotesize, yshift=1ex},
		scale only axis,
		legend image post style={scale=0.8},
		legend style={at={(0.6,0.95)}, font=\legendsize, anchor=north 
			west, legend columns=1, fill=white, draw=white, nodes={scale=1.0, 
				transform shape}, column sep=2pt},
		legend cell align={left}
		]
		
		\addplot [ybar, bar width=1.9, draw=primarycolor, draw opacity=0, line width=0, fill=primarycolor, fill opacity=0.9, mark options={scale=0.5}] table[x index = 0, y index=1] {\cleantable};
		
	\end{axis}
	
	\begin{axis}[
		ybar,
		height=0.4\linewidth,
		width=0.9\linewidth,
		legend pos=outer north east,
		enlarge x limits=0.02,
		xlabel=\empty,
		ylabel=\empty,
		xmin=0,
		xmax=11,
		ymin=0,
		ymax=8,
		xtick=\empty,
		xtick align=inside,
		ytick=\empty,
		xticklabels=\empty,
		yticklabels=\empty,
		yticklabel style = {font=\scriptsize, yshift=0.0ex},
		xticklabel style = {font=\scriptsize},
		ylabel style = {font=\footnotesize, yshift=-4.5ex},
		xlabel style = {font=\footnotesize, yshift=1ex},
		scale only axis,
		legend image post style={scale=0.8},
		legend style={at={(0.6,0.75)}, font=\legendsize, anchor=north west, legend columns=1, fill=white, draw=white, nodes={scale=1.0, 		transform shape}, column sep=2pt},
		legend cell align={left}
		]
		
		\addplot[ybar, bar width=1.9, draw=secondarycolor, draw opacity=0, line width=0, fill=secondarycolor, fill opacity=0.8, mark options={scale=0.5}] table[x index = 0, y index=1] {\poisontable};
		
	\end{axis}
	
	%
	%
\end{tikzpicture}
		\captionsetup{margin*={20pt, 0pt}}
		\label{fig:badnets-cifar10-sce}
	\end{subfigure}
	~~\qquad
	\begin{subfigure}{0.29\textwidth}
		\pgfplotstableread[col sep=space]{%
	bins	clean
	8.289306606457103e-06	28.107
	0.11513753980398178	2.136
	0.2302667796611786	1.147
	0.3453960418701172	0.834
	0.4605252742767334	0.632
	0.5756545066833496	0.54
	0.6907837986946106	0.442
	0.8059130311012268	0.399
	0.921042263507843	0.349
	1.036171555519104	0.337
	1.1513007879257202	0.277
	1.2664300203323364	0.248
	1.3815592527389526	0.261
	1.4966884851455688	0.222
	1.611817717552185	0.173
	1.7269469499588013	0.186
	1.842076301574707	0.19
	1.9572055339813232	0.172
	2.0723347663879395	0.181
	2.1874639987945557	0.136
	2.302593231201172	0.147
	2.417722463607788	0.155
	2.5328516960144043	0.141
	2.6479809284210205	0.137
	2.7631101608276367	0.147
	2.878239393234253	0.126
	2.993368625640869	0.112
	3.1084980964660645	0.106
	3.2236273288726807	0.11
	3.338756561279297	0.115
	3.453885793685913	0.111
	3.5690150260925293	0.107
	3.6841442584991455	0.111
	3.7992734909057617	0.097
	3.914402723312378	0.096
	4.029531955718994	0.111
	4.1446614265441895	0.101
	4.259790420532227	0.102
	4.374919891357422	0.097
	4.490048885345459	0.08
	4.605178356170654	0.109
	4.720307350158691	0.102
	4.835436820983887	0.093
	4.950565814971924	0.09
	5.065695285797119	0.073
	5.180824279785156	0.099
	5.295953750610352	0.099
	5.411082744598389	0.097
	5.526212215423584	0.102
	5.641341209411621	0.093
	5.756470680236816	0.093
	5.871600151062012	0.108
	5.986729145050049	0.09
	6.101858615875244	0.093
	6.216987609863281	0.099
	6.332117080688477	0.102
	6.447246074676514	0.106
	6.562375545501709	0.103
	6.677504539489746	0.111
	6.792634010314941	0.079
	6.9077630043029785	0.11
	7.022892475128174	0.097
	7.138021469116211	0.107
	7.253150939941406	0.111
	7.368279933929443	0.13
	7.483409404754639	0.115
	7.598538875579834	0.134
	7.713667869567871	0.147
	7.828797340393066	0.144
	7.9439263343811035	0.133
	8.05905532836914	0.15
	8.174184799194336	0.164
	8.289314270019531	0.149
	8.404443740844727	0.161
	8.519572257995605	0.181
	8.6347017288208	0.206
	8.749831199645996	0.24
	8.864960670471191	0.292
	8.98008918762207	0.333
	9.095218658447266	0.667
}\cleantable

\pgfplotstableread[col sep=space]{%
	bins	poison
	8.289306606457103e-06	0.359
	0.11513753980398178	0.032
	0.2302667796611786	0.014
	0.3453960418701172	0.01
	0.4605252742767334	0.019
	0.5756545066833496	0.006
	0.6907837986946106	0.009
	0.8059130311012268	0.007
	0.921042263507843	0.006
	1.036171555519104	0.006
	1.1513007879257202	0.005
	1.2664300203323364	0.005
	1.3815592527389526	0.005
	1.4966884851455688	0.003
	1.611817717552185	0.005
	1.7269469499588013	0.005
	1.9572055339813232	0.003
	2.0723347663879395	0.006
	2.302593231201172	0.002
	2.417722463607788	0.004
	2.5328516960144043	0.008
	2.6479809284210205	0.002
	2.7631101608276367	0.002
	2.878239393234253	0.002
	2.993368625640869	0.005
	3.1084980964660645	0.005
	3.2236273288726807	0.002
	3.338756561279297	0.004
	3.5690150260925293	0.003
	3.6841442584991455	0.003
	3.7992734909057617	0.005
	4.029531955718994	0.005
	4.374919891357422	0.002
	4.490048885345459	0.003
	4.835436820983887	0.006
	4.950565814971924	0.005
	5.065695285797119	0.003
	5.295953750610352	0.007
	5.411082744598389	0.007
	5.526212215423584	0.005
	5.641341209411621	0.01
	5.756470680236816	0.004
	5.871600151062012	0.007
	5.986729145050049	0.005
	6.101858615875244	0.006
	6.216987609863281	0.005
	6.332117080688477	0.007
	6.447246074676514	0.004
	6.562375545501709	0.006
	6.677504539489746	0.006
	6.792634010314941	0.011
	6.9077630043029785	0.012
	7.022892475128174	0.018
	7.138021469116211	0.01
	7.253150939941406	0.013
	7.368279933929443	0.013
	7.483409404754639	0.018
	7.598538875579834	0.007
	7.713667869567871	0.015
	7.828797340393066	0.018
	7.9439263343811035	0.028
	8.05905532836914	0.022
	8.174184799194336	0.025
	8.289314270019531	0.032
	8.404443740844727	0.038
	8.519572257995605	0.054
	8.6347017288208	0.076
	8.749831199645996	0.1
	8.864960670471191	0.123
	8.98008918762207	0.296
	9.095218658447266	3.418
}\poisontable

\begin{tikzpicture}
	\begin{axis}[
		ybar,
		height=0.4\linewidth,
		width=0.9\linewidth,
		legend pos=outer north east,
		enlarge x limits=0.02,
		grid=major, 
		grid style={dashed,gray!30},
		xlabel={\rce Loss},
		ylabel={\# Samples $\times 10^3$},
		xmin=0,
		xmax=11,
		ymin=0,
		ymax=8,
		xtick={0,1,...,16},
		xtick align=inside,
		ytick={0, 2, 4, 6, 8, 10},
		xticklabels={0,1,2,3,4,5,6,7,8,9,10,11,12,13,14,15,16},
		yticklabels={0, 2, 4, 6, 8, 10},
		yticklabel style = {font=\scriptsize, yshift=0.0ex},
		xticklabel style = {font=\scriptsize},
		ylabel style = {font=\footnotesize, yshift=-4ex},
		xlabel style = {font=\footnotesize, yshift=1ex},
		scale only axis,
		legend image post style={scale=0.75},
		legend style={at={(0.6,0.95)}, font=\legendsize, anchor=north west, legend columns=1, fill=white, draw=white, nodes={scale=1.0, 		transform shape}, column sep=2pt},
		legend cell align={left}
		]
		
		\addplot [ybar, bar width=1.8, draw=primarycolor, draw opacity=0, line width=0, fill=primarycolor, fill opacity=0.9, mark options={scale=0.5}] table[x index = 0, y index=1] {\cleantable};
		
	\end{axis}
	
	\begin{axis}[
		ybar,
		height=0.4\linewidth,
		width=0.9\linewidth,
		legend pos=outer north east,
		enlarge x limits=0.02,
		xlabel=\empty,
		ylabel=\empty,
		xmin=0,
		xmax=11,
		ymin=0,
		ymax=8,
		xtick=\empty,
		xtick align=inside,
		ytick=\empty,
		xticklabels=\empty,
		yticklabels=\empty,
		yticklabel style = {font=\scriptsize, yshift=0.0ex},
		xticklabel style = {font=\scriptsize},
		ylabel style = {font=\footnotesize, yshift=-4.5ex},
		xlabel style = {font=\footnotesize, yshift=1ex},
		scale only axis,
		legend image post style={scale=0.75},
		legend style={at={(0.6,0.75)}, font=\legendsize, anchor=north west, legend columns=1, fill=white, draw=white, nodes={scale=1.0, 		transform shape}, column sep=2pt},
		legend cell align={left}
		]
		
		\addplot[ybar, bar width=1.8, draw=secondarycolor, draw opacity=0, line width=0, fill=secondarycolor, fill opacity=0.8, mark options={scale=0.5}] table[x index = 0, y index=1] {\poisontable};
		
	\end{axis}
	
	%
	%
\end{tikzpicture}
		\pgfplotstableread[col sep=space]{%
	bins	clean
	8.289306606457103e-06	14.385
	0.0460599884390831	1.107
	0.09211168438196182	0.677
	0.13816338777542114	0.484
	0.18421508371829987	0.378
	0.2302667796611786	0.303
	0.2763184905052185	0.248
	0.32237017154693604	0.243
	0.36842188239097595	0.191
	0.41447359323501587	0.184
	0.4605252742767334	0.173
	0.5065769553184509	0.166
	0.5526286959648132	0.137
	0.5986803770065308	0.131
	0.6447320580482483	0.114
	0.6907837986946106	0.121
	0.7368354797363281	0.088
	0.7828871607780457	0.117
	0.828938901424408	0.089
	0.8749905824661255	0.1
	0.921042263507843	0.083
	0.9670939445495605	0.066
	1.0131456851959229	0.095
	1.0591974258422852	0.058
	1.105249047279358	0.073
	1.1513007879257202	0.099
	1.197352409362793	0.079
	1.2434041500091553	0.063
	1.2894558906555176	0.075
	1.3355075120925903	0.069
	1.3815592527389526	0.06
	1.427610993385315	0.061
	1.4736626148223877	0.062
	1.51971435546875	0.066
	1.5657660961151123	0.048
	1.611817717552185	0.061
	1.6578694581985474	0.061
	1.7039211988449097	0.046
	1.7499728202819824	0.06
	1.7960245609283447	0.061
	1.842076301574707	0.054
	1.8881279230117798	0.046
	1.934179663658142	0.05
	1.9802314043045044	0.051
	2.026283025741577	0.055
	2.0723347663879395	0.043
	2.1183865070343018	0.05
	2.164438247680664	0.045
	2.2104897499084473	0.042
	2.2565414905548096	0.044
	2.302593231201172	0.055
	2.348644971847534	0.058
	2.3946967124938965	0.046
	2.440748453140259	0.044
	2.486799955368042	0.048
	2.5328516960144043	0.046
	2.5789034366607666	0.04
	2.624955177307129	0.031
	2.671006917953491	0.035
	2.7170586585998535	0.033
	2.7631101608276367	0.046
	2.809161901473999	0.035
	2.8552136421203613	0.031
	2.9012653827667236	0.039
	2.947317123413086	0.036
	2.993368625640869	0.044
	3.0394203662872314	0.037
	3.0854721069335938	0.052
	3.131523847579956	0.038
	3.1775755882263184	0.045
	3.2236273288726807	0.039
	3.269678831100464	0.033
	3.315730571746826	0.038
	3.3617823123931885	0.035
	3.407834053039551	0.037
	3.453885793685913	0.04
	3.4999375343322754	0.035
	3.5459890365600586	0.042
	3.592040777206421	0.029
	3.638092517852783	0.034
	3.6841442584991455	0.037
	3.730195999145508	0.031
	3.77624773979187	0.027
	3.8222992420196533	0.037
	3.8683509826660156	0.036
	3.914402723312378	0.03
	3.9604544639587402	0.035
	4.006505966186523	0.042
	4.052557945251465	0.04
	4.098609447479248	0.038
	4.1446614265441895	0.038
	4.190712928771973	0.035
	4.236764430999756	0.047
	4.282816410064697	0.039
	4.3288679122924805	0.037
	4.374919891357422	0.034
	4.420971393585205	0.037
	4.467022895812988	0.035
	4.51307487487793	0.049
	4.559126377105713	0.033
	4.605178356170654	0.032
	4.6512298583984375	0.044
	4.697281837463379	0.034
	4.743333339691162	0.042
	4.789384841918945	0.037
	4.835436820983887	0.038
	4.88148832321167	0.025
	4.927540302276611	0.042
	4.9735918045043945	0.034
	5.019643306732178	0.032
	5.065695285797119	0.041
	5.111746788024902	0.045
	5.157798767089844	0.04
	5.203850269317627	0.041
	5.24990177154541	0.036
	5.295953750610352	0.044
	5.342005252838135	0.042
	5.388057231903076	0.029
	5.434108734130859	0.036
	5.480160713195801	0.036
	5.526212215423584	0.036
	5.572263717651367	0.03
	5.618315696716309	0.038
	5.664367198944092	0.045
	5.710419178009033	0.029
	5.756470680236816	0.032
	5.8025221824646	0.046
	5.848574161529541	0.041
	5.894625663757324	0.039
	5.940677642822266	0.036
	5.986729145050049	0.034
	6.032780647277832	0.047
	6.078832626342773	0.035
	6.124884128570557	0.043
	6.170936107635498	0.042
	6.216987609863281	0.038
	6.263039588928223	0.035
	6.309091091156006	0.037
	6.355142593383789	0.051
	6.4011945724487305	0.052
	6.447246074676514	0.041
	6.493298053741455	0.04
	6.539349555969238	0.038
	6.5854010581970215	0.05
	6.631453037261963	0.036
	6.677504539489746	0.043
	6.7235565185546875	0.05
	6.769608020782471	0.04
	6.815659999847412	0.052
	6.861711502075195	0.053
	6.9077630043029785	0.05
	6.95381498336792	0.054
	6.999866485595703	0.043
	7.0459184646606445	0.057
	7.091969966888428	0.05
	7.138021469116211	0.05
	7.184073448181152	0.052
	7.2301249504089355	0.06
	7.276176929473877	0.059
	7.32222843170166	0.06
	7.368279933929443	0.051
	7.414331912994385	0.062
	7.460383415222168	0.058
	7.506435394287109	0.053
	7.552486896514893	0.051
	7.598538875579834	0.048
	7.644590377807617	0.065
	7.6906418800354	0.067
	7.736693859100342	0.068
	7.782745361328125	0.073
	7.828797340393066	0.076
	7.87484884262085	0.074
	7.920900344848633	0.07
	7.966952323913574	0.077
	8.013004302978516	0.093
	8.05905532836914	0.075
	8.105107307434082	0.079
	8.151159286499023	0.075
	8.197210311889648	0.116
	8.24326229095459	0.097
	8.289314270019531	0.09
	8.335366249084473	0.111
	8.381417274475098	0.128
	8.427469253540039	0.118
	8.47352123260498	0.12
	8.519572257995605	0.14
	8.565624237060547	0.146
	8.611676216125488	0.124
	8.65772819519043	0.168
	8.703779220581055	0.171
	8.749831199645996	0.194
	8.795883178710938	0.188
	8.841934204101562	0.246
	8.887986183166504	0.25
	8.934038162231445	0.307
	8.98008918762207	0.38
	9.026141166687012	0.511
	9.072193145751953	0.672
	9.118245124816895	1.173
	9.16429615020752	12.621
}\cleantable

\pgfplotstableread[col sep=space]{%
	bins	poison
	8.289306606457103e-06	4.956
	0.13816338777542114	0.005
	0.2302667796611786	0.002
	0.2763184905052185	0.002
	0.5986803770065308	0.002
	8.47352123260498	0.002
	8.887986183166504	0.002
	9.16429615020752	0.007
}\poisontable

\begin{tikzpicture}
	\begin{axis}[
		ybar,
		height=0.4\linewidth,
		width=0.9\linewidth,
		legend pos=outer north east,
		enlarge x limits=0.02,
		grid=major, 
		grid style={dashed,gray!30},
		xlabel={$\RCE$ Loss},
		ylabel={\# Samples $\times 10^3$},
		xmin=0,
		xmax=11,
		ymin=0,
		ymax=8,
		xtick={0,1,...,16},
		xtick align=inside,
		ytick={0, 2, 4, ..., 10},
		xticklabels={0,1,2,3,4,5,6,7,8,9,10,11,12,13,14,15,16},
		yticklabels={0, 2, 4, 6, 8, 10},
		yticklabel style = {font=\scriptsize, yshift=0.0ex},
		xticklabel style = {font=\scriptsize},
		ylabel style = {font=\footnotesize, yshift=-4ex},
		xlabel style = {font=\footnotesize, yshift=1ex},
		scale only axis,
		legend image post style={scale=0.8},
		legend style={at={(0.6,0.95)}, font=\legendsize, anchor=north 
			west, legend columns=1, fill=white, draw=white, nodes={scale=1.0, 
				transform shape}, column sep=2pt},
		legend cell align={left}
		]
		
		\addplot [ybar, bar width=1.9, draw=primarycolor, draw opacity=0, line width=0, fill=primarycolor, fill opacity=0.9, mark options={scale=0.5}] table[x index = 0, y index=1] {\cleantable};
		
	\end{axis}
	
	\begin{axis}[
		ybar,
		height=0.4\linewidth,
		width=0.9\linewidth,
		legend pos=outer north east,
		enlarge x limits=0.02,
		xlabel=\empty,
		ylabel=\empty,
		xmin=0,
		xmax=11,
		ymin=0,
		ymax=8,
		xtick=\empty,
		xtick align=inside,
		ytick=\empty,
		xticklabels=\empty,
		yticklabels=\empty,
		yticklabel style = {font=\scriptsize, yshift=0.0ex},
		xticklabel style = {font=\scriptsize},
		ylabel style = {font=\footnotesize, yshift=-4.5ex},
		xlabel style = {font=\footnotesize, yshift=1ex},
		scale only axis,
		legend image post style={scale=0.8},
		legend style={at={(0.6,0.75)}, font=\legendsize, anchor=north west, legend columns=1, fill=white, draw=white, nodes={scale=1.0, 		transform shape}, column sep=2pt},
		legend cell align={left}
		]
		
		\addplot[ybar, bar width=1.9, draw=secondarycolor, draw opacity=0, line width=0, fill=secondarycolor, fill opacity=0.8, mark options={scale=0.5}] table[x index = 0, y index=1] {\poisontable};
		
	\end{axis}
	
	%
	%
\end{tikzpicture}
		\captionsetup{margin*={20pt, 0pt}}
		\label{fig:badnets-cifar10-rce}
	\end{subfigure}
	
	\caption{Comparing data distribution in dataset splitting defenses by different loss functions: \ce~(left), \sce~(middle), and \rce~(right). Distribution of \dbd and \ourmethod are shown in the top and bottom row, respectively. All histograms represent the distribution baselines of each defense at the first splitting moment on the entire \poisonset (\dbd: right before the final training with data labels, \ourmethod: right after the first iteration of \stage{2}) with a \resnet{18} trained on \cifar{10} poisoned by \blend attack.}
	\label{fig:sce-ce-rce}
\end{figure*}
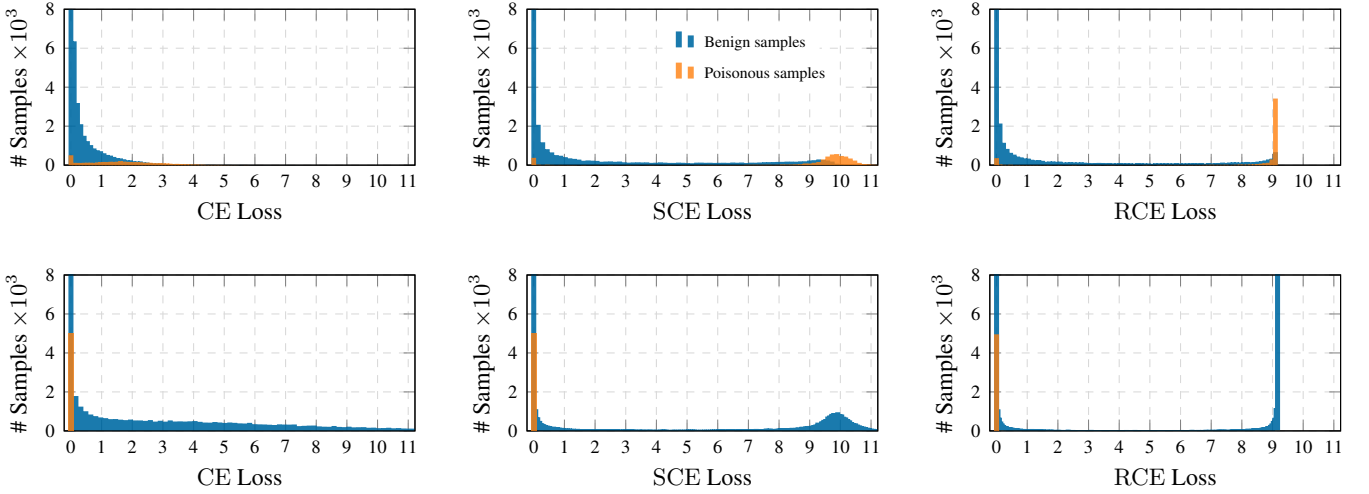

We compare the dataset splitting of \dbd and \ourmethod with different loss functions and show the data distribution of each in \cref{fig:sce-ce-rce}. Using \ce loss alone cannot ensure a clear splitting, particularly in \dbd, due to the strong overlap of of benign and poisonous samples. By the adding the \rce loss item, \sce can distinguish most poisonous samples in the distribution yielded by \dbd's unsupervised learning, while the splitting threshold is not fixed due to the unlimited range of the \ce loss. Using \rce loss in \dbd's splitting, in turn, makes poisonous samples significantly separate from most benign samples, which is beneficial for the dataset splitting. 

Hence, we use \rce loss alone in \dbd defense. Interestingly, using \rce loss improves \dbd's isolation of poisonous samples and, thus, makes \dbd more robust against some backdoor attacks (\cf~\cref{tab:dbd-rce}). However, learning on the benign samples is a difficult task, leading to an overlap of poisonous samples and some hard-to-learn benign samples. Meanwhile, a few poisonous samples originally from the backdoor target class remain with the low loss that causes learning the backdoor again in \dbd. In this case, using \rce loss cannot make \dbd avoid learning on the backdoor or improve the performance of the primary task (\ie natural accuracy). Learning a backdoor oracle makes poisonous samples concentrated in the low loss (\cf the second row of \cref{fig:sce-ce-rce}), particularly by using \rce, ensuring the defense and the natural performance, thus, making \ourmethod excel.

\begin{table}[!h]
	\caption{\dbd defense using different losses for the dataset splitting on poisoned \cifar{10} with a \resnet{18} model.}
	\label{tab:dbd-rce}
	\begin{dbdlosstable}{\linewidth}
		\badnets
		& 92.78	& 99.98	& 16.36
		& \bf 93.24	& 14.65	& 2.91	
		& 91.47	& \bf 8.26	& \bf 1.04
		\\
		\blend
		& \bf 90.75	& 99.98	& 15.48
		& 90.27	& 99.91	& 11.61
		& 90.52	& \bf 96.83	& \bf 6.64
		\\
		\iab
		& \bf 88.24	& 82.76	& 16.08
		& 87.91	& 83.23	& 11.58
		& 86.43	& \bf 53.25	& \bf 4.17
		\\
		\wanet
		& 92.46		& 35.27		& 4.63
		& \bf 93.03		& 15.27		& 2.83
		& 91.95		& \bf 3.38	& \bf 1.95
		\\
		\midrule
	\end{dbdlosstable}
\end{table}

\iftrue
\section{Implementation Details}
\label{app:expm-details}

In this section, we summarize the experimental setup of our evaluation, including the setup for all considered attacks (\cf~\cref{app:attack-setup}) and defenses (\cf~\cref{app:defense-setup}).


\subsection{Experimental Setup}
\label{sec:exp-setup-app}

In the following, we detail the setup of our experiments. Note that we run each experiment on a NVIDIA GeForce RTX~3090 GPU with 24GB memory.

\paragraph{Datasets and models} 
\cref{tab:ds-model} lists the details of considered datasets. For each, we randomly select samples across all classes for the dataset poisoning. We use the original test set to evaluate the natural accuracy (\acc) and poison the test data of non-targeted classes to determine the attack success rate (\asr). Since prior defenses use either \wrn~\cite{Li2021antibackdoor, Zhang2023backdoor} or \resnet{18}~\cite{Huang2022backdoor, Chen2022DST, Gao2023asd} on small-scale datasets, we conduct experiments on both architectures for the best fairness.

\begin{table}[!h]
	\caption{Summary of datasets in our evaluation.}
	\label{tab:ds-model}
	\tablesize
	\begin{dsmodeltable}{\columnwidth}
		\cifar{10}
		& {$3\times32\times32$}
		& 10
		& 50000
		& 10000
		\\ 
		\gtsrb
		& {$3\times32\times32$}
		& 43
		& 39209
		& 12630
		\\ 
		\makecell[l]{\tinyimagenet}
		& {$3\times64\times64$}
		& 200
		& 100000
		& 10000
		\\
		\bottomrule
	\end{dsmodeltable}
\end{table}

\subsection{Attacks Setups}
\label{app:attack-setup}

We implement each attack by conforming to the anti-backdoor learning application scenario, that is, the backdoor is introduced via data poisoning only. In the following, we provide the implementation details of each attack.

\paragraph{\badnets Attack~\cite{Gu2017badnet}} We apply a $2 \times 2$ square pattern with colors and a small apple pattern for poisoning small-scale datasets and \tinyimagenet dataset (as used in defenses \dbd~\cite{Huang2022backdoor} and \asd~\cite{Gao2023asd}), respectively.

\paragraph{\trojan Attack~\cite{Liu2018trojan}} We set the trigger mask size as $3\times3$ at the bottom right corner for all different datasets. The optimization setting for trigger generation is executed by following the original \trojan's implementation.

\paragraph{\blend Attack~\cite{Chen2017blend}} Following the original implementation, we choose the Hello-Kitty pattern as the trigger for poisoning \cifar{10} and \gtsrb, and the random noise trigger for \tinyimagenet. We use the trigger pattern opacity $\alpha=0.1$, as it is proved sufficient to achieve an \perc{100} attack success rate.

\paragraph{\clb Attack~\cite{turner2019labelconsistent}} We implement the \clbs (\clb) attack on all considered datasets but with using poisoning rate of \perc{50} and \perc{100} for two small-scale datasets and \tinyimagenet, respectively. We poison each sample by adopting the projected gradient descent (PGD) method to generate adversarial perturbation with strength $\epsilon=\nicefrac{16}{255}$ and step size $\nicefrac{2}{255}$ for 30 steps.

\paragraph{\iab Attack~\cite{Nguyen2020IAB}} \iab poisons the dataset by using a trigger generator that adaptively generates a trigger pattern on each sample. We train the generator model for each dataset with the default hyper-parameters: the backdoor probability $\rho_{b}=0.1$, the cross-trigger probability $\rho_{c}=0.1$, and the weighting parameter $\lambda_{div}=1.0$ for the diversity loss term.

\paragraph{\wanet Attack~\cite{Nguyen2021wanet}} \wanet originally runs the attack during training, which contradicts the threat model of anti-backdoor learning. Therefore, we poison the dataset by running \wanet by only once. We first generate a warping-based trigger function and its accompanied noise function. Then, we use both pattern functions to poison the dataset. For all considered datasets, we following the default settings of \wanet, \ie using the noise ratio of \num{0.2}, the grid size $k=4$, and the warping strength $s=0.5$.

\subsection{Baseline Defenses}
\label{app:defense-setup}

In this section, we outline the implementation details of all considered defenses that require no reference clean dataset.

\begin{table*}[!t]
	\caption{Comparison of \ourmethod with prior defenses on \gtsrb.
		All results are shown in \%. The best results across all defenses are highlighted in \textbf{bold} font. Settings where the defense fails (\asr > \perc{90}) are marked in {\HLFail orange bold} font.}
	\label{tab:compare-sota-gtsrb}
	\tablesize
	\hspace*{-11.5pt}
	\begin{tikzpicture}
		\newcommand{\boxy}{+0.3}
		\draw [fill=tabbgcolor,draw=tabbgcolor] (-6.125,\boxy) rectangle ($(8.97,\boxy)-(0,0.80)$);
		\renewcommand{\boxy}{-2.55}
		\draw [fill=tabbgcolor,draw=tabbgcolor] (-6.125,\boxy) rectangle ($(8.97,\boxy)-(0,0.80)$);
		\node (table) {\newcommand{\mycsvreader}[3]{%
	\csvreader[
	head to column names,
	filter = \equal{\dataset}{#1} \and \equal{\arch}{#2} \and 
	\equal{\attack}{#3},
	late after line=\\,
	]%
	{res/main_results.csv}%
	{}
	{
		& \noacctxt
		& \noasrtxt
		& \ablacctxt
		& \ablasrtxt
		& \dbdacctxt
		& \dbdasrtxt
		& \dstacctxt
		& \dstasrtxt
		& \cbdacctxt
		& \cbdasrtxt
		& \oursacctxt
		& \oursasrtxt
	}
}

\begin{overalltable}{\textwidth}
	\multirow{16}{*}{\rotatebox{90}{\gtsrb}}
	& \multirow{8}{*}{\wrn}
	& \badnets	\mycsvreader{gtsrb}{wrn}{badnets}
	&
	& \trojan	\mycsvreader{gtsrb}{wrn}{trojan}
	&
	& \blend	\mycsvreader{gtsrb}{wrn}{blend}
	&
	& \clb		\mycsvreader{gtsrb}{wrn}{clb}
	&
	& \iab		\mycsvreader{gtsrb}{wrn}{iab}
	&
	& \wanet	\mycsvreader{gtsrb}{wrn}{wanet}
	\cmidrule{3-15}
	&
	& \avg		\mycsvreader{gtsrb}{wrn}{avg}
	&
	& \worst	\mycsvreader{gtsrb}{wrn}{worstcase}
	\cmidrule{2-15}
	& \multirow{8}{*}{\resnet{18}}
	& \badnets	\mycsvreader{gtsrb}{resnet18}{badnets}
	&
	& \trojan	\mycsvreader{gtsrb}{resnet18}{trojan}
	&
	& \blend	\mycsvreader{gtsrb}{resnet18}{blend}
	&
	& \clb		\mycsvreader{gtsrb}{resnet18}{clb}
	&
	& \iab		\mycsvreader{gtsrb}{resnet18}{iab}
	&
	& \wanet	\mycsvreader{gtsrb}{resnet18}{wanet}
	\cmidrule{3-15}
	&
	& \avg		\mycsvreader{gtsrb}{resnet18}{avg}
	&
	& \worst	\mycsvreader{gtsrb}{resnet18}{worstcase}
	\bottomrule
\end{overalltable}
};
	\end{tikzpicture}
\end{table*}

\paragraph{\abl~\citep{Li2021antibackdoor}}
The \abl's procedure consists of three stages: (1) backdoor training for 20 epochs on the entire poisoned dataset with LGA loss and consequently isolating \perc{1} samples by lowest loss, (2) model fine-tuning on the remaining dataset, and (3) backdoor unlearning the isolated poisonous set for 5 epochs with a learning rate of 0.0001. 
Hyper-parameter $\gamma$ in LGA is sensitive to different attacks, as the learning speed of backdoor varies with different trigger pattern. Backdoor suppression can be inefficient with too small $\gamma$, while a large $\gamma$ may suppress the learning on benign samples, leading to a clean accuracy degradation. We use the best defensive performance through $\gamma$ in $\left\{0.0, 0.1, 0.2, 0.3, 0.4, 0.5\right\}$ for the comparison.

\paragraph{\dbd~\citep{Huang2022backdoor}} 
\dbd uses self-supervised learning to train the model for \num{1000} epochs to extract benign features from the dataset into the model. Then, it fine-tunes the fully connected layer for 10 epochs with supervised learning, which in fact learns benign samples faster than poisonous samples. Hence, poisonous samples have higher \sce loss~\cite{Wang2019symmetric}, which makes them distinguishable from the entire dataset. By splitting the dataset half-and-half, \dbd adopts semi-supervised learning (\ssl), \ie \mixmatch~\cite{Berthelot2019mixmatch}, to train the final clean model. 
The supervised learning of \ssl trains the model on most of benign samples with labels to improve the clean accuracy. The unsupervised learning extracts benign features from poisonous samples without labels to further erase the backdoor.
\dbd doesn't specify other hyper-parameters for individual backdoor attack. Therefore, we directly follow the default setting of \dbd to conduct all experiments.

\paragraph{\dst~\citep{Chen2022DST}}
\dst exploits the higher transformation sensitivity of poisonous samples for the splitting. The secure training of \dst consists of three stages: (1) splitting the dataset to a poisoned set, a clean set and a suspicious set, (2) training the feature extractor on the three split sets via a semi-supervised contrastive learning, and (3) minimizing a mixed loss function to learn a benign classifier. In stage (1), we set the same fixed splitting ratios $\alpha_{c}=\perc{20}$ and $\alpha_{p}=\perc{5}$ for composing the clean set and the poisoned set, respectively. In stage (2), we adopt \num{200} epochs for semi-supervised contrastive learning. Finally, we follow the suggestion of using $\lambda_{p} = 0.001$ and 10 epochs training to minimize the mixed loss function in stage (3). All different data transformations keep the same as in \dst.  

\paragraph{\cbd~\citep{Zhang2023backdoor}}
\cbd first learns a model on the entire poisoned dataset, and then train another model by enlarging the mutual independency to the former model. Since the first model is expected to be fully backdoored and, thus, able to tell poisonous samples, the number of training epochs in the first phase is a sensitive hyper-parameter with different attacks. \citet{Zhang2023backdoor} use the best training epochs in the first phase from $\left\{3, 5, 8\right\}$. Accordingly, we tested different training epochs and used the result with the highest natural accuracy for the comparison.

\paragraph{\ourmethod~(Ours)}
%
In \ourmethod's \stage{1}~-~\threeb, we use \adam optimizer and set the learning rate of~\num{0.001} to ensure a faster convergence to the poisoned set \poisonD. \revision{In \stage{2}, we set the number of iterations $\Niters=20$.} In \stage{4}, we use SGD optimizer with the weight decay \num{0.0005} and the momentum \num{0.9}, and we step-wise schedule the learning rate from~\num{0.1} to~\num{0.01}, ~\num{0.001} and \num{0.0001} at epoch \num{50}, \num{75} and \num{90}, respectively. Note that we remove all data augmentations in \stage{1}~-~\threeb to stabilize the learning of backdoor~\cite{Qiu2021DeepSweep, Li2021antibackdoor}, and in \stage{4}, we activate all necessary data augmentations for achieving a high natural accuracy in the final clean model.


\begin{table*}[!t]
	\caption{Comparing the dataset splitting methods on \gtsrb. Precision (\precision) measures the ratio of poisonous samples in \poisonD. \recall shows the isolation percentage of poisonous samples. $\pratio_{bng}$ is the poisoning ratio in \cleanD. All results are shown in \perc{}. We highlight the best \Fone score and $\pratio_{bng}$ of all defenses in \textbf{bold} font. $\pratio_{bng}$ above \perc{5} is marked in {\HLFail orange bold} font.
	}
	\label{tab:comp_split_gtsrb}
	\hspace*{-12pt}
	\begin{tikzpicture}
		\newcommand{\boxy}{+0.3}
		\draw [fill=tabbgcolor,draw=tabbgcolor] (-6.16,\boxy) rectangle ($(8.97,\boxy)-(0,0.80)$);
		\renewcommand{\boxy}{-2.55}
		\draw [fill=tabbgcolor,draw=tabbgcolor] (-6.16,\boxy) rectangle ($(8.97,\boxy)-(0,0.80)$);
		\node (table) {\newcommand{\splitreader}[3]{%
	\csvreader[
	head to column names,
	filter = \equal{\dataset}{#1} \and \equal{\arch}{#2} \and 
	\equal{\attack}{#3},
	late after line=\\,
	]%
	{res/compare_split.csv}%
	{}
	{
		& \dbdPrec
		& \dbdRecall
		& \dbdFone
		& \dbdPrate
		& \dstPrec
		& \dstRecall
		& \dstFone
		& \dstPrate
		& \oursPrec
		& \oursRecall
		& \oursFone
		& \oursPrate
	}
}

\begin{splittable}{\textwidth}
	\multirow{16}{*}{\rotatebox{90}{\gtsrb}}
	& \multirow{8}{*}{\wrn}
	& \badnets	\splitreader{gtsrb}{wrn}{badnets}
	&
	& \trojan	\splitreader{gtsrb}{wrn}{trojan}
	&
	& \blend	\splitreader{gtsrb}{wrn}{blend}
	&
	& \clb		\splitreader{gtsrb}{wrn}{clb}
	&
	& \iab		\splitreader{gtsrb}{wrn}{iab}
	&
	& \wanet	\splitreader{gtsrb}{wrn}{wanet}
	\cmidrule{3-15}
	&
	& \avg		\splitreader{gtsrb}{wrn}{avg}
	&
	& \worst	\splitreader{gtsrb}{wrn}{worstcase}
	\cmidrule{2-15}
	& \multirow{8}{*}{\resnet{18}}
	& \badnets	\splitreader{gtsrb}{resnet18}{badnets}
	&
	& \trojan	\splitreader{gtsrb}{resnet18}{trojan}
	&
	& \blend	\splitreader{gtsrb}{resnet18}{blend}
	&
	& \clb		\splitreader{gtsrb}{resnet18}{clb}
	&
	& \iab		\splitreader{gtsrb}{resnet18}{iab}
	&
	& \wanet	\splitreader{gtsrb}{resnet18}{wanet}
	\cmidrule{3-15}
	&
	& \avg		\splitreader{gtsrb}{resnet18}{avg}
	&
	& \worst	\splitreader{gtsrb}{resnet18}{worstcase}
	\bottomrule
\end{splittable}};
	\end{tikzpicture}
\end{table*}

\section{Experiments on \gtsrb}
\label{app:expm-gtsrb}

For the dataset \gtsrb, both considered baseline models have the sufficient capacity as they both produce a similar natural accuracy by the naive training on a poisoned dataset (\cf~\cref{tab:compare-sota-gtsrb}). 
Prior defenses can resist the backdooring attack of \badnets, \trojan and \clb, while their robustness significantly decreases under other attacks that are stealthier. Since learning on \gtsrb is easier than \cifar{10}, particularly for \wrn, for the defense \cbd, the contrast between a benign model and a poisoned model is not clear, which leads to a defense failure (against \eg \blend) and a degradation of natural accuracy. 

For defenses of dataset splitting, the fixed splitting ratio makes \dbd and \dst either inefficient in isolating poisonous samples, or filtering out many benign samples by mistake (\cf~\cref{tab:comp_split_gtsrb}). Differently, \ourmethod shows the ability to robustly isolate poisonous samples from each poisoned \gtsrb dataset. The high efficiency in the dataset splitting, \ie high \Fone and $\pratio_{bng}$ below \perc{0.1} (\cf~\cref{tab:comp_split_gtsrb}), allows \ourmethod to successfully eliminate any backdoor attack and preserve the natural accuracy in the final clean model. 

\label{app:compare-gtsrb}

\section{Additional Ablation Study on \ourmethod}
\label{app:ablation-study-settings}


In this section, we investigate \ourmethod's performance from five different aspects: 
(1)~hyper-settings (\cf~\cref{app:hyper-settings}), 
(2)~cross backdoor target labels (\cf~\cref{app:bd-targets}), 
(3)~cross model architectures (\cf~\cref{app:cross-arch}), 
(4)~cross different poisoning rates (\cf~\cref{app:cross-prates})
and (5)~the performance on clean dataset (\cf~\cref{app:clean-dataset}).

\subsection{Hyper-Settings}
\label{app:hyper-settings}
Besides the ablation study in \cref{sec:ablation-study}, we investigate the influence of four different settings and summarize \ourmethod's performance of varying every setting in \cref{fig:hyper-params}. 

\begin{enumerate}[label=\bf (\alph*), wide=0em]
	\item \textbf{$\gamma$ value in \lgaloss.}
	The training with the LGA loss traps poisonous samples with the low loss. \ourmethod performs stably in the dataset splitting while $\gamma < 0.015$. By increasing $\gamma$ to \num{0.02}, the learning by LGA loss makes poisonous samples with the dynamic triggers, \ie \iab and \wanet, less learned by the model. Thus, the consequential unlearning step fosters a mis-splitting on them, in particular for \iab attack.   
	
	\item \textbf{Factor $\lambda$ used in unlearning.}
	The unlearning step works to enlarge the prediction loss on samples of \cleanD. Similar to $\gamma$ in LGA loss, \ourmethod can split the dataset effectively by using a small $\lambda$ value in the unlearning step, resulting the final clean model with the high \acc and very low \asr. However, $\lambda \geq 0.015$ make either the natural performance decrease or the splitting ineffective.
	
	\begin{figure}[!b]
		\hskip -3pt
		\begin{subfigure}{.435\linewidth}
			\vskip -11pt
			\pgfplotstableread[col sep=space]{%
	lgaGamma	acc	asr	acc	asr	acc	asr	acc	asr
	0.5	92.65	0.42	92.56	0.31	92.87	0.65	92.64	1.34
	1	93.34	0.56	93.37	0.58	93.11	0.77	93.56	1.22
	1.5	93.30	0.72	93.13	0.42	92.83	0.61	91.49	3.37
	2	92.94	0.92	93.00	98.30	92.52	99.94	90.54	7.86
}\subsettrain

\begin{tikzpicture}	
	\begin{axis}[
		height=.6\linewidth,
		width=.85\linewidth, 
		enlarge x limits=0.02,
		enlarge y limits=0.04,
		grid=major, 
		grid style={dashed,gray!30},
		xlabel= {$\gamma$ in \lgaloss}, 
		ylabel= {\acc in \%},
		xmin=0.5,
		xmax=2,
		ymin=75,
		ymax=95,
		xtick={0.5, 1, 1.5, 2},
		xticklabels={0.005, 0.010, 0.015, 0.020},
		ytick={75, 80, 85, 90, 95},
		yticklabels={75, 80, 85, 90, 95},
		yticklabel style = {font=\scriptsize, yshift=0ex},
		xticklabel style = {font=\scriptsize},
		ylabel style = {font=\footnotesize, yshift=-4ex},
		xlabel style = {font=\footnotesize, yshift=1ex},
		legend image post style={scale=1.0},
		legend style={at={(0.05,0.8)}, font=\legendsize, anchor=north west, legend columns=1, fill=white, draw=white, nodes={scale=1.0, transform shape}, column sep=1pt, line width=1.5pt},
		legend cell align={left},
		scale only axis,
		every axis plot/.append style={line width=0.7pt},
		]
		
		\addplot+[mark=square*, mark options={solid, mark size=.7pt}, solid, color=primarycolor, line width=1pt] table[x index = 0, y index=1] 
		{\subsettrain};
		
		\addplot+[mark=triangle*, mark options={solid, mark size=.7pt}, solid, color=secondarycolor, line width=1pt] table[x index = 0, y index=3] 
		{\subsettrain};
		
		\addplot+[mark=diamond*, mark options={solid, mark size=.7pt}, solid, color=tertiarycolor, line width=1pt] table[x index = 0, y index=5] 
		{\subsettrain};
		
		\addplot+[mark=*, mark options={solid, mark size=.7pt}, solid, color=black, line width=1pt] table[x index = 0, y index=7] 
		{\subsettrain};
		
		\addlegendentry{\badnets};
		\addlegendentry{\blend};
		\addlegendentry{\iab};
		\addlegendentry{\wanet};
		
	\end{axis}
	
	\begin{axis}[
		height=.6\linewidth,
		width=.85\linewidth, 
		enlarge x limits=0.02,
		enlarge y limits=0.04,
		xlabel= \empty, 
		ylabel= \empty, 
		xmin=0.5,
		xmax=2,
		ymin=0,
		ymax=100,
		xtick={0.5, 1, 1.5, 2},
		xticklabels=\empty,
		ytick={0, 25, 50, 75, 100},
		yticklabels=\empty,
		ylabel near ticks, yticklabel pos=right,
		yticklabel style = {font=\scriptsize, yshift=0ex},
		xticklabel style = {font=\scriptsize},
		ylabel style = {font=\footnotesize, yshift=1.0ex},
		xlabel style = {font=\footnotesize, yshift=2.2ex},
		legend image post style={scale=0.5},
		legend style={at={(0.99,0.80)}, font=\legendsize, anchor=north east, 
			legend columns=1, fill=white, draw=white, nodes={scale=0.7, transform shape}, column sep=0pt, line width=.5pt},
		legend cell align={left},
		scale only axis,
		every axis plot/.append style={line width=0.7pt},
		]
		
		\addplot+[mark=square*, mark options={solid, mark size=.7pt}, densely dashed, color=primarycolor, line width=1pt] table[x index = 0, y index=2] 
		{\subsettrain};
		
		\addplot+[mark=triangle*, mark options={solid, mark size=.7pt}, densely dashed, color=secondarycolor, line width=1pt] table[x index = 0, y index=4] 
		{\subsettrain};
		
		\addplot+[mark=diamond*, mark options={solid, mark size=.7pt}, densely dashed, color=tertiarycolor, line width=1pt] table[x index = 0, y index=6] 
		{\subsettrain};
		
		\addplot+[mark=*, mark options={solid, mark size=.7pt}, densely dashed, color=black, line width=1pt] table[x index = 0, y index=8] 
		{\subsettrain};
		
		
	\end{axis}
	
\end{tikzpicture}
			
			\vskip -6pt
			\pgfplotstableread[col sep=space]{%
	epoch	acc	asr	acc	asr	acc	asr	acc	asr
	6	93.41	0.83	92.30	1.47	93.55	2.94	91.53	5.72
	8	92.59	0.64	93.36	1.12	93.34	1.44	92.49	0.56
	10	93.34	0.56	93.37	0.58	93.11	0.77	93.56	1.22
	12	92.23	0.38	92.90	0.47	93.00	0.32	92.82	0.40
	14	91.97	0.46	92.04	0.31	92.49	0.25	91.06	0.31
}\epochtrain

\begin{tikzpicture}	
	\begin{axis}[
		height=.6\linewidth,
		width=.85\linewidth, 
		enlarge x limits=0.02,
		enlarge y limits=0.04,
		grid=major, 
		grid style={dashed,gray!30},
		xlabel= {Epochs \EpochsPoi}, 
		ylabel= {\acc in \%},
		xmin=6,
		xmax=14,
		ymin=75,
		ymax=95,
		xtick={6,8,10,12,14},
		xticklabels={6,8,10,12,14},
		ytick={75, 80, 85, 90, 95},
		yticklabels={75, 80, 85, 90, 95},
		yticklabel style = {font=\scriptsize, yshift=0ex},
		xticklabel style = {font=\scriptsize},
		ylabel style = {font=\footnotesize, yshift=-4ex},
		xlabel style = {font=\footnotesize, yshift=0.7ex},
		legend image post style={scale=0.5},
		legend style={at={(0.99,0.80)}, font=\legendsize, anchor=north east, 
			legend columns=1, fill=white, draw=white, nodes={scale=0.7, transform shape}, column sep=0pt, line width=.5pt},
		legend cell align={left},
		scale only axis,
		every axis plot/.append style={line width=0.7pt},
		]
		
		\addplot+[mark=square*, mark options={solid, mark size=.7pt}, solid, color=primarycolor, line width=1pt] table[x index = 0, y index=1] 
		{\epochtrain};
		
		\addplot+[mark=triangle*, mark options={solid, mark size=.7pt}, solid, color=secondarycolor, line width=1pt] table[x index = 0, y index=3] 
		{\epochtrain};
		
		\addplot+[mark=diamond*, mark options={solid, mark size=.7pt}, solid, color=tertiarycolor, line width=1pt] table[x index = 0, y index=5] 
		{\epochtrain};
		
		\addplot+[mark=*, mark options={solid, mark size=.7pt}, solid, color=black, line width=1pt] table[x index = 0, y index=7] 
		{\epochtrain};

	\end{axis}
	
	\begin{axis}[
		height=.6\linewidth,
		width=.85\linewidth, 
		enlarge x limits=0.02,
		enlarge y limits=0.04,
		xlabel= \empty, 
		ylabel= \empty, 
		xmin=6,
		xmax=14,
		ymin=0,
		ymax=20,
		xtick={6,8,10,12,14},
		xticklabels=\empty,
		ytick={0, 5, ..., 20},
		yticklabels=\empty, 
		ylabel near ticks, yticklabel pos=right,
		yticklabel style = {font=\scriptsize, yshift=0.3ex},
		xticklabel style = {font=\scriptsize},
		ylabel style = {font=\footnotesize, yshift=1.0ex},
		xlabel style = {font=\footnotesize, yshift=2.2ex},
		legend image post style={scale=0.5},
		legend style={at={(0.015,0.71)}, font=\legendsize, anchor=north west, 
			legend columns=2, fill=white, draw=white, nodes={scale=0.7, transform shape}, column sep=0pt, line width=.5pt},
		legend cell align={left},
		scale only axis,
		every axis plot/.append style={line width=0.7pt},
		]
		
		\addplot+[mark=square*, mark options={solid, mark size=.7pt}, densely dashed, color=primarycolor, line width=1pt] table[x index = 0, y index=2] 
		{\epochtrain};
		
		\addplot+[mark=triangle*, mark options={solid, mark size=.7pt}, densely dashed, color=secondarycolor, line width=1pt] table[x index = 0, y index=4] 
		{\epochtrain};
		
		\addplot+[mark=diamond*, mark options={solid, mark size=.7pt}, densely dashed, color=tertiarycolor, line width=1pt] table[x index = 0, y index=6] 
		{\epochtrain};
		
		\addplot+[mark=*, mark options={solid, mark size=.7pt}, densely dashed, color=black, line width=1pt] table[x index = 0, y index=8] 
		{\epochtrain};
		
		
	\end{axis}
	
\end{tikzpicture}
		\end{subfigure}
		\hskip 4pt
		\begin{subfigure}{.435\linewidth}
			\pgfplotstableread[col sep=space]{%
	ulGamma	acc	asr	acc	asr	acc	asr	acc	asr
	0.5	93.29	0.69	92.85	0.63	93.10	0.31	93.14	0.96
	1.0	93.34	0.56	93.37	0.58	93.11	0.77	93.56	1.22
	1.5	93.19	0.70	93.31	0.78	93.21	1.04	93.21	8.25
	2.0	93.07	0.83	93.66	2.38	93.98	96.03	92.96	14.27
}\subsettrain

\begin{tikzpicture}	
	\begin{axis}[
		height=.6\linewidth,
		width=.85\linewidth, 
		enlarge x limits=0.02,
		enlarge y limits=0.04,
		grid=major, 
		grid style={dashed,gray!30},
		xlabel= {$\lambda$ in \ulloss}, 
		ylabel= \empty, 
		xmin=0.5,
		xmax=2,
		ymin=75,
		ymax=95,
		xtick={0.5, 1, 1.5, 2},
		xticklabels={0.005, 0.010, 0.015, 0.020},
		ytick={75, 80, 85, 90, 95},
		yticklabels=\empty, 
		yticklabel style = {font=\scriptsize, yshift=0ex},
		xticklabel style = {font=\scriptsize},
		ylabel style = {font=\footnotesize, yshift=-4ex},
		xlabel style = {font=\footnotesize, yshift=1.1ex},
		legend image post style={scale=1.0},
		legend style={at={(0.0,0.65)}, font=\legendsize, anchor=north west, 
			legend columns=2, fill=white, draw=white, nodes={scale=.8, transform shape}, column sep=1pt, line width=1.5pt},
		legend cell align={left},
		scale only axis,
		every axis plot/.append style={line width=0.7pt},
		]
		
		\addplot+[mark=square*, mark options={solid, mark size=.7pt}, solid, color=primarycolor, line width=1pt] table[x index = 0, y index=1] 
		{\subsettrain};
		
		\addplot+[mark=triangle*, mark options={solid, mark size=.7pt}, solid, color=secondarycolor, line width=1pt] table[x index = 0, y index=3] 
		{\subsettrain};
		
		\addplot+[mark=diamond*, mark options={solid, mark size=.7pt}, solid, color=tertiarycolor, line width=1pt] table[x index = 0, y index=5] 
		{\subsettrain};
		
		\addplot+[mark=*, mark options={solid, mark size=.7pt}, solid, color=black, line width=1pt] table[x index = 0, y index=7] 
		{\subsettrain};
		
		
	\end{axis}
	
	\begin{axis}[
		height=.6\linewidth,
		width=.85\linewidth, 
		enlarge x limits=0.02,
		enlarge y limits=0.04,
		xlabel= \empty, 
		ylabel= {\asr in \%},
		xmin=0.5,
		xmax=2,
		ymin=0,
		ymax=100,
		xtick={0.5, 1, 1.5, 2},
		xticklabels=\empty,
		ytick={0, 25, 50, 75, 100},
		yticklabels={0, 25, 50, 75, 100},
		ylabel near ticks, yticklabel pos=right,
		yticklabel style = {font=\scriptsize, yshift=0ex},
		xticklabel style = {font=\scriptsize},
		ylabel style = {font=\footnotesize, yshift=1.0ex},
		xlabel style = {font=\footnotesize, yshift=2ex},
		legend image post style={scale=0.5},
		legend style={at={(0.99,0.80)}, font=\legendsize, anchor=north east, 
			legend columns=1, fill=white, draw=white, nodes={scale=0.7, transform shape}, column sep=0pt, line width=.5pt},
		legend cell align={left},
		scale only axis,
		every axis plot/.append style={line width=0.7pt},
		]
		
		\addplot+[mark=square*, mark options={solid, mark size=.7pt}, densely dashed, color=primarycolor, line width=1pt] table[x index = 0, y index=2] 
		{\subsettrain};
		
		\addplot+[mark=triangle*, mark options={solid, mark size=.7pt}, densely dashed, color=secondarycolor, line width=1pt] table[x index = 0, y index=4] 
		{\subsettrain};
		
		\addplot+[mark=diamond*, mark options={solid, mark size=.7pt}, densely dashed, color=tertiarycolor, line width=1pt] table[x index = 0, y index=6] 
		{\subsettrain};
		
		\addplot+[mark=*, mark options={solid, mark size=.7pt}, densely dashed, color=black, line width=1pt] table[x index = 0, y index=8] 
		{\subsettrain};
		
		
	\end{axis}
	
\end{tikzpicture}
			
			\vskip -5pt
			\pgfplotstableread[col sep=space]{%
	epoch	acc	asr	acc	asr	acc	asr	acc	asr
	1	93.34	0.56	93.37	0.58	93.11	0.77	93.56	1.22
	3	93.16	0.69	92.78	0.39	92.23	1.39	92.37	1.34
	5	92.98	0.57	91.36	0.38	91.30	0.76	91.48	0.89
	7	92.31	0.50	87.89	0.04	88.43	0.08	89.24	0.75
	9	91.98	0.51	85.10	0.03	85.20	0.02	85.34	0.38
}\epochtrain

\begin{tikzpicture}	
	\begin{axis}[
		height=.6\linewidth,
		width=.85\linewidth, 
		enlarge x limits=0.02,
		enlarge y limits=0.04,
		grid=major, 
		grid style={dashed,gray!30},
		xlabel= {\paramsref[\bf i] used in \stage{3}}, 
		ylabel= \empty, 
		xmin=1,
		xmax=9,
		ymin=75,
		ymax=95,
		xtick={1,3,5,7,9},
		xticklabels={1,3,5,7,9},
		ytick={75, 80, 85, 90, 95},
		yticklabels=\empty, 
		yticklabel style = {font=\scriptsize, yshift=0ex},
		xticklabel style = {font=\scriptsize},
		ylabel style = {font=\footnotesize, yshift=-4ex},
		xlabel style = {font=\footnotesize, yshift=1.6ex},
		legend image post style={scale=0.5},
		legend style={at={(0.99,0.80)}, font=\legendsize, anchor=north east, 
			legend columns=1, fill=white, draw=white, nodes={scale=0.7, transform shape}, column sep=0pt, line width=.5pt},
		legend cell align={left},
		scale only axis,
		every axis plot/.append style={line width=0.7pt},
		]
		
		\addplot+[mark=square*, mark options={solid, mark size=.7pt}, solid, color=primarycolor, line width=1pt] table[x index = 0, y index=1] 
		{\epochtrain};
		
		\addplot+[mark=triangle*, mark options={solid, mark size=.7pt}, solid, color=secondarycolor, line width=1pt] table[x index = 0, y index=3] 
		{\epochtrain};
		
		\addplot+[mark=diamond*, mark options={solid, mark size=.7pt}, solid, color=tertiarycolor, line width=1pt] table[x index = 0, y index=5] 
		{\epochtrain};
		
		\addplot+[mark=*, mark options={solid, mark size=.7pt}, solid, color=black, line width=1pt] table[x index = 0, y index=7] 
		{\epochtrain};

	\end{axis}
	
	\begin{axis}[
		height=.6\linewidth,
		width=.85\linewidth, 
		enlarge x limits=0.02,
		enlarge y limits=0.04,
		xlabel= \empty, 
		ylabel= {\asr in \%},
		xmin=1,
		xmax=9,
		ymin=0,
		ymax=20,
		xtick={1,3,5,7,9},
		xticklabels=\empty,
		ytick={0, 5, ..., 20},
		yticklabels={0, 5, 10, 15, 20},
		ylabel near ticks, yticklabel pos=right,
		yticklabel style = {font=\scriptsize, yshift=0ex},
		xticklabel style = {font=\scriptsize},
		ylabel style = {font=\footnotesize, yshift=0.2ex},
		xlabel style = {font=\footnotesize, yshift=2.2ex},
		legend image post style={scale=0.5},
		legend style={at={(0.015,0.71)}, font=\legendsize, anchor=north west, 
			legend columns=2, fill=white, draw=white, nodes={scale=0.7, transform shape}, column sep=0pt, line width=.5pt},
		legend cell align={left},
		scale only axis,
		every axis plot/.append style={line width=0.7pt},
		]
		
		\addplot+[mark=square*, mark options={solid, mark size=.7pt}, densely dashed, color=primarycolor, line width=1pt] table[x index = 0, y index=2] 
		{\epochtrain};
		
		\addplot+[mark=triangle*, mark options={solid, mark size=.7pt}, densely dashed, color=secondarycolor, line width=1pt] table[x index = 0, y index=4] 
		{\epochtrain};
		
		\addplot+[mark=diamond*, mark options={solid, mark size=.7pt}, densely dashed, color=tertiarycolor, line width=1pt] table[x index = 0, y index=6] 
		{\epochtrain};
		
		\addplot+[mark=*, mark options={solid, mark size=.7pt}, densely dashed, color=black, line width=1pt] table[x index = 0, y index=8] 
		{\epochtrain};
		
		
	\end{axis}
	
\end{tikzpicture}
		\end{subfigure}
		\caption{Ablation study on \ourmethod's hyper-settings with the model \resnet{18} on poisoned \cifar{10} datasets. \acc and \asr are shown in solid and dashed lines, respectively.}
		\label{fig:hyper-params}
	\end{figure}
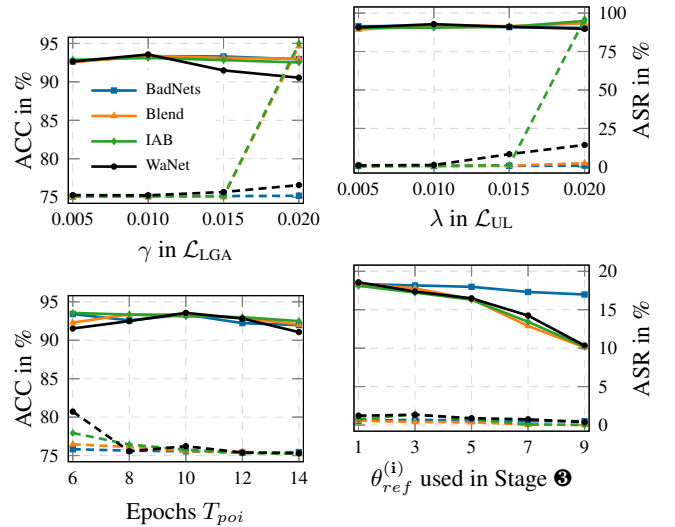
	
	\item \textbf{Number of epochs \EpochsPoi.}
	Different from the settings above, \ourmethod is less sensitive to the number of epochs \EpochsPoi. Using fewer \EpochsPoi (e.g. \num{6} epochs) introduces a bit instability, making the splitting slightly less precise and, thus, yielding a higher \asr. Using more epochs in each round, although beneficial for the stability, consumes more time in the splitting process. Hence, we use $\EpochsPoi=10$ in \ourmethod's implementation.
	
	\item \textbf{Reference model \paramsref[i] used in \stage{3}.}
	In the meta-splitting, \ourmethod uses the reference model of the first round as the notion to split benign and poisonous samples from the overlap. To demonstrate our choice, we use the reference model from different rounds in the meta-splitting (\ie \stage{3}). As shown in \cref{fig:hyper-params}, the reference model only has the impact on splitting benign samples from the overlap. Using a model \paramsref from a later round (\eg round \num{9}) makes the natural accuracy decrease, as the reference model \paramsref[9] converges to the backdoor more, particularly making the \rce loss of benign samples in the backdoor target class also low. This phenomenon first prove the rationality of using \paramsref[1] in the meta-splitting and second further presents the progressive convergence to the backdoor in \stage{2}. 
	
\end{enumerate}

\subsection{Cross Different Backdoor Target Labels}
\label{app:bd-targets}

We evaluate our method across different target labels, \ie $\labelyt \in \left\{0, 1, 2, 3, 4, 5, 6, 7, 8, 9\right\}$, on each poisoned \cifar{10} dataset with the baseline model \resnet{18}. Despite a slight variance across different target labels, the results in \cref{fig:bd-targets} verify the robustness of our method in the isolation of poisonous samples and the preservation of the natural accuracy.

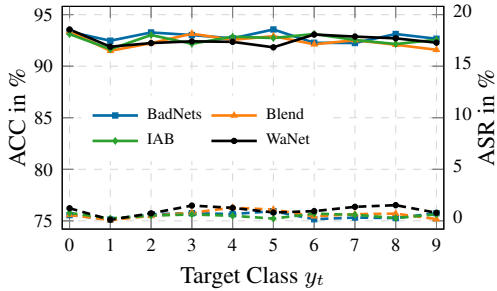
\begin{figure}[!h]
	\centering
	\pgfplotstableread[col sep=space]{%
	target	acc	asr	acc	asr	acc	asr	acc	asr
	0	93.34	0.56	93.37	0.58	93.11	0.77	93.56	1.22
	1	92.47	0.13	91.49	0.08	91.65	0.24	91.89	0.11
	2	93.27	0.64	92.23	0.48	93.02	0.51	92.24	0.73
	3	93.02	0.74	93.15	0.79	92.17	0.62	92.41	1.48
	4	92.70	0.66	92.56	1.28	92.86	0.49	92.36	1.26
	5	93.56	0.94	92.91	1.09	92.73	0.22	91.83	0.81
	6	92.27	0.16	92.12	0.44	93.12	0.70	93.07	0.94
	7	92.25	0.32	92.52	0.65	92.54	0.59	92.87	1.36
	8	93.12	0.27	92.08	0.70	92.15	0.27	92.70	1.52
	9	92.66	0.74	91.59	0.17	92.56	0.63	92.28	0.78
}\targtestrain

\begin{tikzpicture}	
	\begin{axis}[
		height=.35\linewidth,
		width=.6\linewidth, 
		enlarge x limits=0.02,
		enlarge y limits=0.04,
		grid=major, 
		grid style={dashed,gray!30},
		xlabel= {Target Class \labelyt}, 
		ylabel= {\acc in \%},
		xmin=0,
		xmax=9,
		ymin=75,
		ymax=95,
		xtick={0, 1, ..., 9},
		xticklabels={0, 1, 2, 3, 4, 5, 6, 7, 8, 9},
		ytick={75, 80, 85, 90, 95},
		yticklabels={75, 80, 85, 90, 95},
		yticklabel style = {font=\scriptsize, yshift=0ex},
		xticklabel style = {font=\scriptsize},
		ylabel style = {font=\footnotesize, yshift=-4ex},
		xlabel style = {font=\footnotesize, yshift=1ex},
		legend image post style={scale=1.0},
		legend style={at={(0.05,0.6)}, font=\legendsize, anchor=north west, legend columns=2, fill=white, draw=white, nodes={scale=1.0, transform shape}, column sep=1pt, line width=1.5pt},
		legend cell align={left},
		scale only axis,
		every axis plot/.append style={line width=0.7pt},
		]
		
		\addplot+[mark=square*, mark options={solid, mark size=.7pt}, solid, color=primarycolor, line width=1pt] table[x index = 0, y index=1] 
		{\targtestrain};
		
		\addplot+[mark=triangle*, mark options={solid, mark size=.7pt}, solid, color=secondarycolor, line width=1pt] table[x index = 0, y index=3] 
		{\targtestrain};
		
		\addplot+[mark=diamond*, mark options={solid, mark size=.7pt}, solid, color=tertiarycolor, line width=1pt] table[x index = 0, y index=5] 
		{\targtestrain};
		
		\addplot+[mark=*, mark options={solid, mark size=.7pt}, solid, color=black, line width=1pt] table[x index = 0, y index=7] 
		{\targtestrain};
		
		\addlegendentry{\badnets};
		\addlegendentry{\blend};
		\addlegendentry{\iab};
		\addlegendentry{\wanet};
		
	\end{axis}
	
	\begin{axis}[
		height=.35\linewidth,
		width=.6\linewidth, 
		enlarge x limits=0.02,
		enlarge y limits=0.04,
		xlabel= \empty, 
		ylabel= {\asr in \%},
		xmin=0,
		xmax=9,
		ymin=0,
		ymax=20,
		xtick={0, 1, ..., 9},
		xticklabels=\empty,
		ytick={0, 5, 10, 15, 20},
		yticklabels={0, 5, 10, 15, 20},
		ylabel near ticks, yticklabel pos=right,
		yticklabel style = {font=\scriptsize, yshift=0.3ex},
		xticklabel style = {font=\scriptsize},
		ylabel style = {font=\footnotesize, yshift=1.0ex},
		xlabel style = {font=\footnotesize, yshift=2.2ex},
		legend image post style={scale=1.0},
		legend style={at={(0.99,0.80)}, font=\legendsize, anchor=north east, legend columns=1, fill=white, draw=white, nodes={scale=1.0, transform shape}, column sep=0pt, line width=.5pt},
		legend cell align={left},
		scale only axis,
		every axis plot/.append style={line width=0.7pt},
		]
		
		\addplot+[mark=square*, mark options={solid, mark size=.7pt}, densely dashed, color=primarycolor, line width=1pt] table[x index = 0, y index=2] 
		{\targtestrain};
		
		\addplot+[mark=triangle*, mark options={solid, mark size=.7pt}, densely dashed, color=secondarycolor, line width=1pt] table[x index = 0, y index=4] 
		{\targtestrain};
		
		\addplot+[mark=diamond*, mark options={solid, mark size=.7pt}, densely dashed, color=tertiarycolor, line width=1pt] table[x index = 0, y index=6] 
		{\targtestrain};
		
		\addplot+[mark=*, mark options={solid, mark size=.7pt}, densely dashed, color=black, line width=1pt] table[x index = 0, y index=8] 
		{\targtestrain};
		
		
	\end{axis}
	
\end{tikzpicture}
	\caption{Evaluating \ourmethod's defense across different backdoor target labels with a \resnet{18} model on poisoned \cifar{10} datasets. \acc and \asr are shown in solid and dashed lines, respectively.}
	\label{fig:bd-targets}
\end{figure}

\subsection{Cross-Architecture Evaluation}
\label{app:cross-arch}

We implement poisoning attacks with the same settings as mentioned in \cref{app:attack-setup} and evaluate \ourmethod's performance cross different model architectures: \vggeleven~\cite{Simonyan2015Very},
\mobilenet~\cite{Sandler2018mobilenetv2} and \densenet~\cite{Huang2017DenseNet}. As shown in \cref{tab:cross-arch}, our method shows a high adaptability in isolating poisonous samples from each poisoned dataset with different model architectures. At the same time, our method remains robust in preserving a high natural accuracy.

\begin{table}[!t]
	\caption{Cross-architecture evaluation on poisoned \cifar{10} datasets. The method "---" marks \nodefense.}
	\label{tab:cross-arch}
	\tablesize
	\begin{archtable}{\linewidth}
		\multirow{2}{*}{\badnets}
		& ---
		& 90.23	& 99.94	& 92.72	& 99.62	& 94.33	& 100.00 \\
		& \ourmethod
		& 89.80	& 0.51	& 91.72	& 0.63	& 94.30	& 0.57	\\ \midrule
		\multirow{2}{*}{\trojan}
		& --- 
		& 89.76	& 99.98	& 92.47	& 100.00& 94.27	& 99.95	\\
		& \ourmethod
		& 88.10	& 0.67	& 91.33	& 0.51	& 93.30	& 0.50 \\ \midrule
		\multirow{2}{*}{\blend}
		& --- 
		& 89.65	& 99.96	& 92.20	& 99.49	& 94.12	& 99.23	\\
		& \ourmethod
		& 88.35	& 1.07	& 91.69	& 0.42	& 94.01	& 0.50	\\ \midrule
		\multirow{2}{*}{\clb}
		& --- 
		& 90.74	& 99.46	& 92.53	& 99.83	& 94.12	& 99.23	\\
		& \ourmethod
		& 89.89	& 0.76	& 91.33	& 0.36	& 93.16	& 0.39	\\ \midrule
		\multirow{2}{*}{\iab}
		& ---
		& 90.24	& 99.89	& 92.46	& 99.92	& 94.25	& 99.91	\\
		& \ourmethod
		& 89.18	& 1.23	& 91.36 & 0.69	& 93.65	& 1.04	\\ \midrule
		\multirow{2}{*}{\wanet}
		& ---
		& 88.95	& 94.01	& 92.18	& 99.26	& 93.45	& 98.23	\\
		& \ourmethod
		& 86.42	& 2.15	& 91.38	& 0.32	& 92.61	& 1.14	\\ \bottomrule
	\end{archtable}
\end{table}

\subsection{Cross Different Poisoning Rates}
\label{app:cross-prates}
We investigate \ourmethod's ability against backdooring attacks across different poisoning rates.
\ourmethod isolates poisonous samples robustly (\cf~\cref{tab:cross-prates}), while a large number of poisonous samples ($\rho=\perc{20}$) makes the natural accuracy slightly degrade.
For poisoning rates less than \perc{10} (as used in the initial experiments), \ourmethod successfully preserves the natural accuracy while mitigating the attack.

\begin{table}[!h]
	\caption{Evaluation across different poisoning rates.}
	\label{tab:cross-prates}
	\begin{pratestable}{\linewidth}
		\perc{1}
		& 93.67	& 0.47	& 93.06	& 1.37	& 93.59	& 1.06	& 92.68	& 1.13
		\\
		\perc{5}
		& 93.18	& 0.41	& 92.43	& 0.93	& 92.42	& 1.23	& 91.34	& 1.47
		\\
		\perc{10}
		& 93.34	& 0.56	& 93.37	& 0.58	& 93.11	& 0.77	& 93.56	& 1.22
		\\
		\perc{20}
		& 92.13	& 0.58	& 89.64	& 0.06	& 91.24	& 1.64	& 89.56	& 0.47
		\\
		\bottomrule
	\end{pratestable}
\end{table}

\subsection{Performance on Clean Datasets}
\label{app:clean-dataset}

Given a purely clean dataset, our method retains on-par natural accuracy with the original model (\cf~\cref{tab:clean-train}), showing the capability to cope with natural performance preservation when the dataset is not poisoned.

\begin{table}[!h]
	\caption{\ourmethod's performance with clean data. ``Naive'' denotes the natural accuracy by the naive training.}
	\label{tab:clean-train}
	\tablesize
	\begin{nattraintable}{\columnwidth}
		\multirow{2}{*}{\cifar{10}}
		& \resnet{18}	& 93.87 & 93.79
		\\ 
		& \wrn			& 89.42 & 88.52
		\\ \midrule
		\multirow{2}{*}{\gtsrb}
		& \resnet{18}	& 98.10 & 97.38
		\\ 
		& \wrn			& 97.42 & 96.12
		\\ \midrule
		\tinyimagenet
		& \resnet{34}	& 58.35 & 57.78
		\\
		\bottomrule
	\end{nattraintable}
\end{table}

\subsection{Time Consumption in Stages of \ourmethod}
\label{app:time-consume}


\revision{From the time breakdown in \cref{tab:harvey_comp_time}, HARVEY's time overhead primarily comes from \stage{1} and \twob, which handle dataset splitting to isolate poisonous samples. Meanwhile, the final training in \stage{4} is slightly faster than naive training due to the reduced size of the training set.}

\begin{table}[!h]
	\caption{The breakdown of \ourmethod's time consumption.}
	\label{tab:harvey_comp_time}
	\tablesize
	\begin{harveyconsumetable}{\columnwidth}
	\cifar{10}
	& 0.102	& 0.426	& 0.006	& 0.418	& 0.952
	\\
	\gtsrb
	& 0.074	& 0.338	& 0.004	& 0.328	& 0.744
	\\
	\tinyimagenet
	& 0.586	& 2.565	& 0.015	& 2.516	& 5.682
	\\
	\bottomrule
\end{harveyconsumetable}
\end{table}

\section{Robustness to the Adaptive Attack}
\label{app:adaptive-attack}

In the anti-backdoor learning, the complete access to the dataset is available but there is no control of the training process. 
We evaluate our method with an adaptive attack, where the adversary alters the data distribution by duplicating benign samples, aiming to make \ourmethod isolate benign samples rather poisonous samples.

\begin{figure}[!h]
	\centering
	\includegraphics[width=2cm]{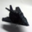}
	\caption{Duplicated Image (Class = 0)}
	\label{fig:dduplicate}
\end{figure}

\begin{table}[!b]
	\caption{Evaluating \ourmethod with a \resnet{18} model against the duplicating attack applied on different poisoned \cifar{10} datasets. All results are shown in \%. ``Target Label'' denotes different methods of the \ddup~attack.}
	\label{tab:ddup-attack}
	\tablesize
	\begin{dduptable}{\columnwidth}
	\multirow{6}{*}{~\poisonTarget $=$ class 0}
	& \badnets	& 92.34	& 0.78	& 62.19	& 0.00
	\\
	& \trojan	& 91.65	& 0.64	& 57.85	& 0.01
	\\
	& \blend	& 91.14	& 0.76	& 66.04	& 0.02 
	\\
	& \clb		& 91.03	& 0.89	& 54.72	& 0.00 
	\\
	& \iab		& 91.42	& 1.12	& 57.41	& 0.00 
	\\
	& \wanet	& 90.38	& 2.57	& 52.76	& 0.51
	\\ 
	\midrule
	%
	\multirow{6}{*}{\makecell{\poisonTarget $\neq$ class 0\\(\poisonTarget $=$ class 1)}}
	& \badnets	& 90.32	& 0.28	& 67.15	& 0.00 
	\\
	& \trojan 	& 90.86	& 0.45	& 72.20	& 0.01 
	\\
	& \blend	& 89.58	& 0.47	& 66.05	& 0.01
	\\
	& \clb		& 89.46	& 0.74	& 69.41	& 0.00
	\\
	& \iab		& 89.00	& 0.00	& 70.71	& 0.00
	\\
	& \wanet	& 88.12	& 0.03	& 69.41	& 0.00
	\\
	\bottomrule
\end{dduptable}
\end{table}

\paragraph{Duplicating benign samples}
To hide the prominence of poisonous samples, an adaptive attacker can alter the dataset distribution by duplicating a single benign sample to establish a fake data cluster with similarly high significance as the real poisonous samples in the dataset.
We choose an image from class~\num{0} (\cf~\cref{fig:dduplicate}) used in the duplication and consider two different duplicating approaches in the \ddup attack: (1)~target label~$=$~class~0 (``Airplane'') and (2)~target label~$\neq$~class~0, for which we choose class~1 (``Automobile'') as the target label.
\cref{tab:ddup-attack} shows \ourmethod's results against the duplicating attack.
Due to the replacement of the single sample duplication, the amount of original benign samples is smaller and thus the clean accuracy decreases. 
For backdoors using dynamic and less visible trigger (\ie \wanet), learning the feature of the duplicated sample is easier, making our method isolate more benign samples alongside poisonous samples, resulting in the splitting with a smaller \Fone and, thus, a lower natural accuracy. Across all attacks with the two proposed \ddup methods, our method yields high performance in the backdoor mitigation.

\fi

\end{document}